%% file: main.tex
\documentclass[11pt]{article}

\usepackage[numbers,compress]{natbib}

\usepackage[hyphens]{url}
\usepackage[pdftex]{hyperref}
\hypersetup{
    unicode=false,          % non-Latin characters in AcrobatÂ¿s bookmarks
    pdftoolbar=true,        % show Acrobat toolbar?
    pdfmenubar=true,        % show Acrobat menu?
    pdffitwindow=false,      % page fit to window when opened
    pdfnewwindow=true,      % links in new window
    colorlinks=true,       % false: boxed links; true: colored links
    linkcolor=DarkBlue,          % color of internal links
    citecolor=DarkGreen,        % color of links to bibliography
    filecolor=DarkRed,      % color of file links
    urlcolor=DarkBlue,          % color of external links
    pdftitle={},
    pdfauthor={},
}
\usepackage[skip=2mm,indent=5mm]{parskip}
\usepackage[small]{caption}
\RequirePackage{amsthm,amsmath,amsfonts,amssymb}

\RequirePackage{graphicx}
\usepackage{subcaption}
\usepackage{multicol}

\usepackage{multirow}
\usepackage[svgnames]{xcolor}
\usepackage[table]{xcolor}
\usepackage{pdflscape}

\newcommand{\myhdashline}{\arrayrulecolor{gray}\hdashline\arrayrulecolor{black}}

\usepackage{colortbl}
\usepackage[normalem]{ulem}
\usepackage{comment}
\usepackage{longtable}
\usepackage{adjustbox}

\usepackage{array}

\usepackage{algorithm}
\usepackage{algpseudocode}

\usepackage{fullpage}
\usepackage[utf8]{inputenc}
\usepackage[T1]{fontenc}
\usepackage{microtype}
\usepackage{csquotes}
\usepackage[osf,sc]{mathpazo}
\RequirePackage[scaled=0.90]{helvet}
\RequirePackage[scaled=0.85]{beramono}
\RequirePackage{textcomp}

\theoremstyle{plain}
\newtheorem{theorem}{Theorem}

\theoremstyle{definition}
\newtheorem{definition}[theorem]{Definition}

\input{macros}

\usepackage{tikz}
\usepackage{subcaption}
\usepackage{enumitem}

\usepackage{fontawesome5}

\usepackage{booktabs}
\usepackage{tcolorbox}
\usepackage{arydshln}

\usepackage{pifont}
\usepackage{wasysym}
\newcommand{\cmark}{\ding{51}}%
\newcommand{\xmark}{\ding{55}}%

\newtcolorbox{practicalbox}{
  colback=blue!5,    % background color
  colframe=cyan!99!black, % border color
  coltitle=black,
  title=Practical Recommendation,
  fonttitle=\bfseries,
  sharp corners,
  boxrule=0.8pt,
  before=\par\medskip\noindent,
  after=\par\medskip
}
\newtcolorbox{practicalboxnotitle}{
  colback=blue!5,    % background color
  colframe=cyan!99!black, % border color
  coltitle=black,
  fonttitle=\bfseries,
  sharp corners,
  boxrule=0.8pt,
  before=\par\medskip\noindent,
  after=\par\medskip
}

\title{\textbf{Large Language Model Hacking:}\\Quantifying the Hidden Risks of Using LLMs\\ for Text Annotation}

\makeatletter
\newcommand{\printfnsymbol}[1]{%
  \textsuperscript{\@fnsymbol{#1}}%
}
\makeatother

\usepackage{authblk}
\author[1]{Joachim Baumann}
\author[1]{Paul Röttger}
\author[2]{Aleksandra Urman}
\author[3]{Albert Wendsjö}
\author[4]{\\Flor Miriam Plaza-del-Arco}
\author[5]{Johannes B. Gruber}
\author[1]{Dirk Hovy}
\affil[1]{Bocconi University}
\affil[2]{University of Zurich}
\affil[3]{University of Gothenburg}
\affil[4]{LIACS, Leiden University}
\affil[5]{GESIS, Leibniz Institute for the Social Sciences}

\date{}

\begin{document}

\maketitle

\begin{abstract}
Large language models (LLMs) are rapidly transforming social science research by enabling the automation of labor-intensive tasks like data annotation and text analysis. However, LLM outputs vary significantly depending on the implementation choices made by researchers (e.g., model selection or prompting strategy). Such variation can introduce systematic biases and random errors, which propagate to downstream analyses and cause Type I (false positive), Type II (false negative), Type S (wrong sign for significant effect), or Type M (correct but exaggerated effect) errors.
We call this phenomenon where configuration choices lead to incorrect conclusions \textit{LLM hacking}.

We find that intentional LLM hacking is strikingly simple.
By replicating 37 data annotation tasks from 21 published social science studies, we show that, with just a handful of prompt paraphrases, virtually anything can be presented as statistically significant.

Beyond intentional manipulation, our analysis of 13 million labels from 18 different LLMs across 2,361 realistic hypotheses shows that there is also a high risk of \textit{accidental} LLM hacking, even when following standard research practices.
We find incorrect conclusions in approximately 31\% of hypotheses for state-of-the-art LLMs, and in half the hypotheses for smaller language models.
While higher task performance and stronger general model capabilities reduce LLM hacking risk, even highly accurate models remain susceptible.
The risk of LLM hacking decreases as effect sizes increase, indicating the need for more rigorous verification of LLM-based findings near significance thresholds.
We analyze 21 mitigation techniques and find that human annotations provide crucial protection against false positives.
Common regression estimator correction techniques can restore valid inference but trade off Type I vs.\ Type II errors.

Overall, our findings advocate for a fundamental shift in LLM-assisted research practices, from viewing LLMs as convenient black-box annotators to seeing them as complex instruments that require rigorous validation. Based on our findings, we publish a list of practical recommendations to prevent intentional LLM hacking and limit the risk of accidental LLM hacking.
\end{abstract}

\begin{figure*}[htb]
    \centering
    \includegraphics[width=\textwidth]{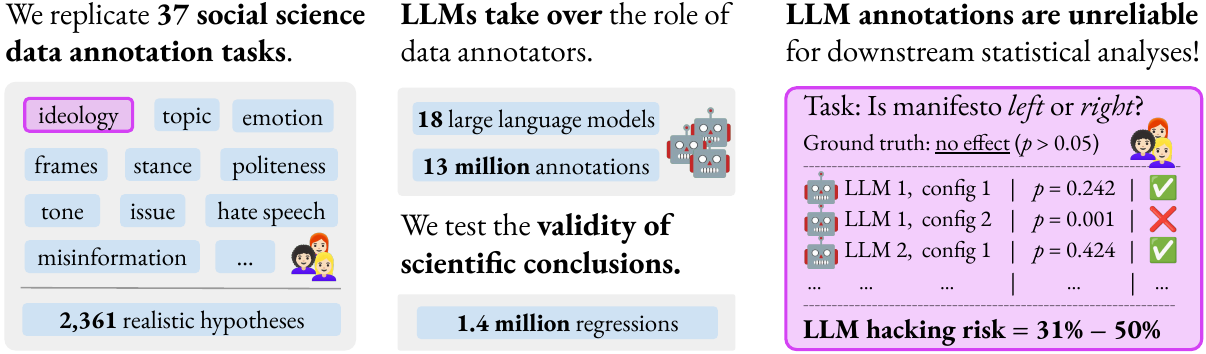}
    \caption{
    (left) We quantify LLM hacking risk through systematic replication of 37 diverse computational social science annotation tasks.
For these tasks, we create a combined set of 2,361 realistic hypotheses that researchers might test using these annotations.
(middle) We collect 13 million LLM annotations across plausible LLM configurations.
These annotations feed into 1.4 million regressions testing the hypotheses. 
(right) For a hypothesis with no true effect (ground truth $p > 0.05$), different LLM configurations yield conflicting conclusions.
Checkmarks indicate correct statistical conclusions matching ground truth; crosses indicate LLM hacking -- incorrect conclusions due to LLM annotation errors.
Across all experiments, LLM hacking occurs in 31-50\% of cases even with highly capable models.
Since minor configuration changes can flip scientific conclusions, from correct to incorrect, LLM hacking can be exploited to present virtually anything as statistically significant.
    }
    \label{fig:figure1}
\end{figure*}

\section{Introduction}

Large language models (LLMs) can easily be instructed to process unstructured data for virtually any analytical purpose~\citep{wei2022finetuned,NEURIPS2022_b1efde53}.
This capability has made it tempting for researchers to outsource time-consuming tasks like data annotation and text analysis to these systems~\citep{liao2025llms}.
LLMs enable the extraction of insights from vast amounts of unstructured text at unprecedented scales~\citep{demszky2023using,ziems-etal-2024-large}.
The rapid integration of LLMs into scientific workflows represents one of the most significant recent methodological shifts in the social sciences and other scientific disciplines.

Unfortunately, beneath this excitement for using LLMs in research lies a fundamental threat to scientific validity that has remained largely unexamined.
Traditionally, computational social scientists used trained student assistants or domain experts to annotate unstructured data to feed their quantitative downstream analyses of interest~\citep{Lombard2002,weber1990basic,krippendorff2018content}.
As the volume of available text data grew exponentially, researchers increasingly turned to computational methods that could scale analysis, either through unsupervised techniques
or by training supervised machine learning (ML) models on human-labeled data~\citep{grimmer2013text,moretti2013distant,Blei2003lds,laver2003extracting}.
In recent years, a large part of the community has enthusiastically embraced the use of LLMs for automated scientific analyses, with many reporting that LLMs can match or even exceed human performance on various annotation tasks~\citep{gilardi2023pnas,ziems-etal-2024-large,tornberg2024large}.
Yet this same literature almost entirely overlooks the fact that a few annotation errors can falsify downstream statistical analyses~\citep{egami2023using,battaglia2024inference,gligoric-etal-2025-unconfident,carlson2025unifying}.
It almost seems as if we tend to forget that high-performing models paired with careful prompt engineering can still produce bad science.
This gap between adoption and validation threatens to undermine the credibility of an entire era of computational social science (CSS) research.

Every LLM-based annotation requires researchers to make numerous configuration choices, including which model to use, how to formulate the prompt, which decoding parameters to set, and how to map outputs to categories.
These choices become a ``garden of forking paths''~\citep{gelman2013garden}, where each decision point branches out into potentially different analytical outcomes.
This effect is magnified by LLMs' sensitivity to seemingly minor input changes~\citep{sclar2024quantifying,salinas-morstatter-2024-butterfly,barrie2024prompt,Atreja_Ashkinaze_Li_Mendelsohn_Hemphill_2025}.
And when LLM outputs determine scientific conclusions -- whether about human behavior, molecular structures, or clinical outcomes -- even minor input variations become consequential~\citep{egami2023using,pangakis2023automated,baly-etal-2020-detect,barrie2024replication}.
For example, when a researcher 
tests whether conservative political discourse contains more economic framing than progressive discourse~\citep{benoit2016crowd,OrnsteinBlasingameTruscott2025} or whether social media posts mentioning certain topics exhibit different sentiment patterns~\citep{gilardi2023pnas}, the conclusion can flip from significant to non-significant based solely on arbitrary researcher decisions (or lack thereof) about how the LLM annotator is configured earlier in the workflow. Insidiously, LLM configuration choices can systematically bias results while maintaining apparent methodological rigor.

In this paper, we introduce the concept of \textit{LLM hacking}, describing the phenomenon where researchers' LLM configuration choices result in incorrect downstream scientific conclusions.\footnote{We use ``hacking'' in the statistical sense (analogous to $p$-hacking). This is unrelated to cybersecurity or unauthorized system intrusion.}
This issue extends beyond simple measurement error.
LLM hacking often emerges from ambiguity about how to configure LLMs, but researchers can also deliberately exploit multiple configuration comparisons to achieve desirable, statistically significant results.
Concerns about the reliability of hypothesis testing ~\citep{Breznau2022HypothesisUncertainty} and the risk of manufactured results~\citep{Head2015phacking,Stefan2023} are as old as scientific data analysis itself.
However, while LLM hacking is related to $p$-hacking in its susceptibility to researcher degrees of freedom~\citep{Simmons2011}, it operates at the data generation stage rather than during the analysis stage.
This creates an entirely new layer of vulnerability.

Our work demonstrates that the scientific risk of LLM hacking is not merely theoretical but a pressing challenge for reproducibility and research integrity in CSS.
Our findings demand a fundamental reconsideration of how LLMs should be integrated into scientific workflows, treating them as measurement tools that require rigorous validation~\citep{wallach2025position}.
We provide practical recommendations, which we hope will help researchers navigate these challenges and harness the transformative potential of LLMs for CSS research, without falling prey to LLM hacking.

\paragraph{Research questions.}
Through a systematic literature review and an extensive CSS replication study, we address four fundamental research questions (RQs):
\begin{itemize}
    \item \textbf{RQ1:} What are current practices for using LLMs as automated annotators in CSS research?
    \item \textbf{RQ2:} How feasible is intentional LLM hacking across different tasks?
    \item \textbf{RQ3:} What is the risk of accidental
    LLM hacking across different tasks and models?
    \item \textbf{RQ4:} To what extent can established mitigation strategies and validation practices reduce the risk of accidental LLM hacking?
\end{itemize}

\subsection{Key contributions}

We introduce the concept of \textbf{Large Language Model (LLM) hacking}~(\S\ref{sec:LargeLanguageModelHacking}).
We then quantify the risk of LLM hacking through a large-scale empirical assessment of over 13 million annotations across 37 diverse CSS tasks (see Table~\ref{tab:tasks_overview}), 18 LLMs, and 2,361 realistic hypothesis tests.
This includes tasks like stance detection or political ideology classification, with hypotheses testing group differences, such as whether Conservative manifestos express more right-wing economic positions than Labour manifestos.
Our experimental setup~(\S\ref{sec:Experiments}) is informed by a systematic literature analysis of 93 papers, which \textbf{reveals enthusiastic advocacy for LLM usage alongside insufficient validation} and limited attention to potential risks~(\S\ref{sec:Literaturereview}).

\subsection{Summary of findings}

In more detail, we report the following key findings:

\textbf{It is strikingly easy to present virtually any desired finding as statistically significant~(\S\ref{ssec:DeliberateLLMhacking}).}
By simply selecting one of the tested models with a handful of prompt paraphrases, malicious actors can arrive at any desired downstream conclusion.
Using just the models and prompts we tested, false positives are feasible for $94.4\%$ of null hypotheses, while true effects can be hidden in $98.1\%$ of cases.
Most alarmingly, statistically significant effects can be reversed entirely in $68.3\%$ of cases with true differences (Type S error).
To make matters worse, the distinction between legitimate and manipulated analyses becomes virtually undetectable post hoc.
This vulnerability means that a researcher seeking to support a predetermined conclusion is almost guaranteed to succeed while maintaining apparent scientific credibility.

Beyond intentional LLM hacking, our experiments empirically demonstrate that there is a high risk of accidental LLM hacking when following reasonable research practices without malicious intent.
\textbf{State-of-the-art (SOTA) models have on average a one-in-three chance of LLM Hacking~(\S\ref{ssec:EmpiricalLLMhackingrisk}).}
LLM hacking affects every model we test (see Table~\ref{tab:PAPER_llm_hacking_risk_by_model_task_averaged}), from small 1B parameter models with $50\%$ risk to SOTA 70B+ parameter models still exhibiting $31\%$ risk.
This means that \textbf{researchers cannot simply rely on larger, more capable models to escape the problem}.
The LLM hacking risk varies strongly across the annotation tasks we evaluated, ranging from $5\%$ for humor detection to over $65\%$ for ideology and frame classification (for some model configurations), showing that no annotation task is immune.
Type II errors dominate, with models more frequently missing true effects than fabricating false ones. Type II errors occur in $31$-$59\%$ of cases (depending on model size).
Type M risk is $41-77\%$, on average, which means that estimated effect sizes do not reflect the truth even when LLM annotations correctly identify a significant effect.

\textbf{Neither high annotation performance nor careful prompting prevent LLM hacking~(\S\ref{ssec:PredictorsofLLMhacking}).}
Through a large regression analysis of all hypothesis-model-prompt combinations (Table~\ref{tab:regression_llm_hacking}), we establish a clear hierarchy of risk factors (Table~\ref{tab:variable_importance}).
\textit{Proximity to statistical significance thresholds} emerges as the strongest predictor, with $p$ values near $0.05$ showing error rates approaching $70\%$.
\textit{Task characteristics} account for $21\%$ of the explained variance, while \textit{model performance} contributes only $8\%$.
Surprisingly, \textit{prompt engineering} choices contribute less than $1\%$ of explained variance, challenging the common belief that careful prompt design can eliminate these risks.

Surprisingly, we find \textbf{no correlation between human inter-annotator agreement and LLM hacking risk}, meaning that even for tasks where human experts perfectly agree, LLM-based annotations can yield unreliable conclusions.

\textbf{$\mathbf{100}$ human annotations outperform $\mathbf{100}$K LLM annotations for controlling false positives~(\S\ref{ssec:MitigatingLLMhackingrisk}).}
Testing 21 different approaches combining human annotation sampling strategies with statistical corrections, we discover an LLM data scale paradox:
\textit{Using human annotations alone often provides the strongest protection against Type I errors}, achieving error rates around $10\%$ with just $100$ human labels, compared to $30-40\%$ for uncorrected hybrid approaches that combine human and LLM annotations.

\textbf{Regression estimator correction techniques restore valid inference but trade Type I for Type II errors.}
Rather than solving the underlying problem, statistical correction methods like Design-based Supervised Learning~\citep{egami2023using} and Confidence-Driven Inference~\citep{gligoric-etal-2025-unconfident} face unavoidable trade-offs between error types.
While these methods successfully reduce Type I errors to nominal levels, they increase Type II errors by up to 60 percentage points.
Even with $1$,$000$ human annotations, overall LLM hacking risk only decreases to about $20\%$ when using human annotations alone, compared to $31-50\%$ baseline risk when combining human and LLM annotations without correction.

\textbf{Researchers should use LLMs wisely, not widely~(Table~\ref{tab:table1}).}
Our systematic experiments allow us to develop evidence-based guidelines for researchers navigating the trade-offs between annotation efficiency and scientific validity.
These concrete recommendations acknowledge both the appeal of LLM automation and its fundamental limitations.
For research contexts where false positives (Type I errors) pose the greatest risk, such as novel discovery claims, we show that ground-truth-only approaches provide the only reliable protection.
If, instead, Type II errors are the primary concern, or when human annotation is truly infeasible, our results suggest that researchers use specific model selection strategies and correction techniques.
We recommend transparency standards, including pre-registration of LLM configuration choices and comprehensive reporting of all tested combinations, as essential safeguards against both accidental and intentional manipulation.
\\\\
Taken together, our results show that, while LLMs offer unprecedented scalability for data annotation, their use in hypothesis testing requires fundamental changes to current practices.
The field needs to shift from treating them as convenient replacements for human annotators to recognizing them as complex instruments requiring careful calibration and validation.

\section{Large Language Model (LLM) Hacking}
\label{sec:LargeLanguageModelHacking}

We formalize \textit{LLM hacking} as a phenomenon occurring when researchers using LLMs for data annotation draw incorrect scientific conclusions.
Depending on the researcher's outcome of interest, wrong conclusions can be the false (non)discovery of an effect or a wrong statistical estimate, for example.
In this study, \textbf{we focus specifically on LLM-generated text annotations used in regression analyses}.
More precisely, the scientific outcome of interest is whether regression coefficients reach statistical significance.
However, the concept of LLM hacking is more general and encompasses any plausible scientific conclusion downstream of LLM usage.

LLM hacking is a consequence of the researchers' vast degrees of freedom in LLM configuration choices.
In the course of using LLMs to prepare data for downstream statistical analysis, researchers have many decisions to make.
Each choice is often plausible but questionable at the same time, since the downstream impact on result validity is hard to quantify.
This issue is exacerbated by the fact that many decisions are not explicitly made by models and model API services, but rather default to predetermined values that researchers are sometimes unaware of.

\begin{definition}[LLM Hacking]
Let $\theta^*$ denote the true outcome of interest, $\Phi$ the space of possible reasonable LLM configurations, and $\hat{\theta}(\phi)$ the outcome obtained using configuration $\phi \in \Phi$. LLM hacking $\mathcal{L}$ occurs when:
\begin{equation}
\mathcal{L}(\phi) = \mathbb{1}[\hat{\theta}(\phi) \neq \theta^*]
\end{equation}
\end{definition}
Here, the reference $\theta^*$ and the observed $\hat{\theta}(\phi)$ represent any potential outcome of interest, such as conclusions from statistical hypothesis tests, but we leave this broadly defined for the sake of generality.
The configuration space $\Phi$ encompasses all choices affecting LLM outputs, such as model selection, prompt formulation, temperature and other decoding parameters, and output mapping strategies (e.g., whether generated tokens are mapped to labels).
$\Phi$ is context-dependent, but must be plausible, excluding configurations that are clearly unreasonable or produce nonsensical outputs (e.g., due to completely mis-specified prompts).

\paragraph{Intentional LLM hacking.}
Even seemingly minor variations like prompt paraphrases or small temperature adjustments can produce dramatically different outcomes that propagate to downstream analyses.
When studying phenomena without true effects, LLM hacking produces Type I errors (false positives).
For real effects, it causes Type II errors (false negatives), Type S errors (wrong direction), or Type M errors (magnitude distortions)~\citep{Meehl_1967,Gelman2014Beyond}.

\begin{definition}[LLM Hacking Risk]
The LLM hacking risk quantifies the probability of obtaining false conclusions when randomly selecting from reasonable configurations:
\begin{equation}
\mathbb{E}_{\phi \sim \Phi}[\mathcal{L}(\phi)]
\end{equation}
\end{definition}

Importantly, this risk exists even for well-intentioned researchers who must necessarily make configuration choices (or are subject to undocumented default settings) without clear guidance on optimal settings.

\paragraph{Intentional LLM hacking.}
Beyond accidental errors, malicious actors could deliberately exploit the configuration flexibility to manufacture desired outcomes.
We define \textit{LLM hacking feasibility} as the existence of at least one configuration producing a specific incorrect conclusion.

\begin{definition}[LLM Hacking Feasibility]
For a specific hypothesis and desired incorrect outcome $\hat{\theta}' \neq \theta^*$, LLM hacking is feasible if:
\begin{equation}
F(\hat{\theta}') = \mathbb{1}[\exists \phi \in \Phi : \hat{\theta}(\phi) = \hat{\theta}']
\end{equation}
\end{definition}

\subsection{Relationship to $p$-hacking}

LLM hacking differs fundamentally from $p$-hacking~\citep{Simmons2011,Stefan2023}, though both produce the same outcome: false statistical conclusions arising from researcher degrees of freedom~\citep{gelman2013garden}.
The key distinction lies in their strategies.
$p$-hacking manipulates analytical choices (e.g., variable selection, outlier removal, subgroup analysis), while LLM hacking manipulates data generation through configuration choices.
Both practices can yield significant $p$ values where none should exist, but LLM hacking achieves this by shaping the annotated data itself rather than how that data is analyzed.

Risks stemming from LLM hacking and $p$-hacking are cumulative.
Hence, studies using LLM-annotated data face both configuration-induced biases and traditional analytical flexibility issues, including selective reporting~\citep{White2000snooping,head2015extent}, HARKing~\citep{kerr1998harking,Hitchcock2004}, and publication bias~\citep{sterling1959publication,rothstein2005publication,ioannidis2008most,simonsohn2014p,Goldfarb2016}.
Notice that while traditional $p$-hacking focuses predominantly on false discoveries (Type I errors), the relative importance of Type I versus Type II errors is context-dependent.
We therefore define LLM hacking to encompass both false positives and false negatives, recognizing that researchers must balance these trade-offs according to their specific research goals.

\section{Literature review: Using LLMs for automated data annotation}
\label{sec:Literaturereview}

To understand the extent of LLM use for data annotation in CSS research, we conducted a comprehensive literature review. Our systematic search across five major databases yielded 1,592 initial papers, refined to 93 relevant studies through rigorous screening (see Appendix~\ref{app:lit_review_method} for full methodology). The selected papers from 2022-2025 benchmark LLM performance or provide guidance for CSS annotation tasks.
We manually extracted relevant information, including the CSS tasks covered, LLM usage recommendations made, and validation approaches discussed (see Appendix~\ref{app:lit_review_results}).

\paragraph{The risk of LLM hacking is widespread.}

Our systematic review reveals a strong consensus toward adopting LLMs for automated data annotation across CSS.
$85$ ($91.4\%$) reviewed papers recommend using LLMs for data annotation tasks (at least under certain conditions), while only $8$ ($8.6\%$) advise against their use.
This overwhelming endorsement spans a diverse range of CSS annotation tasks, including sentiment analysis, stance detection, hate speech identification, and political discourse analysis.

The reviewed studies evaluate LLM performance across numerous CSS tasks, with $77$ ($82.8\%$) using empirical datasets to benchmark annotation quality.
Despite the limitations of using proprietary models for research~\citep{ollion2024dangers}, the GPT family of models dominates the evaluation landscape, appearing in over $80\%$ of studies, while open-weight alternatives like Llama and Qwen are tested less frequently.

\paragraph{Current validation practices for LLM annotation are insufficient.}

$48$ ($56.5\%$) papers disregard model validation (despite recommending the use of LLMs).
Papers that include validations predominantly focus on simple performance metrics.
Only four papers explicitly mention the risk of false conclusions in downstream statistical analyses when using LLM-generated annotations~\citep{egami2023using,pangakis2023automated,baly-etal-2020-detect,barrie2024replication,gligoric-etal-2025-unconfident}.
Two of those~\citep{egami2023using,gligoric-etal-2025-unconfident} propose techniques to correct regression estimates based on LLM annotation bias to reduce Type I risk.
We test both techniques in our experiments, see Section~\ref{ssec:LLMhackingriskmitigationtechniques}.

Despite the widespread adoption of LLMs in research, we still lack any principled guidance on when their use might compromise scientific validity.
We found no empirical evidence identifying which factors determine whether LLM errors will distort downstream hypothesis testing, regression analyses, or other statistical procedures commonly used in CSS research.
It thus remains unclear how well-intentioned researchers can avoid erroneous conclusions and how easily these configuration choices could be exploited to manufacture invalid results.
This omission represents a fundamental gap in our understanding of LLM reliability for scientific inference.

The validation gap in LLM usage for CSS is particularly troubling, given that these same papers advocate for replacing human annotators with automated systems whose reliability remains unverified for specific research contexts.
Despite well-established evidence that LLMs are extremely brittle annotation tools~\citep{egami2023using}, producing different outputs even with minor prompt modifications~\citep{sclar2024quantifying,salinas-morstatter-2024-butterfly,Atreja_Ashkinaze_Li_Mendelsohn_Hemphill_2025}, the prevailing sentiment appears to be that LLMs can seamlessly automate numerous annotation tasks~\citep{gilardi2023pnas,tornberg2024large}.

\section{Experimental setup}
\label{sec:Experiments}

\subsection{Data}

Table~\ref{tab:tasks_overview} lists the 37 annotation tasks that we compiled from 21 datasets into a unified format for systematic evaluation.
Starting from all 1,592 papers identified by our systematic literature review (Appendix~\ref{app:lit_review_method}), we extracted the CSS annotation tasks studied and checked for available datasets.

We sampled our final list of annotation tasks from this pool, prioritizing diversity in data types (e.g., Tweets, news articles, Reddit posts, open-ended survey responses, and party manifestos) and datasets with publicly available ground truth labels (to be able to test LLM annotation reliability) and additional metadata (to formulate realistic hypotheses).

Our final set of tasks represents a range of typical CSS annotation scenarios, including stance detection, topic classification, sentiment analysis, and many more.
These tasks span diverse domains from political science (party manifestos, election tweets) to public health (COVID-19 misinformation) and social psychology (humor detection, politeness classification). The annotation complexity ranges from straightforward binary classifications (relevant/irrelevant) to nuanced multi-class categorizations requiring domain expertise (hatespeech, issue framing).
Our selection includes datasets from early research on automated LLM annotation work~\citep{gilardi2023pnas} alongside more specialized resources such as the British Election Study open-ended responses~\citep{fieldhouse2024british}, ensuring broad coverage of contemporary CSS research applications.

\input{tables/tasks_overview.tex}

\paragraph{Data preprocessing.}
To maximize comparability across annotation tasks, we replicated the preprocessing from prior work (see Appendix~\ref{app:Dataset and Task Overview} for details), with two exceptions: we used universal deduplication (final sizes in Table~\ref{tab:datasets_overview}) and stratified random sampling (considering class and grouping variables) for datasets over $10$,$000$ instances to preserve statistical properties while keeping computational costs manageable.
Furthermore, we decompose multiclass annotation tasks into binary comparisons (e.g., class A vs.\ rest) to make hypotheses comparable across tasks in the downstream statistical analysis.
This resulted in a total of \textbf{250 binary annotation tasks}.

\paragraph{Ground truth annotations.}

If an objectively correct label is available (e.g., in the ideology\_tweets task), we use this as the ground truth.
Otherwise, we obtain ground truth labels from expert annotators (trained research assistants and domain experts) and crowdworkers.
Following prior work on benchmarking LLM annotation performance~\citep{gilardi2023pnas}, we remove all ambiguous datapoints where annotators disagree to ensure high label quality.
This filtering process results in a Krippendorff's alpha of $0.91$, on average (see Table~\ref{tab:tasks_overview}), allowing us to treat the resulting human annotations as noise-free ground truth for our experimental analysis.
Details on annotator types and label aggregation methods are provided in Table~\ref{tab:tasks_overview} and Appendix~\ref{app:Dataset and Task Overview}.

\subsection{Downstream statistical analysis}
\label{ssec:Downstream_statistical_analysis}

We use the LLM-generated annotations as inputs to downstream statistical analyses.
Their conclusions constitute our primary outcome of interest.
Consider, for example, the \textit{manifestos\_econ\_ideology} task that classifies sentences from UK party manifestos as economically left or right-wing~\citep{benoit2016crowd}.
A researcher might test whether Conservative party manifestos contain more right-wing economic positions than Labour party manifestos, assuming the null hypothesis $\mathbf{H_0}$ that there are no differences in class proportions between parties, versus the alternative $\mathbf{H_A}$ that a significant difference exists between the parties.
Similarly, for the \textit{framesI\_tweets} task from~\citet{gilardi2023pnas}, a researcher might test whether Tweets containing the keyword ``Trump'' are more likely to frame content moderation as a problem than those without this keyword.

We evaluate these hypotheses using logistic regression with binary dependent and independent variables.
The dependent variable $y$ represents either the ground truth annotation or the LLM-generated annotation, and the independent variable $x$ indicates group membership in one of two data subsets (e.g., Conservative vs.\ Labour manifestos, or texts containing ``Trump'' vs.\ not).

\paragraph{Outcome of interest.}
For each hypothesis $h$, we run two logistic regressions $\text{logit}(P(y = 1)) = \alpha + \beta x$:
\begin{itemize}
    \item \textbf{Ground truth regression}: $y^{\text{GT}}_h \sim x_h$, yielding coefficient $\beta^{\text{GT}}_h$.
    \item \textbf{LLM-informed regression} (using configuration $\phi$): $y^{\text{LLM}}_{h,\phi} \sim x_h$, yielding coefficient $\beta^{\text{LLM}}_{h,\phi}$
\end{itemize}
We test for each coefficient whether it reaches statistical significance at $\alpha = 0.05$, effectively asking: ``\textit{Does the proportion of positive class labels differ significantly between the two groups?}''\footnote{This could also be answered with a two-proportion z-test. We verified that both methods yield identical results.}
Let $S^{\text{GT}}_h$ and $S^{\text{LLM}}_{h,\phi}$ denote the significance indicators for the coefficients $\beta^{\text{GT}}_h$ and $\beta^{\text{LLM}}_{h,\phi}$, and let $\text{sgn}(\cdot)$ denote the sign function.

Replacing the ground truth with LLM annotations can lead to several discrepancies between the two sets of conclusions~\citep{Meehl_1967,Gelman2014Beyond}:
\begin{itemize}
\setlength{\itemsep}{0pt}\setlength{\parskip}{0pt}
\item \textbf{Type I error}: detecting a difference with LLMs when the ground truth test did not ($S^{\text{GT}}_h = 0$ and $S^{\text{LLM}}_{h,\phi} = 1$)
\item \textbf{Type II error}: missing a difference with LLMs when the ground truth test did ($S^{\text{GT}}_h = 1$ and $S^{\text{LLM}}_{h,\phi} = 0$)
\item \textbf{Type S error}: both tests are significant but the signs differ ($S^{\text{GT}}_h = S^{\text{LLM}}_{h,\phi} = 1$ but $\text{sgn}(\beta^{\text{GT}}_h) \neq \text{sgn}(\beta^{\text{LLM}}_{h,\phi})$)
\item \textbf{Type M error}: both tests are significant, signs agree, but estimated magnitudes differ ($S^{\text{GT}}_h = S^{\text{LLM}}_{h,\phi} = 1$ and $\text{sgn}(\beta^{\text{GT}}_h) = \text{sgn}(\beta^{\text{LLM}}_{h,\phi})$ but $\beta^{\text{GT}}_h \neq \beta^{\text{LLM}}_{h,\phi}$)
\end{itemize}
We emphasize that $\beta^{\text{GT}}_h$ is itself an estimate subject to sampling variability, not the true population parameter.
In our analysis, we use the significance decision based on ground truth labels as a reference point for assessing whether substituting LLM-based annotations would alter the substantive conclusions a researcher might draw.
Although differences in significance levels are not themselves necessarily statistically significant~\citep{Gelman2006Significant}, this operationalization of LLM hacking is directly relevant for CSS, where regression significance remains a central reporting convention.
See Appendix~\ref{app:outcome_of_interest} for robustness checks and an extended discussion of the subject.

\paragraph{Hypotheses.}
We evaluate multiple hypotheses for each annotation task.
We generate these hypotheses through different data groupings, i.e., criteria that split a dataset into two parts.
We consider 
1) keyword-based splits (e.g., texts containing ``economy'' vs.\ not) and 
2) splits based on original metadata (e.g., male vs.\ female authors).
This process yields 2,361 hypotheses across all tasks, of which 48.1\% are based on original metadata.
See Appendix~\ref{app:Dataset_groupings} for more details.

\subsection{LLM configuration space}

We rely on models and prompts used in prior work, wherever possible, to ensure a plausible configuration space $\Phi$.

\paragraph{Models.}
We evaluate 18 models from 4 model providers representing the current state of the art. For each model family, we test different scales: 
\begin{enumerate}
    \item Llama 3 family (1B-70B)~\citep{grattafiori2024llama},
    \item Qwen 2.5~\citep{qwen2025qwen25technicalreport} and 3~\citep{qwen2025qwen3technicalreport} families (1.5B-72B), 
    \item Gemma family (1B-27B)~\citep{team2025gemma}, and
    \item GPT-4o variants~\citep{openai2024gpt4ocard}.
\end{enumerate}
We use the instruction-tuned versions for all models, but omit ``-instruct'' suffices for brevity.
We test both open-weight and proprietary models to evaluate the full spectrum of LLM capabilities typically used in CSS research.
All models and exact versions used are listed in Appendix~\ref{app:Models}.

\paragraph{Prompts, decoding, and output mapping.}
We formulate new prompts based on the original data annotation guidelines when prior prompts are unavailable.
We create few-shot prompt versions when these guidelines include specific examples.
To test sensitivity to prompt formulation, we generated additional semantically equivalent prompt paraphrases to have at least five plausible prompt versions per task (details in Appendix~\ref{app:Prompt paraphrases}).
Table~\ref{tab:tasks_overview} shows that this process yielded 5-7 prompts per task.
The full set contains 199 prompts total.
72 ($36.2\%$) are our generated paraphrases, 178 ($89.4\%$) are zero-shot prompts, and 21 ($10.6\%$) are few-shot prompts.
We set the model temperature to 0 for reproducibility and the maximum number of generated tokens to $20$.
We match generated tokens to class labels using regular expressions to ensure valid category selection.
We account for cases where models fail to follow instructions or produce invalid outputs (see Table~\ref{app:fraction_na}), finding that larger models demonstrate substantially better instruction adherence.

\subsection{Baselines}
\label{ssec:baselines}

We consider the following three random baselines:
\begin{itemize}
\item \textbf{Baseline 1 (random conclusions)}: Randomly assigns statistical conclusions ($\{S=0, S=1 \land \beta>0, S=1 \land \beta<0\}$) with 50\% probability of finding no difference ($S=0$), representing pure chance in hypothesis testing.
\item \textbf{Baseline 2 (random labels)}: Assigns annotation labels uniformly at random from the set of valid classes, simulating completely uninformative annotations.
\item \textbf{Baseline 3 (random errors)}: Generates annotations with controlled F1 scores by flipping random labels to incorrect classes, allowing us to distinguish LLM behavior from random noise.
\end{itemize}
We compute baseline risks using $1$,$000$ bootstrap samples for baselines 1 and 2, and $100$ samples for each F1 score level in baseline 3.

\subsection{Metrics}
\label{ssec:metrics}
We use weighted F1 as a performance metric throughout.
If not mentioned otherwise, we average metrics across annotation tasks $T$ with hypotheses $H_t$ for each task $t$ and for a set of LLM configuration choices $\Phi$ (consisting of 18 models and up to seven prompts per task).\footnote{
While some configurations may be more likely to be chosen in practice, we use uniform averaging to avoid making assumptions about researcher preferences.
Our configuration space $\Phi$ is already restricted to plausible choices by including only well-performing models across different scales and cost tiers, and by using reasonable prompt variations.}
Let $H_t^0 = \{h \in H_t : S^{\text{GT}}_h = 0\}$ denote hypotheses where ground truth is non-significant, and $H_t^1 = \{h \in H_t : S^{\text{GT}}_h = 1\}$ denote hypotheses where ground truth is significant.
This gives us the following empirical risk measures:

The \textbf{Type I risk quantifies the false positive rate}: the probability that LLM annotations detect a significant effect when none exists in the ground truth, i.e., $1-\text{specificity}$.
\begin{equation}
\text{Type I Risk} = \frac{1}{|T|} \sum_{t \in T} \frac{1}{|H_t^0|} \sum_{h \in H_t^0} \frac{1}{|\Phi|} \sum_{\phi \in \Phi} \mathbb{1}[S^{\text{LLM}}_{h,\phi} = 1]
\end{equation}
The \textbf{Type II risk quantifies the false negative rate}: the probability that LLM annotations fail to detect true effects present in the ground truth data, i.e., $1-\text{sensitivity}$.
\label{eq:TypeIRisk}
\begin{equation}
\text{Type II Risk} = \frac{1}{|T|} \sum_{t \in T} \frac{1}{|H_t^1|} \sum_{h \in H_t^1} \frac{1}{|\Phi|} \sum_{\phi \in \Phi} \mathbb{1}[S^{\text{LLM}}_{h,\phi} = 0]
\label{eq:TypeIIRisk}
\end{equation}
Notice that the Type II risk is sometimes denoted by $\beta$ in hypothesis testing, and $1 - \beta$ corresponds to the power of the test.

The \textbf{Type S risk captures the sign error rate}: among true effects, the probability that LLM annotations detect a significant effect in the opposite direction. This is a particularly concerning form of error that fundamentally mischaracterizes the relationship between variables.
\begin{equation}
\text{Type S Risk} = \frac{1}{|T|} \sum_{t \in T} \frac{1}{|H_t^1|} \sum_{h \in H_t^1} \frac{1}{|\Phi|} \sum_{\phi \in \Phi} \mathbb{1}[S^{\text{LLM}}_{h,\phi} = 1, \text{sgn}(\beta^{\text{GT}}_h) \neq \text{sgn}(\beta^{\text{LLM}}_{h,\phi})]
\label{eq:TypeSRisk}
\end{equation}
Finally, the \textbf{empirical LLM hacking risk measures the overall probability of reaching incorrect statistical conclusions}:
\begin{equation}
\text{LLM Hacking Risk} = \frac{\text{Type I Risk} + \text{Type II Risk} + \text{Type S Risk}}{2}
\label{eq:LLMHackingRisk}
\end{equation}
This measure averages the risk of errors involving null hypotheses (Type I) with the risk of errors involving alternative hypotheses (Type II and S), providing a balanced assessment of the overall risk of reaching incorrect statistical conclusions when using LLM annotations.
Note that some fields use the convention of a Type II risk of $0.2$ (power of $80\%$).
However, there are no formal standards for power, and the trade-off between Type I and Type II risk heavily depends on the application context, which is why we use a balanced measure here.

Even in the absence of LLM hacking, effect sizes of the downstream conclusion may still be over- or underestimated.
To empirically quantify Type M errors, we calculate the ratio of effect sizes, measured as differences in group proportions, for cases where both ground truth and LLM annotations yield significant results with matching signs.
Let $\Delta p^{\text{GT}}_h$ and $\Delta p^{\text{LLM}}_{h,\phi}$ denote the differences in class proportions between groups for ground truth and LLM annotations, respectively.
We define:
\begin{equation}
\text{Type M Risk} = \frac{1}{|T|} \sum_{t \in T} \frac{\sum_{h \in H_t^1} \sum_{\phi \in \Phi} \left| \frac{|\Delta p^{\text{LLM}}_{h,\phi}|}{|\Delta p^{\text{GT}}_h|} - 1 \right| \cdot \mathbb{1}[S^{\text{LLM}}_{h,\phi} = 1, \text{sgn}(\beta^{\text{GT}}_h) = \text{sgn}(\beta^{\text{LLM}}_{h,\phi})]}{\sum_{h \in H_t^1} \sum_{\phi \in \Phi} \mathbb{1}[S^{\text{LLM}}_{h,\phi} = 1, \text{sgn}(\beta^{\text{GT}}_h) = \text{sgn}(\beta^{\text{LLM}}_{h,\phi})]}
\end{equation}
The \textbf{Type M risk quantifies the average relative magnitude error for correctly identified effects}.
A value of 0 indicates perfect magnitude estimation, while a value of $0.5$ means that LLM-based effect size estimates deviate from true effect sizes by an average of $50\%$.

\subsubsection{Feasibility of intentional LLM hacking}

The \textbf{LLM hacking feasibility rates quantify how often it is possible to obtain a wrong scientific conclusion through intentional LLM hacking}.
Empirically, it is measured as the proportion of hypotheses for which at least one configuration exists that results in a Type I, Type II, or Type S error. More precisely:
\begin{equation}
\text{Type I Error Feasibility Rate} = \frac{1}{|T|} \sum_{t \in T} \frac{1}{|H_t^0|} \sum_{h \in H_t^0} \mathbb{1}[\exists \phi \in \Phi : S^{\text{LLM}}_{h,\phi} = 1]
\end{equation}
\begin{equation}
\text{Type II Error Feasibility Rate} = \frac{1}{|T|} \sum_{t \in T} \frac{1}{|H_t^1|} \sum_{h \in H_t^1} \mathbb{1}[\exists \phi \in \Phi : S^{\text{LLM}}_{h,\phi} = 0]
\end{equation}
\begin{equation}
\text{Type S Error Feasibility Rate} = \frac{1}{|T|} \sum_{t \in T} \frac{1}{|H_t^1|} \sum_{h \in H_t^1} \mathbb{1}[\exists \phi \in \Phi : S^{\text{LLM}}_{h,\phi} = 1, \text{sgn}(\beta^{\text{GT}}_h) \neq \text{sgn}(\beta^{\text{LLM}}_{h,\phi})]
\end{equation}
Here, the indicator functions denote the LLM hacking feasibility for hypothesis $h$.
These rates quantify the potential for deliberate manipulation while maintaining apparent methodological credibility, particularly when constraining $\Phi$ to methodologically defensible choices (e.g., using only top-performing models).

Conversely, we measure the feasibility of correct scientific conclusions as the proportion of hypotheses for which at least one configuration exists that yields the correct outcome.
We measure this for null and alternative hypotheses separately:
\begin{equation}
\text{H}_0 \; \text{Correctness Feasibility Rate} = \frac{1}{|T|} \sum_{t \in T} \frac{1}{|H_t^0|} \sum_{h \in H_t^0} \mathbb{1}[\exists \phi \in \Phi : S^{\text{LLM}}_{h,\phi} = 0]
\end{equation}
\begin{equation}
\text{H}_\text{A} \; \text{Correctness Feasibility Rate} = \frac{1}{|T|} \sum_{t \in T} \frac{1}{|H_t^1|} \sum_{h \in H_t^1} \mathbb{1}[\exists \phi \in \Phi : S^{\text{LLM}}_{h,\phi} = 1, \text{sgn}(\beta^{\text{GT}}_h) = \text{sgn}(\beta^{\text{LLM}}_{h,\phi})]
\end{equation}
A low correctness feasibility rate would suggest that LLMs may not be suitable for the given annotation task.

\subsection{Mitigation techniques using human-annotated samples}
\label{ssec:LLMhackingriskmitigationtechniques}

When researchers have access to a limited number of human expert annotations ($n_{\text{human}}$), these ground truth (GT) samples can become a valuable resource to mitigate LLM hacking risk.
Table~\ref{tab:mitigation} summarizes our mitigation strategies (M1-M9) along three dimensions: 
(1) how datapoints are sampled to be annotated by humans, 
(2) how the collected human annotations are used for downstream statistical analyses, and 
(3) how the LLM is selected (provided that LLM annotations are used for estimation).

For each dimension, we test three strategies we identified in the literature.
This gives an extensive testbed of 21 alternative techniques to mitigate LLM hacking.
In the following, we describe each of the approaches across these three dimensions in more detail.

\begin{table}[h]
\centering
\small
\caption{LLM hacking mitigation strategies (abbreviated with M1-M9) using human-annotated samples. The third dimension (how the model is selected) only applies to the strategies using LLM annotations, i.e., M2, M3, M5, M6, M8, and M9.}
\label{tab:mitigation}
\begin{tabular}{clccc}
\toprule
& & \multicolumn{3}{c}{\textbf{Data Usage Strategy} (2)} \\
\cmidrule(lr){3-5}
& & \textbf{GT Only} & \textbf{GT + LLM} & \textbf{GT + LLM + Correction} \\
\cmidrule(lr){2-5}
\multirow{3}{*}{\rotatebox{90}{\parbox{1.3cm}{\centering\textbf{Sampling\\Strategy} (1)}}} & \multicolumn{1}{|l|}{\textbf{Random}} & M1 & M2 & M3 (DSL) \\
& \multicolumn{1}{|l|}{\textbf{Low Confidence}} & M4 & M5 & M6 (DSL) \\
& \multicolumn{1}{|l|}{\textbf{Active}} & M7 & M8 & M9 (CDI) \\
\bottomrule
\multicolumn{3}{c}{} & \multicolumn{2}{c}{$\underbrace{\hspace{6cm}}_{\text{\small \textbf{Model Selection Strategy} (3)}}$} \\
\multicolumn{3}{c}{} & \multicolumn{2}{c}{\small Random, GPT-4o, Best-performing} \\
\end{tabular}
\end{table}

\subsubsection{Sampling strategies for human annotations}

\paragraph{Random sampling.} Selects ground truth samples uniformly at random with probability $\frac{n_{\text{human}}}{n}$ for each instance.
This is the sampling strategy used by~\citet{egami2023using}.

\paragraph{Low confidence sampling.}

Several papers mention that researchers may over-sample data that is difficult to annotate automatically, for example, by increasing the sampling probability for texts that LLMs are more uncertain about~\citep{li-etal-2023-coannotating,egami2024using}.
We implement this strategy by sampling the $n_{\text{human}}$ instances with the lowest LLM prediction confidence.
Following~\citet{lin2022teaching},~\citet{li-etal-2023-coannotating}, and~\citet{gligoric-etal-2025-unconfident}, we consider verbalized model confidence probabilities as a proxy for the model's expected performance.
This approach is promising since LLMs are mostly able to provide information about their uncertainty themselves~\citep{wang-etal-2021-want-reduce,kadavath2022language}.
See Appendix~\ref{app:VerbalizedLowconfidencesampling} for more details on the verbalized confidence score elicitation.

\paragraph{Active sampling.}

Active learning has emerged as a technique to strategically annotate the most informative samples~\citep{van-der-meer-etal-2024-annotator,BOSLEY_KUZUSHIMA_ENAMORADO_SHIRAITO_2025}.
Active learning typically uses a supervised ML model to iteratively select the next batch of texts to be annotated.
In Natural Language Processing, active learning has a long history of improving data annotation by sampling datapoints that are either difficult or most representative of the full data~\citep{margatina-etal-2021-active,zhang-etal-2022-survey}.

We use the implementation provided by~\citet{gligoric-etal-2025-unconfident}, which uses an XGBoost model that predicts LLM annotation errors from its annotations and verbalized confidence scores.
Data is then sampled proportionally to the predicted error, mixed with 10\% uniform sampling for increased stability.
This approach shares the same objective as low confidence sampling but is more sophisticated, as it learns the relationship between verbalized confidence and actual annotation error rather than simply using verbalized confidence as a proxy.
See Appendix~\ref{app:Activesamplingimplementationdetails} for more implementation details.

\subsubsection{Using human annotations for downstream statistical analyses}

\paragraph{Ground truth only (Table~\ref{tab:mitigation} cells M1, M4, and M7).}

Using only the small human-annotated sample instead of any LLM annotations is the simplest approach.
This provides unbiased estimates but with high variance due to the limited sample size.
\citet{egami2023using} and \citet{gligoric-etal-2025-unconfident} call this ``Gold-Standard Only'' and ``Human only'', respectively.

\paragraph{Ground truth + LLM annotations (M2, M5, M8).}
This second approach treats sampled ground truth labels as perfect replacements for LLM annotations on those instances, using LLM annotations for all other instances.
This approach reduces the variance but potentially introduces bias from incorrect LLM predictions.

\paragraph{Ground truth + LLM + Corrected estimator (M3, M6, M9).}
Finally, we combine human and LLM annotations using bias-corrected estimators that account for LLM annotation errors while leveraging the full dataset.
We consider the following estimator correction techniques:
\begin{itemize}
    \item \textbf{Design-based Supervised Learning (DSL).}
    \citet{egami2023using} introduced DSL to correct downstream estimators for LLM prediction bias.
    DSL uses a doubly-robust procedure that combines LLM predictions with human annotations to create bias-corrected pseudo-outcomes. 
    It maintains valid statistical inference even when LLM predictions are arbitrarily biased, requiring only that the sampling probability for human annotations is known and bounded away from zero.
    See Appendix~\ref{app:dsl_details} for implementation details.
    \item \textbf{Confidence-Driven Inference (CDI).} Introduced by~\citet{gligoric-etal-2025-unconfident}, CDI is built on top of the active sampling strategy described above.
    To produce an unbiased downstream estimate, CDI takes the active sampling probabilities into account and combines human annotations (where available) with LLM annotations.
    The final regression estimate additionally uses a learned tuning parameter that controls the trust placed in LLM annotations to minimize the estimator variance while maintaining unbiasedness.
    See Appendix~\ref{app:cdi_details} for implementation details.
\end{itemize}
DSL and CDI are both optimized for producing confidence intervals that cover the true effect.
Thus, they essentially reduce the risk for Type I errors. However, Type II errors are not explicitly controlled for.

\subsubsection{Model selection for automated data annotation}

When LLM annotations are incorporated into downstream analyses (M2, M3, M5, M6, M8, M9), researchers must decide which model to use.
This choice can significantly impact the quality and reliability of downstream results.

The predominant model validation approach identified in the literature involves ranking candidate models based on their performance on a small subset of available ground truth data, then selecting the best-performing model for all subsequent annotations~\citep{Linegar2023,hussain2024tutorial,alizadeh2025open}.
To evaluate the efficacy of this strategy, we compare it against two simple baselines (random and GPT-4o model selection) that do not require any ground truth data:
\begin{itemize}
    \item \textbf{Random}: As a baseline, we randomly select one model from the set of available LLMs with equal probability. This strategy serves as a lower bound for model selection performance and requires no ground truth data.
    \item \textbf{GPT-4o}: Researchers often default to the best-performing/most well-known model within their budget.
    We implement this heuristic by consistently selecting GPT-4o, which represents one of the most widely used models
    at the time of this study.
    \item \textbf{Best-performing}: When ground truth annotations are available, we can select LLMs based on their performance on the annotated subset. This data-driven approach uses the human-annotated sample as a test set to identify the most suitable model for the specific task at hand.
\end{itemize}

\section{Results}

\begin{figure*}[thb]
    \centering
    \includegraphics[width=0.9\textwidth]{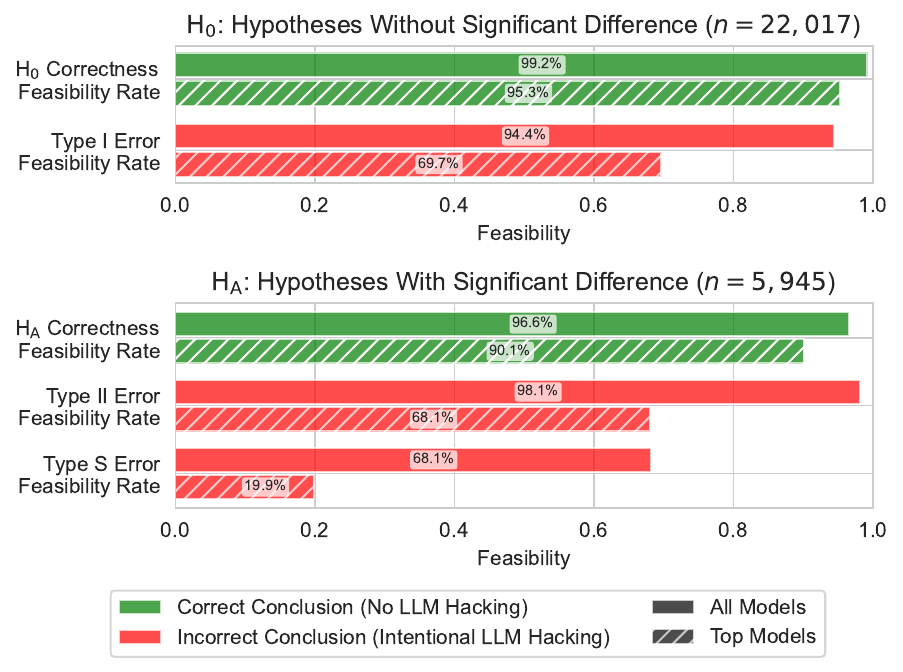}
    \caption{
    Average feasibility rates of LLM hacking (red bars) and correct conclusions (green bars) across annotation tasks. Red bars show the proportion of hypotheses where at least one configuration yields an incorrect conclusion (lower is better, indicating more robust results). Green bars show the proportion where at least one configuration yields the correct conclusion (higher is better, indicating greater potential for accurate conclusions).
    The top panel shows feasibility rates for hypotheses without significant differences, while the bottom panel shows feasibility rates for hypotheses with significant differences.
    Analysis restricted to top models includes:
    Llama-3.1-70B,
    Qwen2.5-32B,
    Qwen2.5-72B,
    Qwen3-32B,
    Gemma-3-27b,
    GPT-4o-mini, and
    GPT-4o.
    }
    \label{fig:llm_hacking_feasibility_task_weighted}
\end{figure*}

\begin{figure*}[tb]
    \centering
    \includegraphics[width=0.9\textwidth]{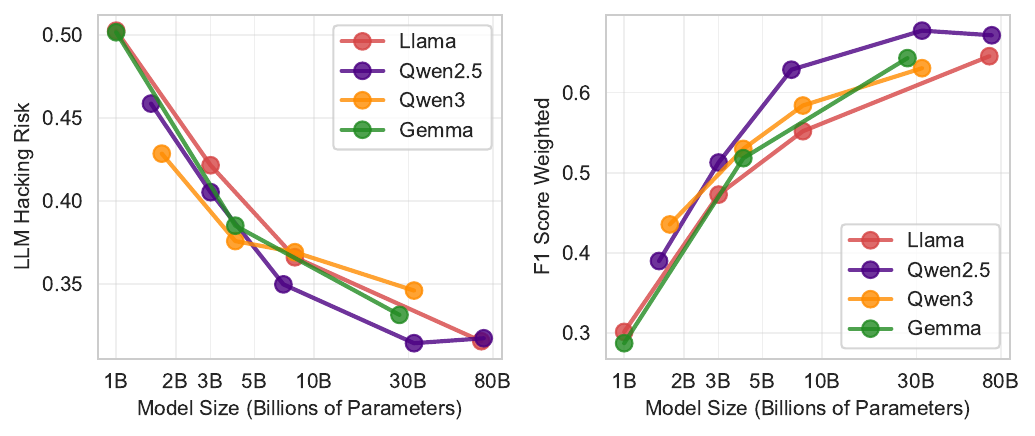}
    \caption{Scaling relationships for LLM hacking risk and annotation performance. Left panel shows task-averaged LLM hacking risk decreasing with model size across all model families, with larger models consistently outperforming smaller ones. Right panel shows corresponding improvements in weighted F1 annotation performance. Both metrics exhibit clear scaling trends, though substantial risk remains even for the largest models.}
    \label{fig:PAPER_scaling_laws}
\end{figure*}

\begin{table}[tb]
\input{tables/PAPER_llm_hacking_risk_by_model_task_averaged.tex}
\label{tab:PAPER_llm_hacking_risk_by_model_task_averaged}
\end{table}

\begin{figure*}[tb]
    \centering
    \includegraphics[width=0.9\textwidth]{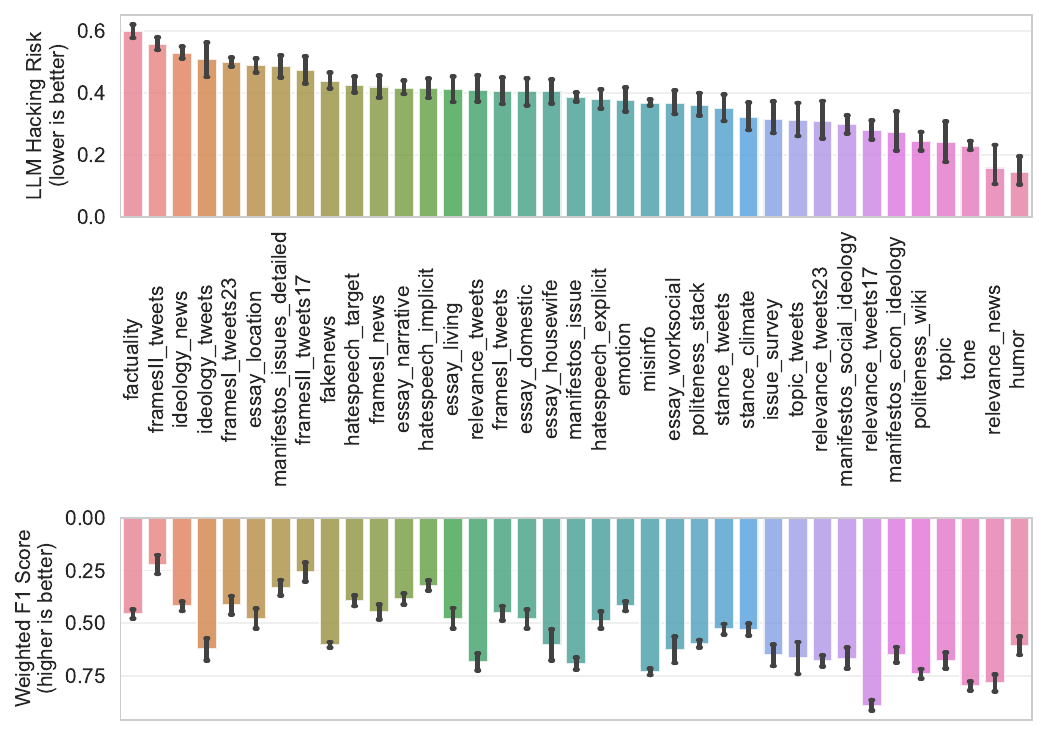}
    \caption{Average weighted F1 scores and LLM hacking risk across all 37 annotation tasks (sorted by decreasing LLM hacking risk). Error bars show 95\% confidence intervals.
    }
    \label{fig:PAPER_performance_f1_score_weighted_by_task}
\end{figure*}

\subsection{Intentional LLM hacking}
\label{ssec:DeliberateLLMhacking}

\textbf{Deliberate model selection and prompt formulation can make almost anything be presented as statistically significant}.
Even when restricting analyses to top-performing models -- which a reviewer might consider methodologically sound -- the intentional hacking feasibility remains alarmingly high.

Figure~\ref{fig:llm_hacking_feasibility_task_weighted} shows that for hypotheses without true differences, a significant effect can be manufactured (Type I error) in $94.4\%$ of cases by using at least one model-prompt combination we tested.
Similarly, fabricating a Type II error (i.e., hiding a true effect) is feasible in $98.1\%$ of the cases with true differences.
Even more surprisingly, Type S errors, i.e., finding significant effects in the opposite direction, are still feasible in $68.3\%$ of cases with true effects, enabling the complete reversal of scientific conclusions while maintaining an appearance of methodological rigor.
Feasibility rates are consistently high across all annotation tasks (see Figure~\ref{fig:llm_hacking_feasibility_by_task} in the Appendix).

The fact that both correct conclusions (feasible in $99.2\%$ and $96.2\%$ of cases) and incorrect conclusions (feasible in $94.4\%$ and $98.1\%$ of cases) can be achieved through different configurations underscores the fundamental unreliability of LLMs as data annotators.
This near-universal feasibility means researchers can cherry-pick configurations to support virtually any desired outcome.
Note also that our high reported hacking rates are conservative estimates, as we tested only a limited set of models and prompts.
A malicious actor testing additional configurations could push these success rates even closer to $100\%$.

\textbf{Even when considering only the seven best-performing models, Type I and Type II error feasibility rates remain very high ($69.7\%$ and $68.1\%$, respectively)}.
This rate suggests that the vulnerability to manipulation is not merely a consequence of poor-quality models but rather a fundamental characteristic of LLM-based annotation approaches, where even methodologically defensible choices can produce incorrect outcomes.

In general, \textbf{constraining the configuration space $\Phi$ or requiring replicability across multiple independent configurations reduces LLM hacking feasibility}.
We list several such combinations in Table~\ref{tab:llm_hacking_feasibility_restrictions} in Appendix~\ref{app:feasibility}.
Pre-registration of all LLM configuration choices, combined with reporting results from all pre-registered configurations, thus makes intentional LLM hacking more difficult by eliminating post-hoc selection opportunities and by demanding consistency of findings across diverse setups.

\begin{practicalbox}
Document all tested model-prompt combinations, not just the final choice.
As a reviewer, be suspicious of studies reporting results from a single LLM configuration without justification.
Pre-register all LLM configuration choices, including model selection procedures, prompt formulations, and decoding parameters, to make this a standard practice in the field.
\end{practicalbox}

\subsection{Empirical LLM hacking risk}
\label{ssec:EmpiricalLLMhackingrisk}

\begin{figure*}[thb]
    \centering
    \includegraphics[width=\textwidth]{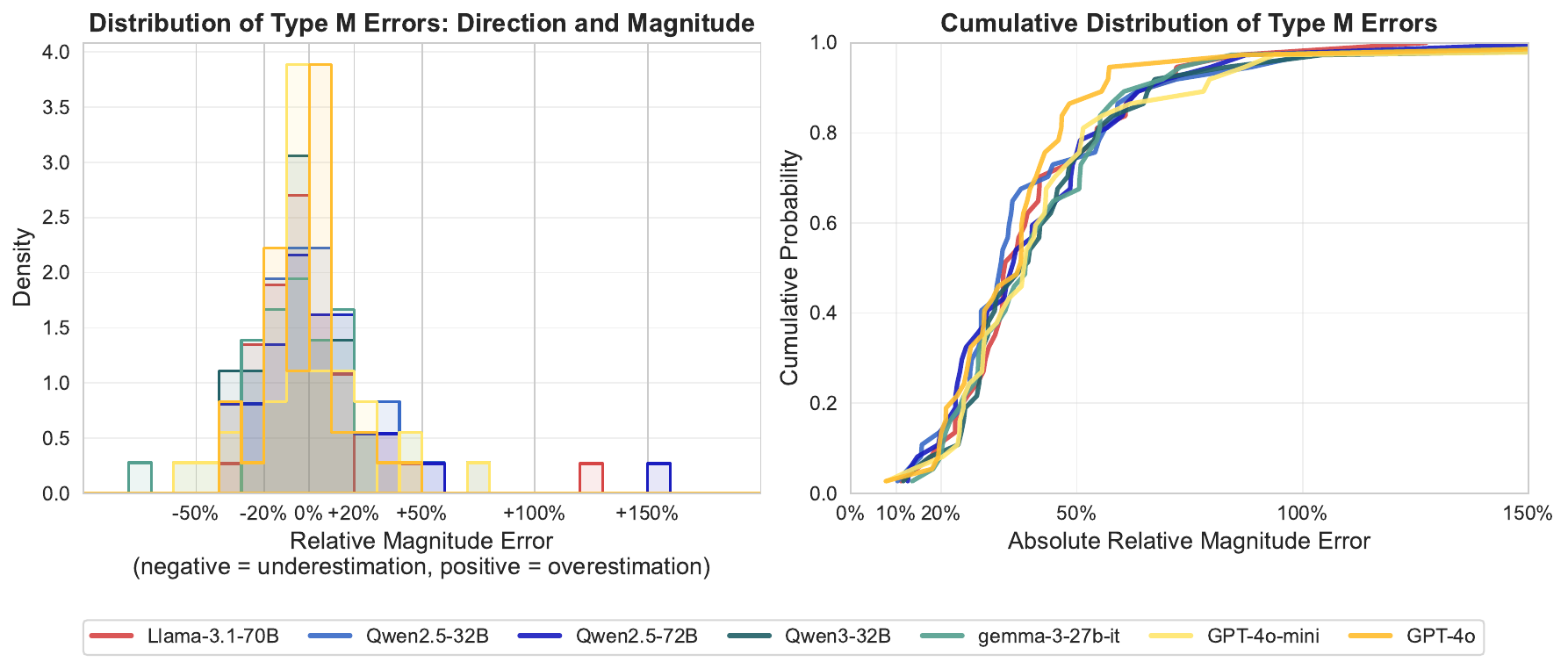}
    \caption{Type M error analysis for seven top models: Relative magnitude errors indicate to what extent estimated effect sizes derived from LLM annotations deviate from true effect sizes in the absence of LLM hacking.
    Left: The histogram visualizes the distribution of Type M errors.
    Right: The Cumulative probability shows the chance of landing below a certain error level, as indicated on the x-axis.}
    \label{fig:type_m_error_distributions}
\end{figure*}

\begin{figure*}[thb]
    \centering
    \includegraphics[width=0.7\textwidth]{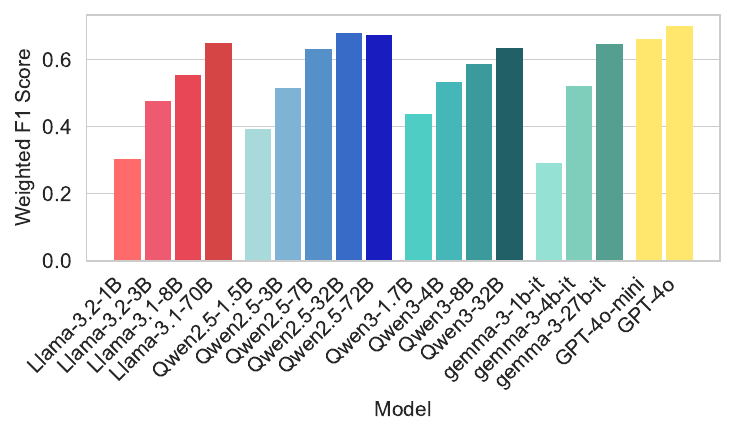}
    \caption{Average weighted F1 scores by model across all 37 annotation tasks.}
    \label{fig:performance_f1_score_weighted_by_model}
\end{figure*}

\textbf{Even SOTA LLMs produce incorrect scientific conclusions in a substantial fraction of cases}.
Table~\ref{tab:PAPER_llm_hacking_risk_by_model_task_averaged} and Figure~\ref{fig:PAPER_scaling_laws} show that empirical LLM hacking risk ranges from $31\%$ for the best-performing 70B parameter models to $50\%$ for 1B parameter models.
LLM hacking risk varies across annotation tasks, from as low as $5\%$ for humor detection to over $65\%$ for ideology, factuality, and frame classification tasks (factuality, ideology\_tweets, framesII\_tweets, and manifestos\_issues\_detailed), for some models (see Figure~\ref{fig:PAPER_performance_f1_score_weighted_by_task} and Table~\ref{app:llm_hacking_risk_by_task_model} in Appendix~\ref{tab:PAPER_llm_hacking_risk_by_task_model}).

\textbf{Unlike random noise that fails to detect effects, LLMs produce plausible-seeming annotations that yield unreliable statistical conclusions in unpredictable ways}.
Table~\ref{tab:PAPER_llm_hacking_risk_by_model_task_averaged} shows that baseline 1 (random conclusions) yields 62.3\% LLM hacking risk, while baseline 2 (random labels) produces 52.5\% risk.
Random labels have extremely low Type I risk (7.8\%) but extremely Type II risk (92.9\%).
The lack of signal in random annotations makes spurious discoveries unlikely but renders true effect detection nearly impossible.

\textbf{In $\mathbf{4}$-$\mathbf{16\%}$ of cases, LLM annotations result in a completely reverse effect}, making something that increases look like it decreases, as shown in the Type S risk column in Table~\ref{tab:PAPER_llm_hacking_risk_by_model_task_averaged}.
\textbf{Even when LLM annotations allow for the correct identification of a significant effect, the estimated effect sizes are heavily biased}.
The Type M risk reveals that LLM-based effect size estimates deviate from true effect sizes by $41$-$77\%$ on average across models.
This systematic inaccuracy makes precise effect size estimation particularly challenging with LLM annotations.
The right panel of Figure~\ref{fig:type_m_error_distributions} illustrates this limitation: even for the best-performing model (GPT-4o), only $2.7\%$ of correctly identified significant effects have magnitude estimates within $10\%$ of the true effect size.

\textbf{Type I risk is substantial, though Type II errors dominate the error landscape across all model families}.
So on average, LLMs more often fail to detect true differences than fabricate false ones.
While larger models within each family consistently show lower risk, the improvements follow a pattern of diminishing returns.
Figure~\ref{fig:PAPER_scaling_laws} illustrates this scaling relationship. Going from a Llama model with 8B to 70B parameters reduces the risk by only $5$ percentage points (from $36.6\%$ to $31.6\%$), whereas going from 1B to 8B parameters yields $13.6$ percentage point improvement (from $50.3\%$ to $36.6\%$).
Similarly, going from Qwen2.5 with 7B to 72B parameters moderately improves the risk  ($35.0\%$ to $31.8\%$ risk), while scaling smaller models shows larger improvements.
This correlation suggests fundamental limitations that additional scaling alone may not overcome, particularly beyond 30-70B parameters.
Figures~\ref{fig:PAPER_scaling_laws} (right panel) and~\ref{fig:performance_f1_score_weighted_by_model} show similar patterns for annotation performance scaling.

\textbf{Performance metrics tell only part of the story}.
Figure~\ref{fig:PAPER_performance_f1_score_weighted_by_task} shows a large variation in task difficulty, with weighted F1 scores ranging from about $0.2$
to above $0.8$ for some tasks.
Yet even high performance does not prevent LLM hacking, as several tasks with F1 scores exceeding $0.93$ still exhibit hacking risks above $50\%$ (see Figure~\ref{fig:PAPER_llm_hacking_vs_f1_score_weighted_subplots_by_task}).
This disconnect between annotation performance and downstream validity underscores that \textbf{traditional performance metrics inadequately capture the risks to scientific inference}.

\textbf{LLMs deviate from expected error patterns of random noise}.
The green lines in Figure~\ref{fig:PAPER_llm_hacking_vs_f1_score_weighted_subplots_by_task} show baseline 3 results.
For several models and tasks, LLMs perform worse than this random error baseline at equivalent F1 scores, on average.
This indicates that LLM errors may not be uniformly distributed but rather systematically biased in ways that particularly harm downstream statistical inference.

\begin{practicalbox}
Use the most capable available models for annotation tasks (where model size is a proxy for capability).
Models with 70B+ parameters show approximately $31\%$ LLM hacking risk compared to about $50\%$ for 1-2B parameter models.
However, expect diminishing returns with model scale.
\end{practicalbox}

\subsection{Predictors of LLM hacking}
\label{ssec:PredictorsofLLMhacking}

If we understand what drives LLM hacking risk, we can devise targeted mitigation strategies, informing adequate research practice. To this end, we run an OLS regression analysis (Table~\ref{tab:regression_llm_hacking}) to identify key predictors.
In particular, we fit the following OLS regression model:
\begin{align}
\text{LLM Hacking} &\sim \text{weighted F1 score} \times (\text{Task} + \text{Model}) \nonumber \\
&\quad + \text{significant difference found} \nonumber \\
&\quad + \text{normalized distance from significance threshold} \nonumber \\
&\quad + \text{prompt type} + \text{prompt detail}
\end{align}
We include fixed effects to control for baseline differences across tasks and models.
The inclusion of interaction terms allows the influence of model performance (F1 score) to vary across tasks and models, while the significance-related predictors capture whether a statistical difference was detected and how far results lie from the $p = 0.05$ threshold.

Table~\ref{tab:variable_importance} reveals a clear hierarchy in the predictors of LLM hacking risk.
The dominant factor is whether a significant difference was found in the original analysis, accounting for $56.6\%$ of the explained variance.
Furthermore, the normalized distance from the significance threshold adds another $10.2\%$.
This shows that \textbf{LLM hacking is more likely to occur near decision boundaries, which is when results hover close to significance thresholds}.
Task-specific characteristics contribute the second-largest share at $20.8\%$, indicating that certain annotation tasks are inherently more vulnerable to configuration-dependent outcomes.
Tasks differ not only in their average LLM hacking risk and annotation performance (Figure~\ref{fig:PAPER_performance_f1_score_weighted_by_task}), but also in how strongly performance improvements reduce hacking risk (Figure~\ref{fig:PAPER_llm_hacking_vs_f1_score_weighted_subplots_by_task}).

Model performance (F1 score) explains only $7.7\%$ of the variance.
All else equal, prompt engineering choices alone -- often emphasized in practice -- cannot prevent LLM hacking (prioritizing few- over zero-shot prompts: $0.1\%$, and providing detailed over concise prompt formulations: $0.01\%$).
This distribution challenges conventional wisdom about LLM reliability, suggesting that even high-performing models and careful prompt engineering cannot eliminate hacking risk, especially when results lie near statistical boundaries.

In the following subsections, we examine each predictor to understand the mechanisms underlying LLM hacking vulnerability.

\begin{table}[thb]
\centering
\caption{Variable Importance in Predicting LLM Hacking}
\label{tab:variable_importance}
\begin{adjustbox}{max width=\textwidth}
\begin{tabular}{lc}
\toprule
Predictor & Relative Importance (in \%) \\
\midrule
Significant Difference Found & 56.6\% \\
Task (Fixed Effects) & 20.8\% \\
Normalized Distance from Significance Threshold & 10.2\% \\
Weighted F1 score & 7.7\% \\
Model (Fixed Effects) & 2.5\% \\
Weighted F1 score $\times$ Task & 1.9\% \\
Weighted F1 score $\times$ Model & 0.1\% \\
Shot Type & 0.1\% \\
Prompt Detail Level & 0.01\% \\
\midrule
\multicolumn{2}{l}{\textit{Note:} Relative importance calculated using the LMG method by~\citep{lindeman1980introduction} with the \texttt{relaimpo} package~\citep{relaimpo}.} \\
\multicolumn{2}{l}{Values represent the proportion of total explained variance (R$^2$ = $0.154$) attributable to each predictor} \\
\multicolumn{2}{l}{after accounting for correlations among predictors.} \\
\bottomrule \\
\end{tabular}
\end{adjustbox}
\end{table}

\subsubsection{Model capability and performance}

\begin{figure*}[htbp]
    \centering
    \vspace{-1cm}
    \includegraphics[width=0.8\textwidth]{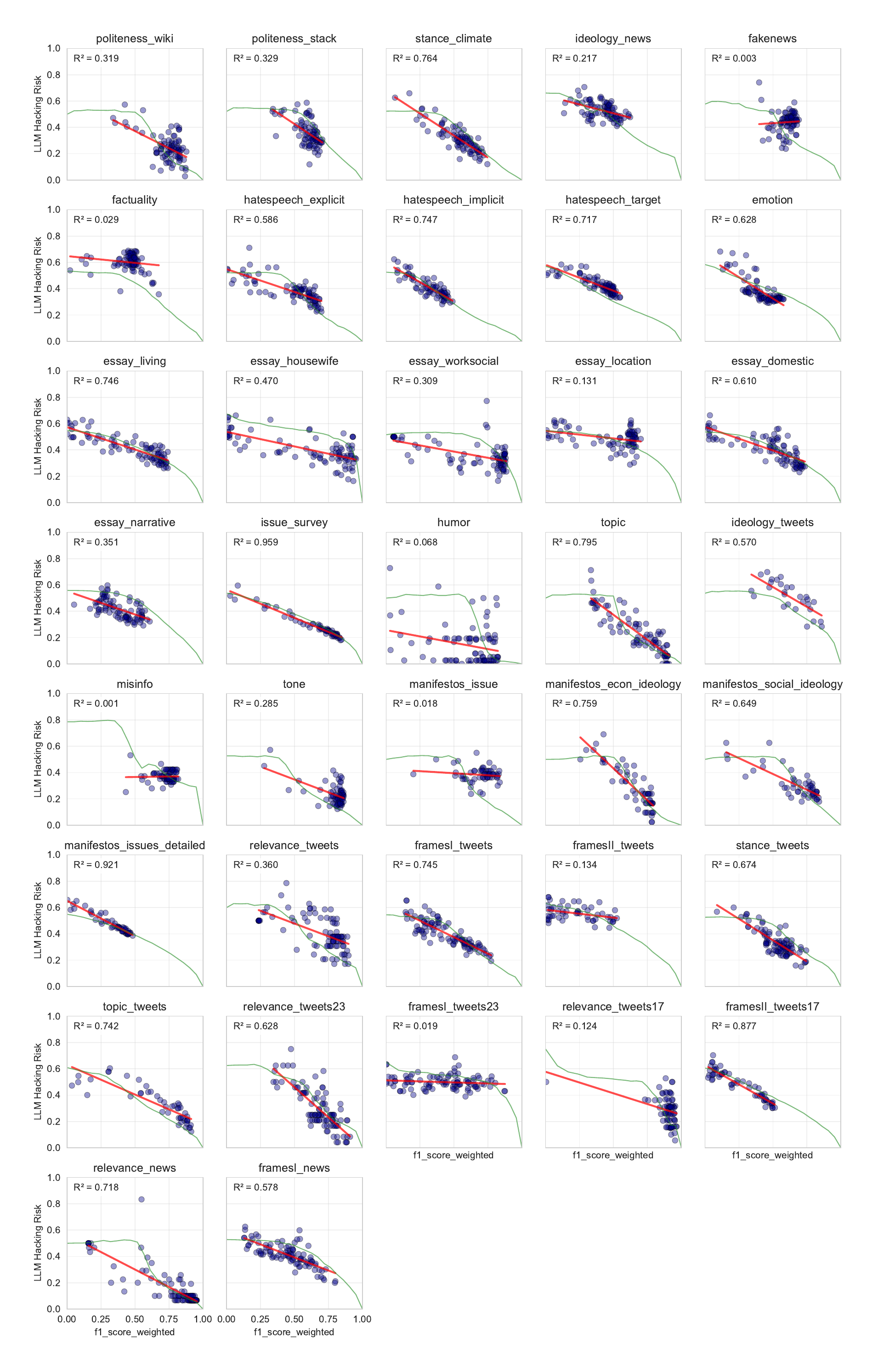}
    \caption{LLM hacking risk versus model performance by task. Each dot represents a model-prompt combination, where LLM hacking risk is calculated across all hypotheses within that task. Red lines represent linear regression fits.
    The green line indicates baseline 3 results.
    }
    \label{fig:PAPER_llm_hacking_vs_f1_score_weighted_subplots_by_task}
\end{figure*}

\textbf{Higher annotation performance strongly predicts lower LLM hacking risk, in general}.
The negative slopes in Figure~\ref{fig:PAPER_llm_hacking_vs_f1_score_weighted_subplots_by_task} display this relationship across all tasks, with correlations ranging from $-0.02$ (essentially no relationship) to $-0.46$ (strong negative correlation).
We list exact correlation scores in Table~\ref{tab:llm_hacking_f1_score_weighted_correlations} in the Appendix.
The variation in correlation strength across tasks suggests that for some annotation scenarios, improving performance directly translates to more reliable downstream inferences.
However, even highly accurate annotations can lead to false conclusions.
For example, we find instances with $93\%$ weighted F1 score
and $50\%$ LLM hacking risk in the \textit{essay\_housewife} task.
The \textit{relevance\_tweets17} task even shows a $96\%$ weighted F1 score and an LLM hacking risk of $16\%$.

General model capability, measured with scores from the popular MMLU-PRO benchmark~\citep{MMLU_PRO_NEURIPS2024}, shows a consistent negative relationship with hacking risk (see  Figure~\ref{fig:llm_hacking_vs_mmlu} in Appendix~\ref{ssec:LLMannotationreliability}).
Models scoring above $50\%$ on MMLU-PRO exhibit lower LLM hacking risk ($30\%$, on average), while those below $20\%$ show risks approaching $50\%$.
This relationship holds across model families, suggesting that broad capabilities transfer to more reliable annotation behavior even for specialized tasks.
This is in line with research suggesting selecting LLMs based on public benchmarks~\citep{Linegar2023,hussain2024tutorial}.
However, minor prompt reformulations can drastically change the outputs, which results in a large model performance variance across prompt paraphrases (see Appendix~\ref{ssec:LLMannotationreliability} for more details).

\begin{practicalbox}
Use models and prompts with high annotation performance for hypothesis testing.
When F1 scores fall below $0.5$, LLM hacking risk typically exceeds $40\%$.
However, be aware that high performance alone does not guarantee the correctness of downstream conclusions.
\end{practicalbox}

\begin{figure*}[thb]
    \centering
    \begin{subfigure}[b]{0.495\textwidth}
        \centering
        \includegraphics[width=\textwidth]{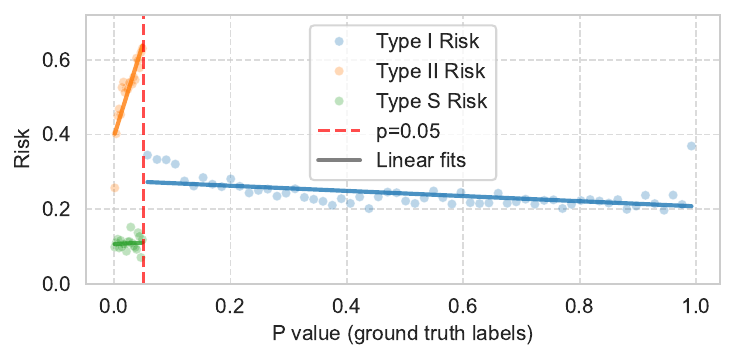}
        \caption{Risk by $p$ values derived from ground truth annotations.}
        \label{fig:PAPER_llm_hacking_risk_by_p_value_gt}
    \end{subfigure}
    \hfill
    \begin{subfigure}[b]{0.495\textwidth}
        \centering
        \includegraphics[width=\textwidth]{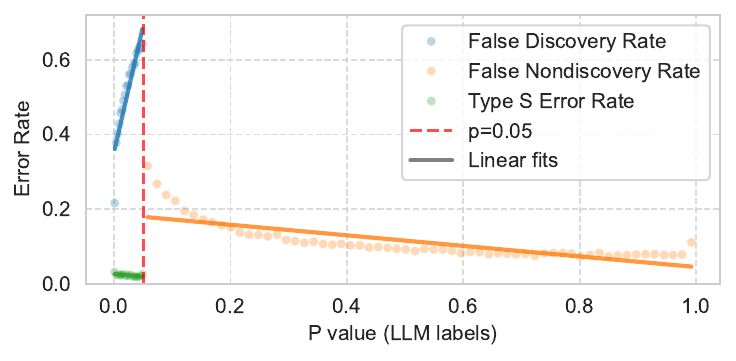}
        \caption{Error rates by $p$ values derived from LLM annotation.}
        \label{fig:PAPER_llm_hacking_risk_by_p_value_llm}
    \end{subfigure}
    \caption{LLM hacking by $p$ values. Both panels show elevated LLM hacking near the significance threshold ($p=0.05$).}
    \label{fig:PAPER_llm_hacking_risk_by_p_value}
\end{figure*}

\textbf{Results near conventional significance thresholds prove extraordinarily unreliable when using LLM-generated annotations}. Figure~\ref{fig:PAPER_llm_hacking_risk_by_p_value} reveals how proximity to the $p = 0.05$ threshold creates fundamental instability in research conclusions.

The left panel of Figure~\ref{fig:PAPER_llm_hacking_risk_by_p_value_gt} shows error risks conditional on ground truth $p$ values. This knowledge is of course only available in controlled experiments where the true labels are known.
We observe spikes in Type I risk (false positives) as ground truth $p$ values approach $0.05$, reaching close to $70\%$ Type I risk probability.
Type II and Type S error risks are also elevated near the significance threshold.
In other words: when the true effect is borderline, LLM annotations will likely yield contradicting conclusions.
What is more concerning, though, is that Type I, Type II, and Type S Risks all remain high even as $p$ values are distant from $0.05$.
Thus, \textbf{LLM annotation errors can obscure or reverse even robust phenomena}; that is, even if true effects are strong, LLM annotations may still produce wrong downstream conclusions.
These findings suggest that the unreliability stems not merely from noise around decision boundaries but from systematic LLM limitations.

The right panel (Figure~\ref{fig:PAPER_llm_hacking_risk_by_p_value_llm}) presents the more practically relevant scenario: error rates as a function of $p$ values derived from the LLM annotations themselves.
Since researchers using LLM annotations typically lack access to ground truth, this represents the information actually available when making research decisions.
Here we use discovery rates rather than risk metrics, since a condition on ground truth outcomes (as we do in Equations~\eqref{eq:TypeIRisk}-\eqref{eq:LLMHackingRisk}) does not make sense when also conditioning on LLM-derived $p$ values (x-axis in the right panel of Figure~\ref{fig:PAPER_llm_hacking_risk_by_p_value_llm}).
We provide exact metric definitions in Appendix~\ref{app:AdditionalMetrics}.
Intuitively, the False Discovery Rate quantifies the fraction of false positives among all discoveries.
And the False Nondiscovery Rate quantifies the fraction of false negatives among all non-discoveries.

Both False Discovery and False Nondiscovery rates surge dramatically for LLM-derived $p$ values around $0.05$.
This shows that \textbf{LLM annotations that produce borderline significant results are systematically unreliable}.
The critical insight is that while ground truth $p$ values remain unknown in practice, researchers can observe the $p$ values resulting from their LLM-based analyses as a form of reliability.
In other words, the proximity to decision boundaries serves as an observable warning signal in practice.
The persistence of elevated error rates even for strong effects ($p$ values near 0 and 1) reveals a fundamental limitation of LLM-based annotation.
Unlike random measurement error, which primarily affects borderline cases, \textbf{even highly significant effects can be false positives when relying on annotations produced by LLMs}.

\begin{practicalbox}
Exercise extreme caution when interpreting any LLM-based statistical results, not just borderline cases.
Our findings reveal that:
\begin{itemize}
    \item Even when LLM-derived $p$ values are far from $0.05$, error rates are substantial.
    \item Proximity to significance thresholds serves as a warning signal, with False (Non)Discovery Rates peaking near $p = 0.05$.
    \item LLM annotation errors can fundamentally reverse conclusions even for robust phenomena.
\end{itemize}
Therefore, never rely on a single LLM configuration for hypothesis testing, but always conduct sensitivity analyses across multiple models and prompts. Furthermore, treat all LLM-derived statistical conclusions as preliminary, requiring validation through alternative methods or data sources.
Last but not least, report the full distribution of findings (including $p$ values and effect sizes) across different LLM configurations rather than a single point estimate.
\end{practicalbox}

\subsubsection{Prompt engineering effects}

Table~\ref{tab:regression_llm_hacking} shows that zero-shot prompting increases risk relative to few-shot prompting, while brief prompts underperform detailed ones.
Though the variable importance analysis (Table~\ref{tab:variable_importance}) shows that prompt engineering influence is modest compared to model and task characteristics, accounting for less than $1\%$ of explained variance.
However, these are not causal effects, and prompt detail effects are likely hidden in performance as a predictor.
Despite limited overall impact, all else equal, prompt engineering remains one of the few factors directly under researcher control.
A good prompt can increase performance scores and thereby reduce LLM hacking risk.
However, researchers should not expect prompt optimization alone to be a strong countermeasure to LLM hacking risk.

\begin{practicalbox}
Prefer few-shot over zero-shot prompting and use detailed task descriptions over brief instructions.
While prompt engineering provides limited protection against LLM hacking per se, testing various different prompt formulations can help improve robustness.
\end{practicalbox}

\subsubsection{Task characteristics and human annotator agreement}
\label{sssec:human-IAA-vs-LLMhacking}

\begin{figure*}[htb]
    \centering
    \includegraphics[width=0.9\textwidth]{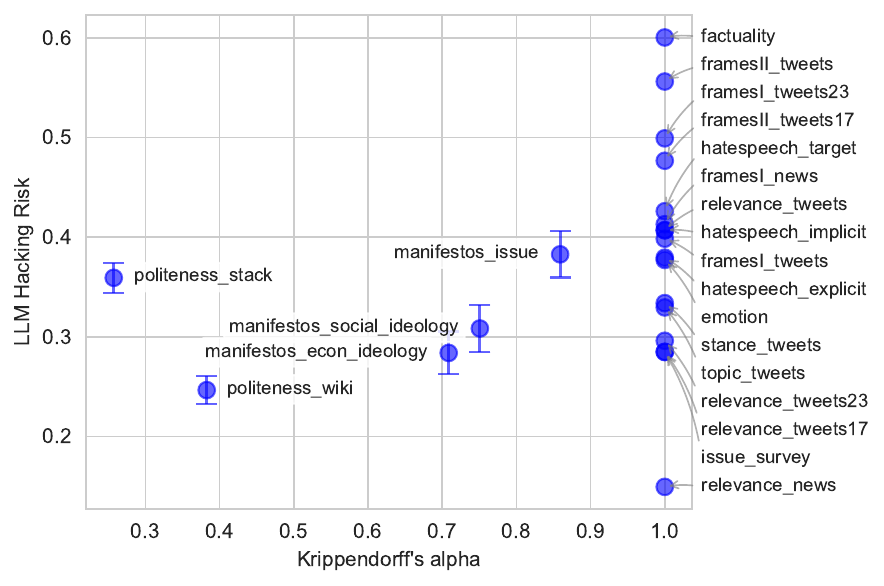}
    \caption{LLM hacking risk versus inter-annotator agreement measured as the Krippendorff's alpha of the human annotations used as the ground truth.
    Krippendorff's alpha can only be calculated for tasks with more than one available annotation per datapoint (see Table~\ref{tab:tasks_overview} for an overview and Appendix~\ref{app:Dataset and Task Overview} for more details on the exact agreement calculations).
    Each point represents a task with error bars indicating 95\% CI (we omit error bars for $K.\alpha=1$ for readability).
    }
    \label{fig:_PAPER_llm_hacking_by_Krippendorffs_alpha}
\end{figure*}

We do not find a significant correlation of LLM hacking risk with human inter-annotator agreement, as visualized in Figure~\ref{fig:_PAPER_llm_hacking_by_Krippendorffs_alpha} ($p>0.1$).
We measured human agreement using Krippendorff's alpha, which allows for different data types and annotation scales across all annotation tasks~\citep{krippendorff2018content}.
This independence suggests that LLM errors follow different patterns than human disagreement.
This means that \textbf{even tasks with perfect human agreement can exhibit high LLM hacking risk, while tasks with human disagreement sometimes show lower risk}.
However, notice that low agreement may stem from multiple factors, such as task difficulty, subjectivity, instruction quality, and annotator selection biases.
Future work should disentangle how task characteristics affect LLM hacking risk.

To further investigate LLM-human alignment, we examine the relationship between LLM hacking risk and metrics from the Alternative Annotator Test (alt-test)~\citep{calderon-etal-2025-alternative}, which statistically evaluates whether LLMs can replace individual human annotators.
We find that high performance on individual annotation alignment (i.e., passing the alt-test) does not guarantee reliability for downstream inference (see Appendix~\ref{app:alt_test} for detailed analysis).

The absence of correlation with human agreement carries important implications for research practice.
Researchers cannot use high human agreement or strong individual annotator alignment as justification for fully automated annotation without validation.
Even seemingly objective tasks with near-perfect human consensus can yield unreliable downstream inferences when annotated by LLMs.

\begin{practicalbox}
Human annotations and high human annotator agreement are important aspects of the general data annotation pipeline.
However, do not rely on human annotator agreement to determine whether LLM annotation is appropriate.
High human agreement does not predict low LLM hacking risk.
Validate LLM performance on your specific task and downstream analysis, regardless of human agreement rates.
\end{practicalbox}

\subsection{Mitigating unintentional LLM hacking risk}
\label{ssec:MitigatingLLMhackingrisk}

\textbf{Access to even small numbers of human annotations enables multiple mitigation strategies, though each involves fundamental trade-offs}.
Figure~\ref{fig:type_i_vs_type_ii_s_tradeoff} shows that no single strategy dominates across different error types.
The most sophisticated statistical corrections can perform worse on average than simpler alternatives.
Our findings suggest that human expert annotation plays a crucial role in mitigating LLM hacking risks.

Of the nine LLM hacking mitigation techniques described in Section~\ref{ssec:LLMhackingriskmitigationtechniques}, we here only focus on M1, M2, M3, and M9.
Other techniques (M4, M5, M6, M7, M8) only differ in the sampling method used for the expert annotations and are shown in the full Figure~\ref{fig:llm_hacking_mitigation_techniques_ALL} in the appendix.
Differences across sampling methods are minor compared to the ground truth data usage approach we analyze and the fundamental trade-offs between error types we observe.
To account for different annotation budgets, we consider $n_{\text{human}} = \{25, 50, 100, 250, 500, 1000\}$ known ground truth annotations.
To ensure comparability across $n_{\text{human}}$ values, we only include tasks where a specific number of known annotations amounts to at most 70\% of the entire dataset.
We show more detailed results in Appendix~\ref{app:MitigatingLLMhackingrisk}.

\subsubsection{Uncorrected LLM-informed regressions provide invalid inference}

Without bias correction techniques, approaches that use LLM annotations only or that simply combine GT samples with LLM annotations (M2) yield Type I risks of 30-38\% (see Figure~\ref{fig:type_i_vs_type_ii_s_tradeoff}), far exceeding the nominal 5\% significance level.
Hence, uncorrected hybrid approaches are unsuitable for rigorous scientific inference.
In contrast, \textbf{statistical correction methods restore valid inference}.
CDI (M9) successfully controls Type I error to match the nominal significance level with as few as $100$ human annotations.
DSL (M3) also reduces Type I errors significantly.

\begin{figure*}[htb]
    \centering
    \includegraphics[width=0.9\textwidth]{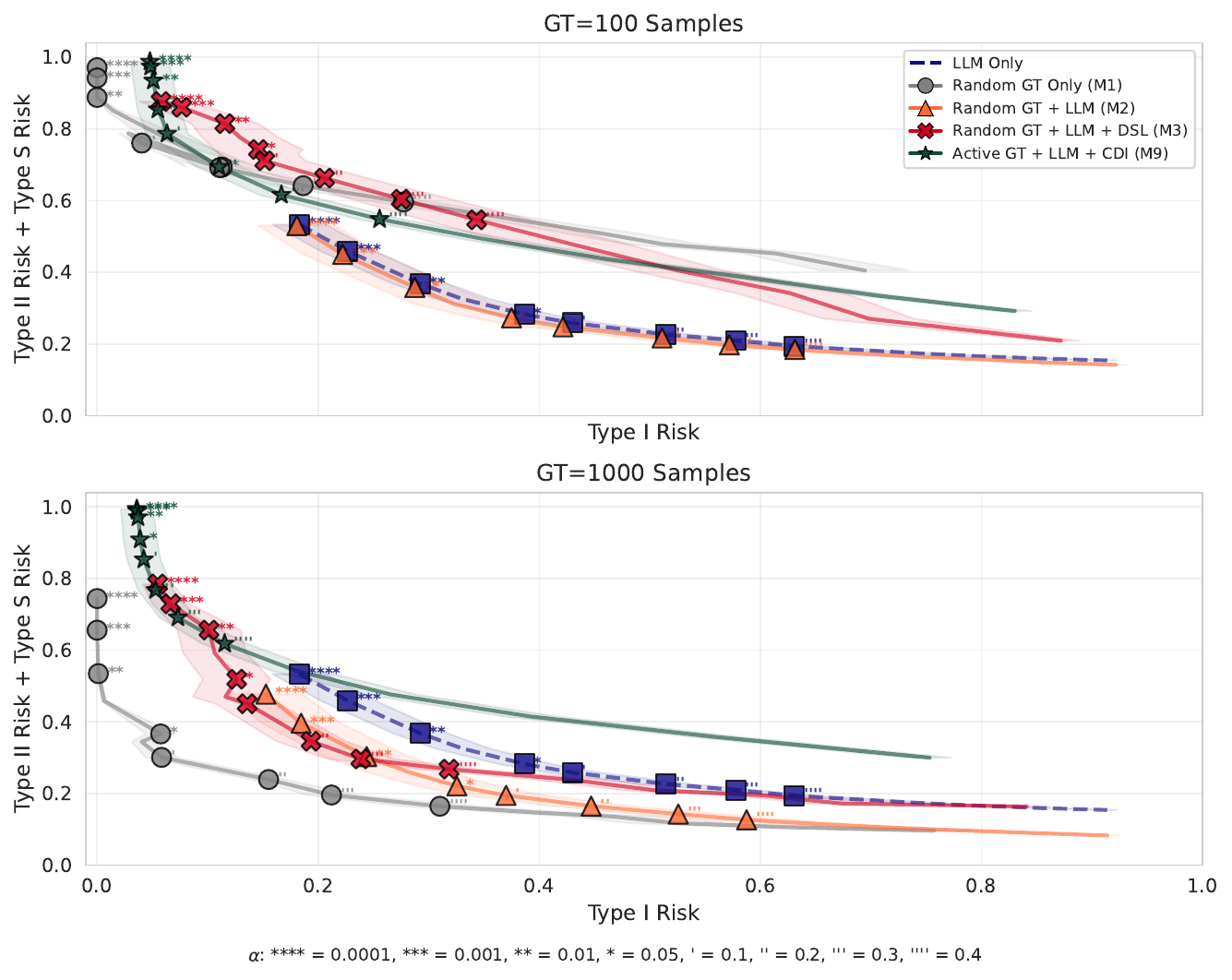}
    \caption{Type I versus Type II+S risk trade-off curves across significance thresholds. Each curve represents performance as the significance threshold $\alpha$ varies from 0.0001 to 0.4, with markers indicating specific $\alpha$ values. Shaded regions indicate 95\% confidence bands.}
    \label{fig:type_i_vs_type_ii_s_tradeoff_siglevels}
\end{figure*}

For $1000$ GT samples, the \textbf{ground truth only approach from the Pareto frontier for Type I vs.\ Type II+S error trade-offs}.
Figure~\ref{fig:type_i_vs_type_ii_s_tradeoff_siglevels} shows that across different significance thresholds, using human annotations alone (M1, gray line) consistently achieves the optimal balance.
That is, for any given Type II+S risk level, GT only achieves the lowest Type I risk, and vice versa.

\begin{practicalbox}
Collect as many high-quality expert annotations as feasible.
Using CDI-corrected estimates or randomly sampled human annotations alone provides strong protection against Type I errors.
\end{practicalbox}

\subsubsection{LLM bias correction methods create error trade-offs}

\begin{figure*}[htb]
    \centering
    \includegraphics[width=0.9\textwidth]{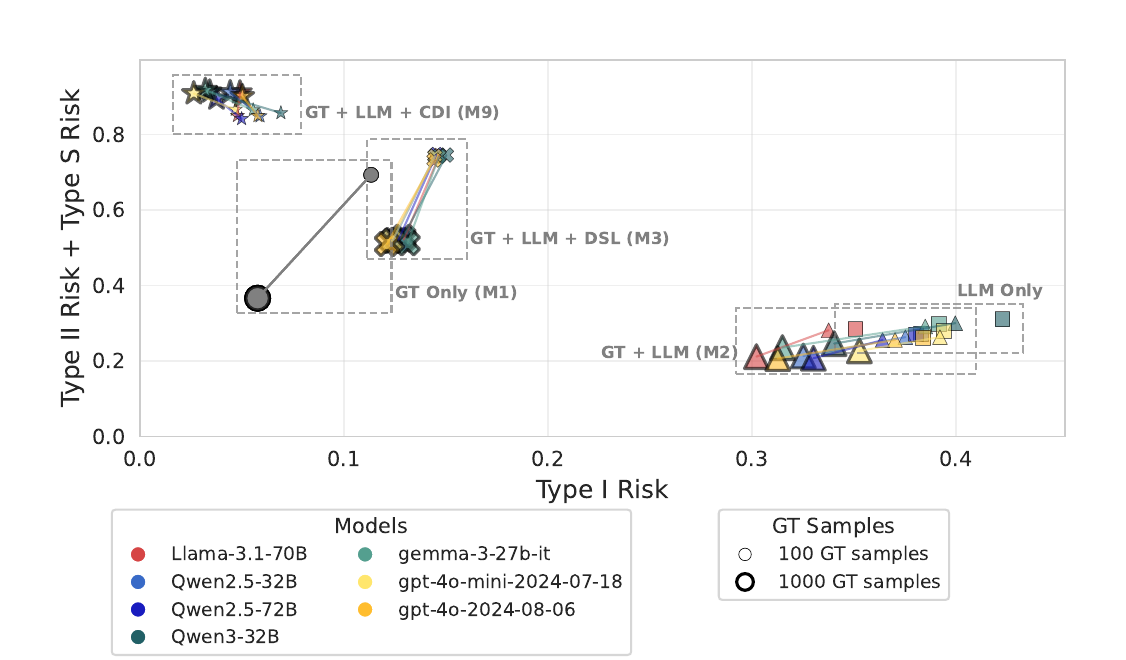}
    \caption{Trade-off between Type I errors and combined Type II+S errors across mitigation strategies.
    Marker types represent different mitigation techniques with varying numbers of ground truth (GT) samples ($100$ or $1$,$000$).
    The origin $(0,0)$ represents the ideal scenario with no statistical errors, but in practice, balancing the trade-off between Type I and Type II+S errors depends on the specific application context.
    }
    \label{fig:type_i_vs_type_ii_s_tradeoff}
\end{figure*}

\begin{figure*}[htb]
    \centering
    \includegraphics[width=\textwidth]{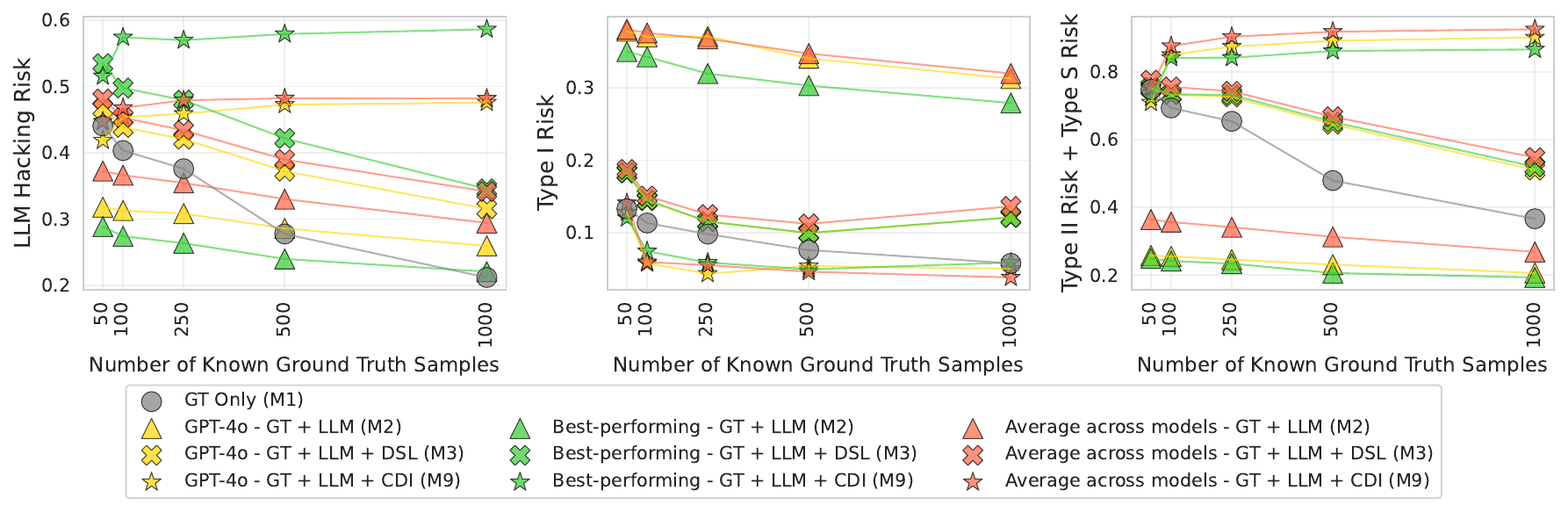}
    \caption{Model selection strategy comparison across ground truth sample sizes and mitigation techniques.}
    \label{fig:model_selection_comparison}
\end{figure*}

Statistical correction techniques combine the efficiency of LLM annotation with the reliability of human labels.
However, this Type I error control comes with an increased Type II risk, which is up to 60 percentage points higher than naive combinations of LLM and GT annotations, as visualized in Figure~\ref{fig:type_i_vs_type_ii_s_tradeoff}.
Notice that this reflects an inherent statistical trade-off rather than a methodological failure.
Design-based Supervised Learning (M3) and confidence-Driven Inference (M9) successfully reduce Type I errors compared to a naive combination of human and LLM annotations (M2), achieving rates comparable to ground-truth-only approaches (M1).
\textbf{Relying only on a moderate sample of human annotations provides the best of both worlds, with low Type I and moderate Type II risk.}

\begin{practicalbox}
Choose bias correction methods based on research priorities.
Use collected ground truth samples only or CDI-corrected LLM annotations when Type I errors are most concerning (e.g., novel discovery claims).
Use a combination of GT samples + LLM annotations (M2) when Type II errors are most problematic (e.g., replication studies).
Understand that all corrections involve trade-offs.
\end{practicalbox}

\subsubsection{Human annotations are crucial}

The baseline risk without any mitigation ranges from about $31$-$50\%$ across models, while solely relying on $1$,$000$ human annotations reduces this to about $20\%$.
Surprisingly, abandoning the use of LLMs and just collecting human samples turns out to be the most effective mitigation strategy, outperforming even sophisticated estimator correction techniques, on average.
Figure~\ref{fig:model_selection_comparison} shows that increasing human annotations consistently reduces LLM hacking risk across all mitigation techniques.
However, if researchers can only collect very few human samples, additional annotations from strong LLMs can reduce risk.

If LLM annotations are used, \textbf{model selection based on performance with human-annotated subsamples provides additional protection against LLM hacking}.
For our 18 models tested, selecting the best performer for each hypothesis consistently reduces risk by about $4$ percentage points compared to using GPT-4o for all tasks (see green lines in Figure~\ref{fig:model_selection_comparison}).
In this case, GPT-4o is selected as optimal in $49\%$ of cases, making a default selection of GPT-4o a strong baseline (see Figure~\ref{fig:model_selection_best_models} in Appendix~\ref{app:model_selection_best_models}).

\begin{practicalbox}
Prioritize human annotation collection over LLM sophistication.
Even $100$ human annotations outperform LLM-based approach in terms of Type I risk.
1,000 human annotations outperform any LLM-based approach in terms of LLM hacking risk generally, on average.
When relying on LLM annotations (e.g., because Type II risk is relevant or because even moderate-scale human annotation is truly infeasible), use human-annotated validation sets to select the best-performing model for your specific task.
When using regression estimate correction techniques like CDI or DSL, it is crucial to consider their inherent Type I vs.\ Type II error trade-off.
\end{practicalbox}

\section{Limitations}

Our LLM hacking risk quantification assumes noise-free ground truth annotations and effective significance thresholds ($p = 0.05$) for identifying true effects.
While we reduce noise by filtering datapoints with annotator disagreement (resulting in an average Krippendorff's alpha of $0.91$ across tasks, see $K.\alpha$ in Table~\ref{tab:tasks_overview}), real human annotations contain their own biases and errors, which means that our empirical LLM hacking risk quantification may over- or underestimate real risk levels.
However, detailed annotator bias analyses for our datasets are unavailable, making this limitation difficult to address quantitatively.

LLM hacking feasibility rates remain unaffected by whether the ground truth regressions represent true population effects.
Even if true effects differ from our estimates, the ability to achieve different downstream conclusions through configuration choices remains.
This fundamental vulnerability exists regardless of annotation quality.
However, our empirical risk estimates compare conclusions between ground truth and LLM-informed regressions, and may over- or underestimate risk with regard to true population effects. 
We present extensive robustness analyses in Appendix~\ref{app:outcome_of_interest}, which show that our risk estimates remain largely stable when accounting for ground truth uncertainty.

Our risk estimates depend on the configuration space considered.
Misspecified prompts could inflate LLM hacking risk.
We ensure plausibility of configurations by replicating models and prompts from prior research, and by excluding prompts that do not yield valid LLM outputs.
Our extensive analyses across models and performance levels show that LLM hacking risk persists even with model scaling and high annotation performance.
The problem is not whether risk exists, but that we find no evidence that it will disappear.

\section{Discussion}

LLMs are rapidly becoming standard tools for data annotation across scientific disciplines.
Researchers advocate for their cost-effectiveness and claim superior performance to crowdworkers~\citep{gilardi2023pnas,tornberg2024large}.
Our findings challenge enthusiasm for using LLMs as automated research assistants.
With a large-scale replication study, we demonstrate that LLM-generated annotations lead to incorrect downstream statistical conclusions in 31-50\% of cases.
The problem extends beyond simple inaccuracy.
Researchers cannot predict when, how, or which LLMs will fail.
This unpredictability undermines the validity of any LLM-based hypothesis testing.
Without rigorous safeguards, LLM adoption poses an emerging threat to scientific integrity.

\paragraph{LLM hacking concerns extend well beyond CSS.}
Any discipline that relies on AI-generated annotations faces the same risks, including the social sciences, medicine, or law~\citep{thirunavukarasu2023large,deroy2023questioning,Bozdag2024,tseng2025evaluating}.
For example, economists have realized that downstream analyses are invalid unless measurement error and model uncertainty from the first-stage AI predictor are explicitly addressed~\citep{battaglia2024inference,Ludwig2025EconometricFramework,modarressi2025causal,carlson2025unifying}.
Similarly, \citet{feuerriegel2025using} have recently raised concerns about using AI in the behavioral sciences.
Whereas existing frameworks establish conditions and methods for valid inference, our contribution is to demonstrate empirically, at scale, how ignoring these safeguards systematically produces false discoveries, and to provide actionable standards that can guide applied research practice.

The risks we identify empirically validate concerns about measurement validity in AI systems~\citep{wallach2025position}, demonstrating how inadequate validation of LLM-based measurement instruments undermines scientific inference.
We provide the first systematic risk quantification, detailed predictor analysis, and evidence-based recommendations for the use of LLMs as automated research assistants for text annotation tasks.
While our guidelines cannot eliminate LLM hacking risk, they establish minimal standards for methodological rigor.
Reviewers should use these criteria to assess LLM-based conclusions.
Researchers must follow them to distinguish valid findings from configuration artifacts.

\paragraph{Human-LLM annotation collaboration.}
Our results suggest that human annotations and task-specific validation remain indispensable.
Pure LLM approaches underperform hybrid strategies that leverage human expertise.
Even $100$ human annotations provide better Type I error control than sophisticated LLM-based corrections.
Our findings complement recent validation frameworks like the alt-test~\citep{calderon-etal-2025-alternative} by showing that different evaluation criteria can yield conflicting assessments of LLM reliability (see Appendix~\ref{app:alt_test}).
While the alt-test validates individual annotation alignment, our approach directly evaluates the preservation of scientific conclusions.
Future systems should prioritize human-in-the-loop designs~\citep{Meng2018paradises,he-etal-2024-annollm,yao-etal-2023-beyond} rather than full automation~\citep{tseng2025evaluating}.
However, there are open questions regarding optimal collaboration methodologies~\citep{li-etal-2023-coannotating,kim-etal-2024-meganno,vaccaro2024combinations}.

\paragraph{Multiverse LLM annotation approaches.}
Multiverse analysis~\citep{Steegen2016Multiverse} offers one path forward: reporting coefficient distributions across all reasonable LLM configurations rather than single point estimates.
This transparency reveals the fragility of the robustness of findings.
Future work should explore the practical feasibility of multiverse approaches and possible insights from analyzing $p$-curves from LLM annotations~\citep{simonsohn2014p}.

\paragraph{Annotator hacking.}
Human annotators introduce their own biases~\citep{al-kuwatly-etal-2020-identifying,geva-etal-2019-modeling,Davani2024}.
Guideline changes alone can shift annotations substantially~\citep{duepublico_mods_00042132}.
Theoretically, ``annotator hacking'' through selective recruitment or instruction manipulation is possible.
However, the cost and logistics of human annotation make intentional large-scale manipulation impractical.
Besides, decades of experience have established reliable practices for human annotation.
Our study aims to develop equivalent standards for the LLM era, defining questionable research practices when using automated annotators.

\paragraph{What distinguishes LLMs from traditional supervised learning}
The risks we identify apply broadly to any supervised learning model used for data annotation.
However, traditional supervised models trained on specific datasets offer limited researcher degrees of freedom, primarily architectural choices and hyperparameters.
In contrast, LLMs enable researchers to generate entirely new classifiers through prompt engineering alone, where each paraphrase or reformulation creates a different annotator model.
This ease of generating annotations without training or even without having any ground truth data fundamentally changes the landscape of potential manipulation.

\subsection{Evidence-based recommendations for practice}

Table~\ref{tab:table1} synthesizes our empirical findings into actionable guidelines.
These recommendations acknowledge both the appeal of LLM annotation and its fundamental limitations.
Researchers must choose strategies based on their specific error tolerances, for example, prioritizing Type I error control for novel discoveries. Similarly, reviewers can use these guidelines when assessing the robustness of LLM-based annotation procedures reported.
Additionally, transparency requirements provide essential safeguards against both accidental and intentional manipulation.
While these practices cannot eliminate LLM hacking risk, they establish minimum standards for scientific rigour with increasing research automation.

\begin{practicalboxnotitle}
\begin{table}[H]
\centering
\caption{
\textbf{
Evidence-based guidelines for LLM-assisted data annotation in research
}
}
\centering
\begin{tabular}{p{4.5cm}p{9.5cm}}
\toprule
\textbf{Category} & \textbf{Practical Recommendations} \\
\midrule
\multicolumn{2}{l}{\textbf{A. LLM Hacking Risk Mitigation}} \\
\hdashline
\textbf{Task-Unspecific Mitigation} \newline \textit{(no human samples required)} & 
\begin{itemize}[leftmargin=*, topsep=0pt, itemsep=0pt]
    \item Use largest available models (70B+ show $\sim$27\% risk vs 47\% for 1B)
    \item Exercise extreme caution when $p$ values are close to significance threshold (risk >70\% near $p=0.05$)
    \item Prefer few-shot over zero-shot prompting and use detailed task descriptions over brief instructions.
\end{itemize} \\
\hdashline
\textbf{Human-Annotated Sample-Enabled} \newline \textit{(human-annotated samples required)} & 
\begin{itemize}[leftmargin=*, topsep=0pt, itemsep=0pt]
    \item Collect as many expert annotations as possible.
    \item The lowest Type I error rates are achieved by simply using randomly sampled human expert annotations, outperforming all LLM-based annotation techniques.
    \item Regression estimator correction techniques (e.g., DSL or CDI) can help in some cases but have practical limitations (e.g., they trade-off Type I vs.\ Type II errors and require enough expert annotations).
    \item Low annotation accuracy is a strong predictor of LLM hacking risk. Thus, avoid using low-quality annotations.
    \item But testing multiple models and selecting based on sample performance only yields minor improvements
    \item Do not rely on human annotator agreement to decide whether annotations should be automated or not. High human agreement rates are not associated with low LLM hacking risk.
\end{itemize} \\
\midrule
\multicolumn{2}{l}{\textbf{B. Transparency \& Reproducibility}} \\
\hdashline
\textbf{Documentation} & 
\begin{itemize}[leftmargin=*, topsep=0pt, itemsep=0pt]
    \item Report all models, versions, prompts, and parameters tested
    \item Document selection criteria and decision process
    \item Release both LLM and human annotations with analysis code
\end{itemize} \\
\hdashline
\textbf{Pre-registration} & 
\begin{itemize}[leftmargin=*, topsep=0pt, itemsep=0pt]
    \item Specify criteria for model selection, prompts, and parameters before analysis
    \item Declare hypotheses and statistical tests in advance
    \item Document planned sensitivity analyses
\end{itemize} \\
\bottomrule
\end{tabular}
\label{tab:table1}
\end{table}
\end{practicalboxnotitle}

\section*{Acknowledgements}
We would like to thank the many people who provided invaluable feedback on earlier drafts of this work:
Alexander Hoyle,
Ann-Kristin Kölln,
Carla Coccia,
Carlo Schwarz,
Cinoo Lee,
Debora Nozza,
Emanuele Moscato,
Florian E. Dorner,
Germain Gauthier,
Ingmar Weber,
Julian Walterskirchen,
Kristina Gligoric,
Nilanjana Dutt,
Philip Resnik,
Rujuta Bhagwat,
Solomon Messing,
Tijana Zrnic,
and
Vivian Nastl.
We are grateful to the Swiss AI Initiative for providing us with compute resources for this large-scale empirical study.

\clearpage

\bibliographystyle{plainnat}
\bibliography{references}

\appendix
\newpage

\section{Literature review methodology}
\label{app:lit_review_method}

We conducted a systematic literature review, which included the following four stages.

\paragraph{Stage 0: Database search.}
In March 2025, we searched five databases for papers on LLM usage in CSS research using the following search criteria:
\begin{itemize}
    \item Search period: January 1, 2022 -- March 5, 2025
    \item Databases: Scopus (362 papers), Web of Science (133), Semantic Scholar (17), ACL Anthology (57), Google Scholar (999), manually added (24)
    \item Keywords: combinations of annotation terms (human/expert/crowdsourced annotation, text classification) AND LLM terms (GPT, large language model, LLM) AND performance terms (benchmark, outperform) AND domain terms (social/communication/political science, sociology)
\end{itemize}
This \textbf{initial search yielded 1,592 papers}. We removed 102 duplicates and 1,030 papers with fewer than 3 citations per year, resulting in 453 papers for screening.

\paragraph{Stage 1: Title and abstract screening.}
Next, we reviewed the title and abstract of all 453 papers to select all that are relevant for our research. We used two inclusion criteria:
\begin{itemize}
    \item Papers benchmarking LLM performance for social science tasks
    \item Papers making recommendations for LLM usage in social science research
\end{itemize}
We excluded non-social science domains (e.g., legal reasoning). This screening identified \textbf{93 relevant papers}.

\paragraph{Stage 2: Data extraction.}
For each of the 93 relevant papers, we then extracted:
\begin{itemize}
    \item Task characteristics (type, data format, language)
    \item Models tested and prompting techniques
    \item Validation approaches mentioned
    \item LLM usage recommendations
    \item Ground truth construction
    \item Performance metrics reported
\end{itemize}

\paragraph{Stage 3: Dataset identification.}
From all of the 93 relevant papers, we identified the ones including a dataset of a CSS data annotation task. We then selected a subset of 21 datasets according to the following criteria:
\begin{itemize}
    \item Inclusion criterion: Public availability
    \item Inclusion criterion: CSS annotation tasks
    \item Exclusion criterion: Tasks cannot be formulated as multiple choice question answering task
    \item Sampled for: Available metadata for hypothesis testing
    \item Sampled for: data and task diversity
    \item Sampled for: Wide usage/citation in the field
\end{itemize}
This process informed our selection of 21 datasets (Table~\ref{tab:datasets_overview}), covering 37 annotation tasks (Table~\ref{tab:tasks_overview}), for the experiments.

\begin{table}[htb]
    \centering
    \begin{tabular}{l p{9.5cm} r}
         \toprule
          \textbf{Database} & \textbf{Search Query} & \textbf{Papers Found} \\
         \midrule
         & \multirow{2}{9.5cm}{{\small \footnotesize{ALL(("human annotation” OR "expert annotation” OR "crowdsourced annotation” OR "human coder*” OR "human coding” OR "text classification” OR "text categorization” OR "manual annotation”) AND ("gpt” OR "large language model” OR "LLM*”) AND ("benchmark” OR "outperform”) AND ("social science” OR "communication science” OR "sociology” OR "political science”)) AND PUBYEAR AFT 2021}}} &  \\
         &  &  \\
         &  &  \\
        \href{https://www.scopus.com/search/form.uri}{Scopus} &  & 362 \\
         &  &  \\
         &  &  \\
         &  &  \\\hdashline
         & \multirow{2}{9.5cm}{{\small \footnotesize{(human annotation OR expert annotation OR crowdsourced annotation OR human coder OR human coding OR text classification OR text categorization OR manual annotation) AND (gpt OR large language model OR LLM) AND (benchmark OR outperform) AND (social science OR communication science OR sociology OR political science)}}} &  \\
        \href{www.webofscience.com}{Web of Science} &  & 133 \\
         &  &  \\\cdashline{1-1}\cdashline{3-3}
         &  &  \\
        \href{https://www.semanticscholar.org/}{Semantic Scholar} &  & 17 \\
         &  &  \\\hdashline
         & \multirow{2}{9.5cm}{{\small \footnotesize{("human annotation*” OR "expert annotation*” OR "crowdsourced annotation*” OR "human coder*” OR "human coding*” OR "text classification*” OR "text categorization*” OR "manual annotation*”) AND ("gpt” OR "large language model*” OR "LLM*”) AND ("benchmark*” OR "outperform*”) AND ("social science*” OR "communication science*” OR "sociology” OR "political science*”)}}} &  \\
        \href{https://aclanthology.org/}{ACL Anthology} &  & 57 \\
         &  &  \\\cdashline{1-1}\cdashline{3-3}
         &  &  \\
        \href{https://scholar.google.com/}{Google Scholar} &  & 999 \\
         &  &  \\\hdashline
        Manually added & - & 24 \\
         \bottomrule
         & & \textit{Total: 1,592}\\
    \end{tabular}
    \caption{Exact search query and number of papers found for each database considered in the literature search.}
    \label{tab:lit_search_keywords}
\end{table}

\section{Structured results of literature review for in-scope articles}
\label{app:lit_review_results}

This section provides detailed results from our systematic literature review on LLM usage for data annotation in CSS.
We analyzed 93 papers to understand current practices, recommendations, and gaps in the field as discussed in Section~\ref{sec:Literaturereview}.

Table~\ref{tab:literature_review} presents a comprehensive overview of all reviewed papers, including their recommendations regarding LLM usage and validation practices.
Papers are sorted by their recommendation type to highlight the distribution of opinions in the field.
A large fraction of the papers we reviewed recommend using LLMs for automated data annotation without rigorous model validation.
Research often suggests that LLMs should be used as data annotators on the grounds of measured model performance for a given task, while highlighting the need for rigorous model validation.
Only $8$ of identified research papers argue against the use of LLMs for automated data annotations.
Only four of the reviewed papers raise concerns of erroneous LLM annotations affecting downstream statistical conclusions~\citep{egami2023using,pangakis2023automated,baly-etal-2020-detect,barrie2024replication,gligoric-etal-2025-unconfident}.
Notice also that about 75\% of studies focused exclusively on English text.

\input{tables/literature_review_table.tex}

\section{Additional experimental details}

\subsection{Models}
\label{app:Models}

We test 18 models:
\begin{itemize}
    \item meta-llama/Llama-3.2-1B-Instruct
    \item meta-llama/Llama-3.2-3B-Instruct
    \item meta-llama/Llama-3.1-8B-Instruct
    \item meta-llama/Llama-3.1-70B-Instruct
    \item Qwen/Qwen2.5-1.5B-Instruct
    \item Qwen/Qwen2.5-3B-Instruct
    \item Qwen/Qwen2.5-7B-Instruct
    \item Qwen/Qwen2.5-32B-Instruct
    \item Qwen/Qwen2.5-72B-Instruct
    \item Qwen/Qwen3-1.7B
    \item Qwen/Qwen3-4B
    \item Qwen/Qwen3-8B
    \item Qwen/Qwen3-32B
    \item google/gemma-3-1b-it
    \item google/gemma-3-4b-it
    \item google/gemma-3-27b-it
    \item gpt-4o-mini-2024-07-18
    \item gpt-4o-2024-08-06
\end{itemize}
All 18 models are instruction-tuned model variants.
We used the model snapshots gpt-4o-2024-08-06 and gpt-4o-mini-2024-07-18 that the OpenAI API points to for the model families gpt-4o and gpt-4o-mini at the time we ran our experiments.
Open-weight model inference used the HuggingFace Transformers library~\citep{wolf-etal-2020-transformers} with full precision (no quantization) to ensure maximum annotation quality.

\subsection{Dataset groupings}
\label{app:Dataset_groupings}

Each dataset is split into binary groups using both metadata-based and text-based approaches to generate realistic hypothesis tests.
For datasets with original metadata (e.g., author gender, political party affiliation), we create groupings based on these attributes.
Additionally, we generate default applicable to any text dataset based on:
\begin{itemize}
\item \textbf{Keyword presence}: Binary splits based on whether text contains specific keywords, selected through:
  \begin{itemize}
  \item The 3 most frequent words in the dataset (excluding stopwords)
  \item The 3 most frequent words per ground truth class (up to 15 classes)
  \item Words that maximize the absolute difference in class proportions between groups
  \end{itemize}
\item \textbf{Text length}: Splitting at the median character count (short vs.\ long texts)
\item \textbf{Random splits}: Multiple random 50/50 splits and unbalanced splits (20/80, 40/60) using different random seeds
\end{itemize}
This automated approach generates diverse, data-driven hypotheses while ensuring that each grouping creates a meaningful contrast that could plausibly be of research interest.

\subsection{Prompt paraphrases}
\label{app:Prompt paraphrases}

Most annotation tasks in this study have been used in prior work to benchmark LLM performance.
Where available, we use prompts from prior work.
To test sensitivity to prompt formulation, we generated paraphrased versions to ensure at least five prompt variations per task.
Paraphrases were generated using OpenAI's o3-2025-04-16 reasoning model with parameters \texttt{\{max\_completion\_tokens=$20000$, reasoning\_effort=high\}}.
The paraphrasing prompt template used was:

\begin{tcolorbox}[colframe=gray!90, colback=gray!10, title=Prompt used to instruct o3 model to paraphrase data annotation prompt., fonttitle=\bfseries, boxrule=0.2pt, left=5pt, right=5pt, top=5pt, bottom=5pt, arc=0.5mm, boxsep=3pt, before skip=8pt, after skip=8pt]
\small
\texttt{%
Paraphrase the following prompt that will be used for automated data annotation.
\\\\
Key requirements:\\
- Preserve the exact same annotation task\\
- Keep all placeholders (e.g., \{text\}) unchanged\\
- Ensure the paraphrased prompt guides the model to output ONLY the keys from this mapping dictionary:
  \{prompt\_output\_mapping\_placeholder\}\\
- Base your paraphrase on the content in the original prompt without adding new information\\
- Improve clarity and structure while maintaining the original meaning
\\\\
Original prompt to paraphrase:\\
\{prompt\_placeholder\}
\\\\
Provide only the paraphrased prompt.
}
\end{tcolorbox}
Here, \{prompt\_output\_mapping\_placeholder\} is filled with the list of valid class labels, and \{prompt\_placeholder\} is filled with the base prompt to be paraphrased.
Generated paraphrases were manually reviewed to be appropriate for the task and adjusted only when necessary (e.g., to correctly escape special characters).
This systematic paraphrasing approach allowed us to test whether minor variations in prompt formulation affect downstream statistical conclusions.
All original prompts, paraphrases, and their compatible output mappings are available in our code repository for full reproducibility.

\subsection{Low confidence sampling with verbalized confidence scores}
\label{app:VerbalizedLowconfidencesampling}

We elicit verbalized LLM annotation confidence using similar prompts as~\citet{gligoric-etal-2025-unconfident}, though slightly adapted to match each annotation task. For example, for the \textit{tone} task, we use the following prompt, where 
\{previous\_answer\_placeholder\} corresponds to the LLMs' annotation of \{text\}:
\begin{tcolorbox}[colframe=gray!90, colback=gray!10,
fonttitle=\bfseries, boxrule=0.2pt, left=5pt, right=5pt, top=5pt, bottom=5pt, arc=0.5mm, boxsep=3pt, before skip=8pt, after skip=8pt]
\small
\texttt{%
How likely is it that the tone of the following political advertisement is \{previous\_answer\_placeholder\}?
\\\\
Output only a single number between 0 and 1, without any context or explanation.
\\\\
Political advertisement: \{text\}
\\\\
Probability:
}
\end{tcolorbox}

Histograms and calibration plots for verbalized confidence scores across all annotation tasks are presented in Appendix~\ref{app:Verbalized_confidence_scores}.

\subsection{Active sampling implementation details}
\label{app:Activesamplingimplementationdetails}

We use the same prediction model and training parameters as~\citet{gligoric-etal-2025-unconfident}: an XGBoost model with $2$,$000$ boosting rounds, a step size of $0.001$, a maximum depth of $3$, and a squared-error objective.
Since we use human annotation budgets of $n_{\text{human}} = [25, 50, 100, 250, 500, 1000]$ in our experiments, the first $25$ burn-in samples (with at least one example of both classes) are used for initialization, and we retrain the XGBoost model after every batch.

\subsection{DSL implementation details}
\label{app:dsl_details}

We implement DSL with default parameters of 5-fold cross-fitting and 10 sample splits, using the R package \texttt{DSL}~\citep{dslRpackage}.
When these fail due to insufficient labeled data, we adaptively try alternative combinations: cross-fit values $\{5, 4, 3, 2, 6, 7, 8, 9, 10\}$ and sample-split values $\{10, 8, 6, 4, 3, 2, 12, 15\}$ in order until successful estimation is achieved.

For low confidence sampling, we set sampling probabilities inversely proportional to the LLM's verbalized confidence scores (mixed with $10\%$ uniform sampling for stability): $\pi_i = 0.1 \cdot \pi_{\text{base}} + 0.9 \cdot (1 - c_i)$, where $c_i$ is the LLM's verbalized confidence and $\pi_{\text{base}} = n_{\text{human}}/n$, ensuring higher sampling probability for instances with lower confidence.

\subsection{CDI implementation details}
\label{app:cdi_details}

The Confidence-Driven Inference (CDI) method introduced by~\citet{gligoric-etal-2025-unconfident} combines human annotations with LLM predictions using an optimized trust parameter $\lambda \in [0,1]$ that minimizes estimator variance while maintaining unbiasedness.

To handle numerical instabilities with sparse labeled data, we implement adaptive initialization by attempting $\lambda_{\text{init}} \in \{1.0, 0.9, 0.7, 0.4, 0.1, 0.0\}$ in descending order until stable optimization is achieved.
In case of quasi-complete separation or extreme sparsity (resulting in negative variance), we fall back to $\lambda = 0$ (human annotations only).

\subsection{Experiment compute costs}

Experiments were run on an internal cluster with Nvidia GH200 and Nvidia A100-80GB GPUs.
API costs for GPT-4o and GPT-4o-mini models totaled approximately $\$520$ for all data annotations and verbalized confidence elicitation across the $145,277$ datapoints and multiple prompt variations.

Total GPU hours amounted to about $11$,$000$ hours, including LLM inference for data annotation, verbalized confidence score elicitation, downstream statistical analyses, and regression estimator correction techniques (DSL and CDI).

\section{Additional Metrics}
\label{app:AdditionalMetrics}

The \textbf{False Discovery Rate quantifies the fraction of false positives among all discoveries}: among all cases where LLM annotations detect a significant effect, what proportion are incorrect discoveries?
\begin{equation}
\text{False Discovery Rate} = \frac{\sum_{t \in T} \sum_{h \in H_t} \sum_{\phi \in \Phi} \mathbb{1}[S^{\text{GT}}_h = 0, S^{\text{LLM}}_{h,\phi} = 1]}{\sum_{t \in T} \sum_{h \in H_t} \sum_{\phi \in \Phi} \mathbb{1}[S^{\text{LLM}}_{h,\phi} = 1]}
\end{equation}
The \textbf{False Nondiscovery Rate quantifies the fraction of false negatives among all non-discoveries}: among all cases where LLM annotations fail to detect a significant effect, what proportion incorrectly misses true effects?
\begin{equation}
\text{False Nondiscovery Rate} = \frac{\sum_{t \in T} \sum_{h \in H_t} \sum_{\phi \in \Phi} \mathbb{1}[S^{\text{GT}}_h = 1, S^{\text{LLM}}_{h,\phi} = 0]}{\sum_{t \in T} \sum_{h \in H_t} \sum_{\phi \in \Phi} \mathbb{1}[S^{\text{LLM}}_{h,\phi} = 0]}
\end{equation}
The \textbf{Type S Error Rate captures the fraction of sign errors among true discoveries}: among all cases where both ground truth and LLM annotations detect significant effects, what proportion have incorrect signs?
\begin{equation}
\text{Type S Error Rate} = \frac{\sum_{t \in T} \sum_{h \in H_t} \sum_{\phi \in \Phi} \mathbb{1}[S^{\text{GT}}_h = 1, S^{\text{LLM}}_{h,\phi} = 1, \text{sgn}(\beta^{\text{GT}}_h) \neq \text{sgn}(\beta^{\text{LLM}}_{h,\phi})]}{\sum_{t \in T} \sum_{h \in H_t} \sum_{\phi \in \Phi} \mathbb{1}[S^{\text{GT}}_h = 1, S^{\text{LLM}}_{h,\phi} = 1]}
\end{equation}

These metrics provide a complementary perspective to the Type I risk and Type II risk metrics by conditioning on LLM decisions rather than ground truth outcomes.
In line with the false discovery rate framework of~\citet{Benjamini1995}, these rates focus on the reliability of discoveries made by the LLM testing procedure, providing insight into the practical trustworthiness of LLM-based findings.

\section{Additional results}
\label{app:Additionalresults}

\subsection{Complete LLM hacking risk breakdown by task and model}
\label{app:llm_hacking_risk_by_task_model}

Table~\ref{tab:PAPER_llm_hacking_risk_by_task_model} provides the complete breakdown of LLM hacking risk across all 37 annotation tasks and 18 models evaluated.
There are large variations both across tasks (ranging from 5\% for humor detection with larger models to over 75\% for some relevance classification tasks) and within tasks across different models.

\begin{table}[htb]
\input{tables/PAPER_llm_hacking_risk_by_task_model.tex}
\label{tab:PAPER_llm_hacking_risk_by_task_model}
\end{table}

\subsection{Instruction-following failure rates}
\label{app:fraction_na}

Table~\ref{tab:PAPER_fraction_na_table} reports the fraction of cases where models produced outputs that could not be mapped to valid annotation categories.
These instruction-following failures occur when models generate responses outside the specified format or provide explanatory text instead of categorical labels.
Higher failure rates occur for smaller models.
We exclude model-task-prompt combinations with failure rates exceeding 1\% from our analyses to ensure robust statistical comparisons, though this conservative threshold means our reported hacking risks may underestimate the true extent of the problem for models with poor instruction-following capabilities.

\begin{table}[htb]
\input{tables/PAPER_fraction_na_table.tex}
\label{tab:PAPER_fraction_na_table}
\end{table}

\subsection{LLM hacking predictors}
\label{ssec:regression}

Table~\ref{tab:regression_llm_hacking} presents the OLS regression results predicting binary LLM hacking outcomes using a Linear Probability Model.
While the R$^{2}$ of $0.154$ appears modest, this is typical for binary outcomes where explained variance is inherently constrained.
The model achieves $83.0\%$ classification accuracy, demonstrating strong predictive performance.
The highly significant F-statistic (F = $2$,$352.722$, $p < 0.001$) confirms that the predictors jointly have significant explanatory power.
With over $1.4$ million observations, the analysis provides substantial statistical power to detect meaningful relationships between prompt characteristics, model performance, and hacking susceptibility.

{\scriptsize
\begin{longtable}{@{\extracolsep{1pt}}lc} 

\caption{OLS Regression Results: LLM hacking risk by prompt type, prompt detail, weighted F1 score with task (T) and model (M) interactions, and significance-related predictors} 
\label{tab:regression_llm_hacking} \\
\toprule
 & Dependent variable: LLM Hacking \\ 
\midrule
\endfirsthead

\multicolumn{2}{c}%
{{\tablename\ \thetable{} -- continued from previous page}} \\
\toprule
 & Dependent variable: LLM Hacking \\ 
\midrule
\endhead

\midrule \multicolumn{2}{r}{{Continued on next page}} \\ 
\endfoot

\bottomrule
\endlastfoot
\input{tables/regression_llm_hacking.tex}
\end{longtable}
}

\subsection{Correlations between LLM hacking and F1 score by task}
\label{ssec:correlations_f2_llmhacking}

\begin{table}[htb]
\input{tables/PAPER_llm_hacking_f1_score_weighted_correlations_table}
\end{table}

\subsection{LLM annotation reliability}
\label{ssec:LLMannotationreliability}

\begin{figure*}[htb]
    \centering
    \includegraphics[width=\textwidth]{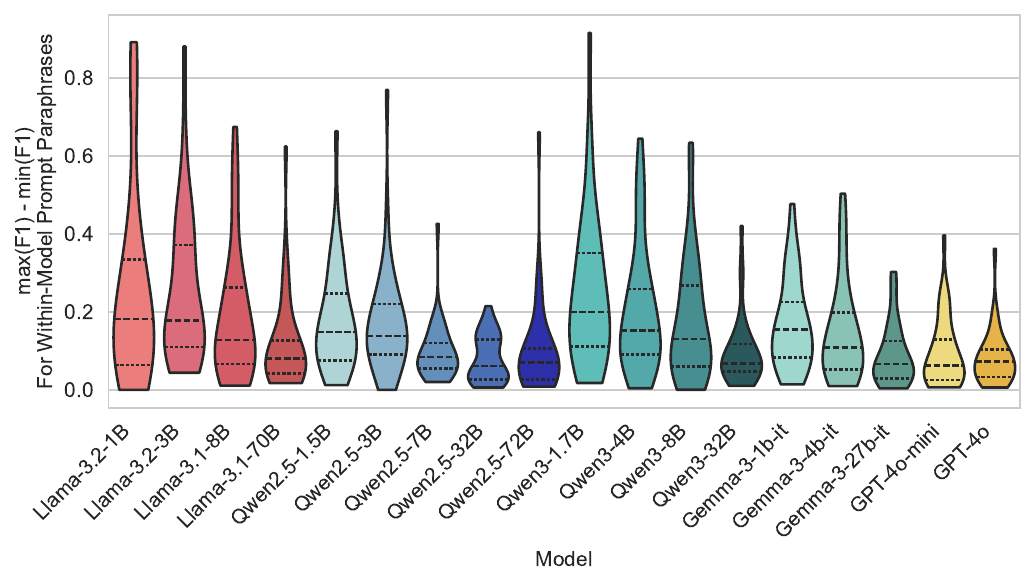}
    \caption{Distribution of performance spread, measured as the absolute differences in weighted F1 across multiple prompt variations for each model.
    The large y-axis values indicate low output consistency across different prompting strategies.}
    \label{fig:performance_reliability_abs_diff_violin}
\end{figure*}

Figure~\ref{fig:performance_reliability_abs_diff_violin}shows that LLMs are unreliable annotators. Performance varies by >10 percentage points on average when comparing the best- with the worst-performing prompt.
These results are in line with the findings by~\citet{sclar2024quantifying,Atreja_Ashkinaze_Li_Mendelsohn_Hemphill_2025}.
Smaller models show wider distributions of performance differences across prompts, indicating less stable annotation behavior compared to larger models.

\subsubsection{Using larger, more capable models reduced LLM hacking risk}

\begin{figure*}[htb]
    \centering
    \includegraphics[width=0.9\textwidth]{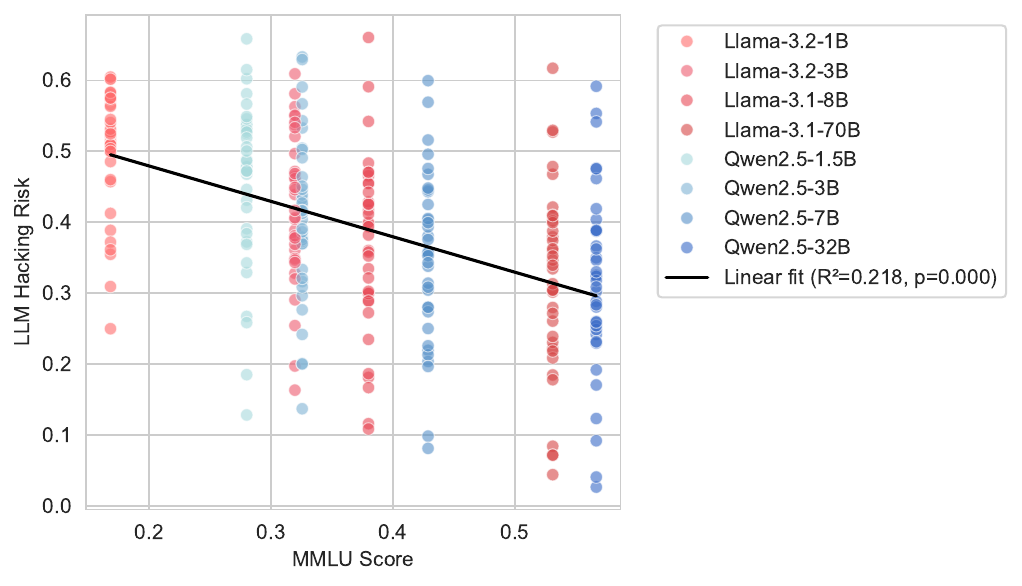}
    \caption{Negative correlation between LLM hacking risk and general model performance (MMLU-PRO scores) for eight Llama and Qwen models with known MMLU-PRO scores.}
    \label{fig:llm_hacking_vs_mmlu}
\end{figure*}

Figure~\ref{fig:llm_hacking_vs_mmlu} demonstrates a negative correlation between LLM hacking risk and general model capability as measured by MMLU-PRO benchmark scores.
The analysis includes eight models from the Llama and Qwen families with publicly available MMLU-PRO scores.

While this suggests that model capability improvements can reduce LLM hacking risk, the fact that even the highest-scoring models retain considerable risk (up to 52\%) highlights the limitations of relying solely on model scaling.

\subsection{LLM hacking feasibility by annotation task}
\label{app:feasibility}

\begin{figure*}[htb]
    \centering
    \includegraphics[width=\textwidth]{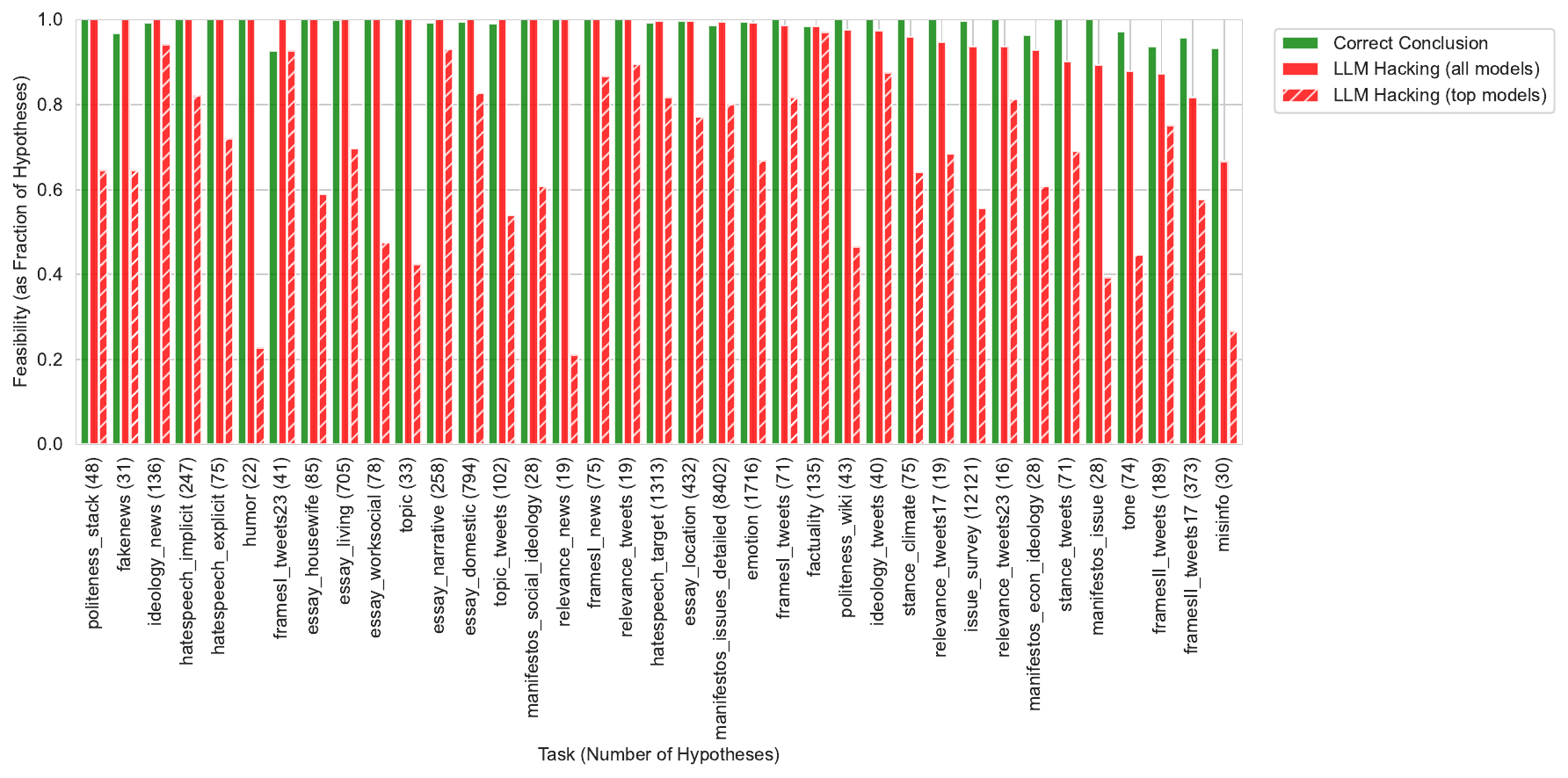}
    \caption{LLM hacking and correct conclusion feasibility by annotation task. Each task is labeled with the total number of unique hypotheses tested (in parentheses). Feasibility represents the fraction of hypotheses within each task where at least one model-prompt combination achieved the specified outcome.}
    \label{fig:llm_hacking_feasibility_by_task}
\end{figure*}

Expanding the results of Section~\ref{ssec:DeliberateLLMhacking}, we provide the full overview of LLM hacking feasibility by annotation task in Figure~\ref{fig:llm_hacking_feasibility_by_task}.

Table~\ref{tab:llm_hacking_feasibility_restrictions} presents the mean feasibility rates (as percentages with values averaged across all tasks).
We consider the following restrictions to the configuration space to compare different settings for seemingly rigorous methodological setups.
\begin{itemize}
    \item \textbf{any model and any prompt} represents the unrestricted case where any single model-prompt combination can be selected.
    \item \textbf{any top model} restricts to the following seven high-performing models only: 
    Llama-3.1-70B,
    Qwen2.5-32B,
    Qwen2.5-72B,
    Qwen3-32B,
    Gemma-3-27b,
    GPT-4o-mini, and
    GPT-4o.
    \item \textbf{at least 2 prompts} requires finding feasible solutions across multiple prompts for the same model.
    \item \textbf{at least 2 models} requires finding feasible solutions across multiple models for the same prompt.
\end{itemize}

\begin{table}[htbp]
\centering
\caption{LLM Hacking Feasibility for various risk types across different model-prompt configuration restrictions.
Results highlighted in green and red are shown in Figure~\ref{fig:llm_hacking_feasibility_task_weighted} in the main text.
}
\label{tab:llm_hacking_feasibility_restrictions}
\begin{tabular}{cl|cc:cc:cc}
\toprule
 & \textbf{Metric} & \textbf{\rotatebox{90}{\shortstack{Any model and \\ any prompt}}} & \textbf{\rotatebox{90}{\shortstack{Any top model and \\ any prompt}}} & \textbf{\rotatebox{90}{\shortstack{Any model with \\ at least 2 prompts}}} & \textbf{\rotatebox{90}{\shortstack{Any top model with \\ at least 2 prompts}}} & \textbf{\rotatebox{90}{\shortstack{Any prompt with \\ at least 2 models}}} & \textbf{\rotatebox{90}{\shortstack{Any prompt with \\ at least 2 top models}}} \\
\midrule
 & Overall LLM Hacking Feasibility Rate & 96.3\% & 66.5\% & 92.8\% & 56.7\% & 93.3\% & 58.4\% \\
 & Overall Correctness Feasibility Rate & 98.9\% & 95.1\% & 98.0\% & 91.9\% & 97.9\% & 92.0\% \\ \myhdashline
 \multirow{4}{*}{$\mathbf{\text{\textbf{H}}_0}$} & $\text{H}_0$ Correctness Feasibility Rate & \colorbox{green}{99.2\%} & \colorbox{green}{95.3\%} & 98.8\% & 93.2\% & 99.2\% & 92.5\% \\
 & Type I Error Feasibility Rate & \colorbox{red}{94.4\%} & \colorbox{red}{69.7\%} & 88.6\% & 59.0\% & 89.8\% & 61.1\% \\
 & Type I Error Feasibility Rate ($\beta^{\text{LLM}}>0$) & 71.2\% & 38.3\% & 54.6\% & 28.9\% & 54.6\% & 30.6\% \\
 & Type I Error Feasibility Rate ($\beta^{\text{LLM}}<0$) & 82.5\% & 42.5\% & 68.2\% & 33.9\% & 68.9\% & 35.4\% \\ \myhdashline
 \multirow{5}{*}{$\mathbf{\text{\textbf{H}}_\text{\textbf{A}}}$} & $\text{H}_\text{A}$ Correctness Feasibility Rate & \colorbox{green}{96.6\%} & \colorbox{green}{90.1\%} & 94.4\% & 84.3\% & 94.1\% & 85.2\% \\
 & Type II Error Feasibility Rate & \colorbox{red}{98.1\%} & \colorbox{red}{68.1\%} & 96.2\% & 61.4\% & 96.1\% & 62.1\% \\
 & Type S Error Feasibility Rate & \colorbox{red}{68.1\%} & \colorbox{red}{19.9\%} & 47.2\% & 13.3\% & 51.5\% & 14.5\% \\
 & Type S Error Feasibility Rate ($\beta^{\text{LLM}}>0$) & 78.5\% & 24.4\% & 59.2\% & 16.8\% & 62.3\% & 18.3\% \\
 & Type S Error Feasibility Rate ($\beta^{\text{LLM}}<0$) & 54.4\% & 13.7\% & 32.0\% & 8.6\% & 37.2\% & 9.3\% \\
\bottomrule
\end{tabular}
\end{table}

\subsection{Verbalized confidence scores}
\label{app:Verbalized_confidence_scores}

\begin{figure*}[htbp]
    \centering
    \includegraphics[width=0.75\textwidth]{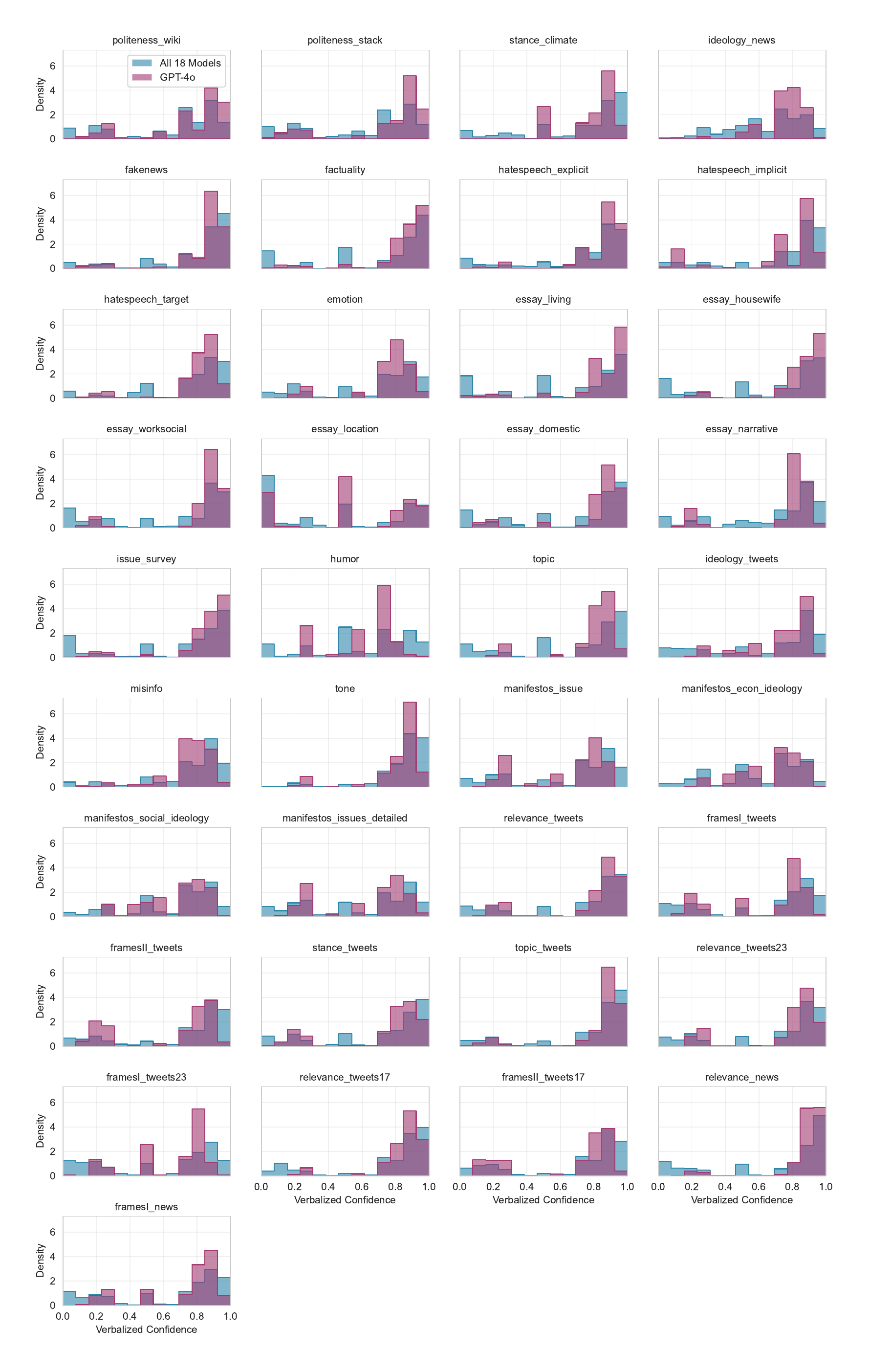}
    \caption{Distribution of verbalized confidence scores across 37 annotation tasks. Blue histograms show all 18 models; orange shows GPT-4o alone. Bimodal distributions indicate models distinguish uncertain from confident predictions.}
    \label{fig:verbalized_confidence_histogram}
\end{figure*}

Figure~\ref{fig:verbalized_confidence_histogram} shows the distribution of verbalized confidence scores across all tasks.
GPT-4o generally shows higher confidence compared to the aggregate of all 18 models.

\begin{figure*}[htbp]
    \centering
    \includegraphics[width=0.75\textwidth]{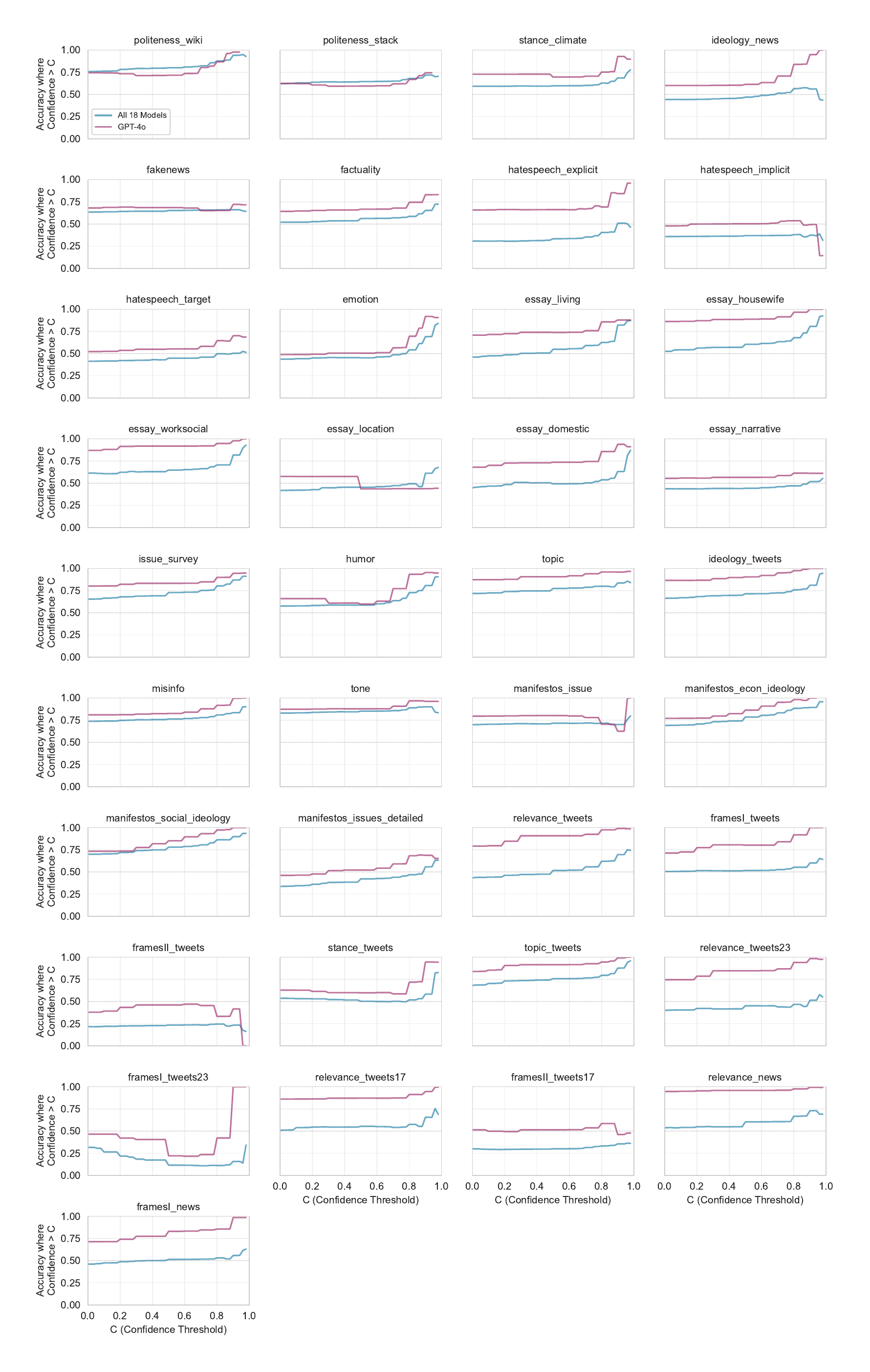}
    \caption{Calibration curves showing accuracy for predictions above confidence threshold $C$.
    LLM annotation accuracy with respect to human annotations on the y-axis, among instances where the confidence score is greater than a threshold $C$ as indicated on the x-axis.}
    \label{fig:verbalized_confidence_accuracy_calibration}
\end{figure*}

Figure~\ref{fig:verbalized_confidence_accuracy_calibration} shows that verbalized confidence and actual accuracy are not very well calibrated.
GPT-4o shows better calibration of verbalized confidence for most tasks, compared to the average of all 18 models.

\subsection{Alternative annotator test comparison}
\label{app:alt_test}

To better understand the relationship between LLM annotation quality and downstream statistical reliability, we compare our LLM hacking risk metrics with results from the Alternative Annotator Test (alt-test) proposed by~\citet{calderon-etal-2025-alternative} for the five annotation tasks without perfect human annotator agreement (see $K.\alpha$ in Table~\ref{tab:tasks_overview}.
The alt-test provides a statistical framework for determining whether LLMs can replace individual human annotators by comparing their alignment with annotation consensus.

\paragraph{Alt-test metrics.}
The alt-test uses two primary metrics:
\begin{itemize}
\item \textbf{Winning rate (WR $\omega$)}: The proportion of human annotators that the LLM statistically ``wins'' against, calculated as the fraction of rejected null hypotheses in paired comparisons.
If $\omega \geq 0.5$, the LLM is said to pass the alt-test, indicating the LLM performs comparably to or better than the majority of human annotators.
\item \textbf{Average advantage probability (AP $\rho$)}: The probability that LLM annotations are as good as or better than those of a randomly chosen human annotator.
\end{itemize}
To account for the higher cost and effort associated with human annotation, we set the alt-test's cost-benefit hyperparameter to $\varepsilon=0.1$.

\paragraph{LLM hacking and the alt-test address fundamentally different questions.}
The alt-test asks: ``Can this LLM replace an individual human annotator?''
In contrast, our LLM hacking analysis asks: ``Will using this LLM's annotations preserve the validity of downstream statistical conclusions?''
This distinction is crucial because our ground truth represents aggregated expert consensus (typically with high inter-annotator agreement), not individual annotator opinions.

\begin{figure*}[htb]
    \centering
    \includegraphics[width=\textwidth]{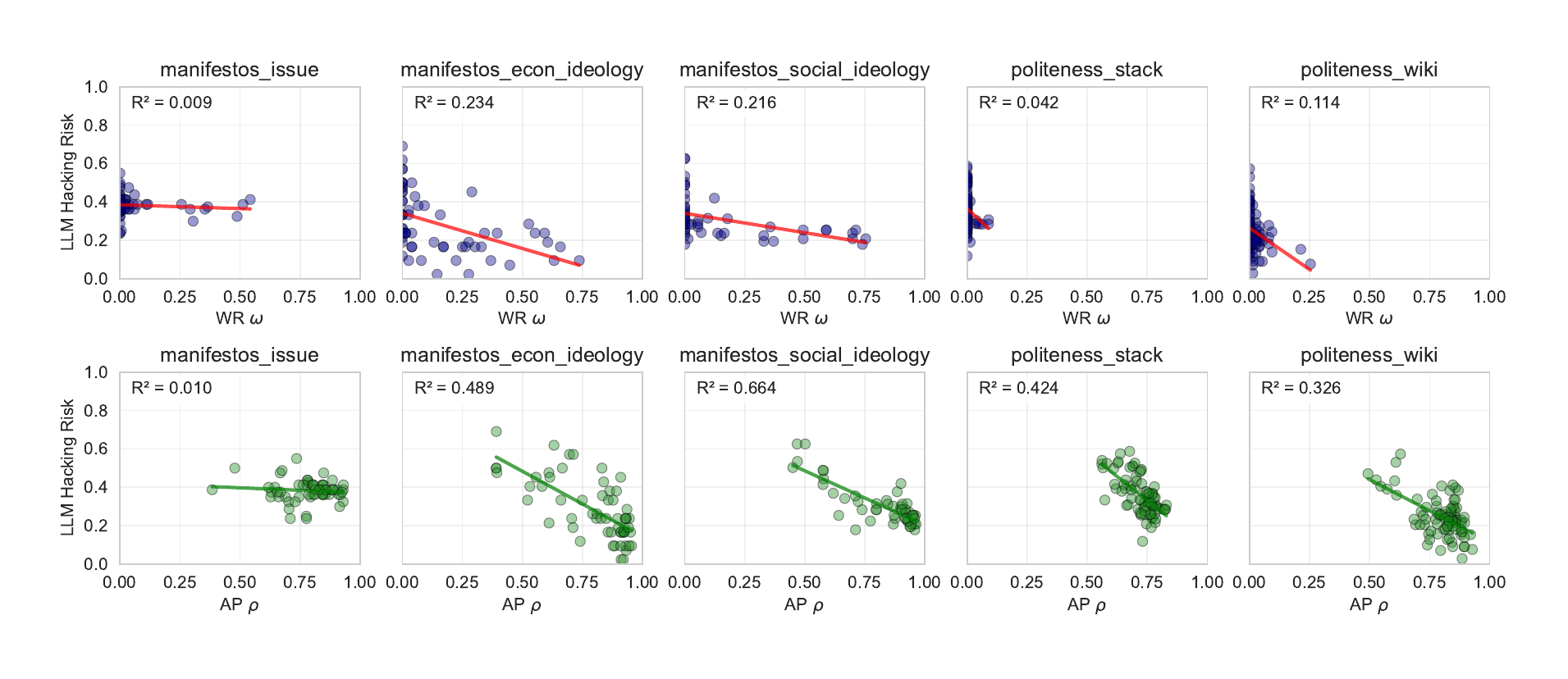}
    \caption{Relationship between LLM hacking risk and alt-test metrics across five annotation tasks with $K.\alpha<1$.}
    \label{fig:PAPER_llm_hacking_vs_alt_test}
\end{figure*}

\paragraph{Correlation analysis.}
Figure~\ref{fig:PAPER_llm_hacking_vs_alt_test} shows heterogeneous relationships between alt-test metrics and LLM hacking risk across tasks.
For most annotation tasks (except \textit{manifestos\_issue}), we observe moderate to strong negative correlations with AP, suggesting that LLMs with higher individual annotator alignment tend to produce lower hacking risk.

The WR generally shows weaker negative correlations with LLM hacking risk compared to AP.
Overall, these results demonstrate that \textbf{passing the alt-test -- while valuable for validating individual annotation quality -- does not guarantee reliable downstream statistical inference}.
An LLM might successfully replicate individual annotator behavior (high $\omega$ and $\rho$) yet still introduce systematic biases that distort regression coefficients and hypothesis test conclusions.
This underscores the importance of task-specific validation that considers both annotation quality and the intended downstream analyses.
 
\subsection{Mitigating LLM hacking risk}
\label{app:MitigatingLLMhackingrisk}

Here, we extend the results from Section~\ref{ssec:MitigatingLLMhackingrisk}, providing a complete overview of all mitigation techniques and all annotation tasks.
We also provide additional details on model selection strategies.

\subsubsection{Sampling strategy comparison}
\label{app:sampling_strategy_comparison}

\begin{figure*}[htb]
    \centering
    \includegraphics[width=\textwidth]{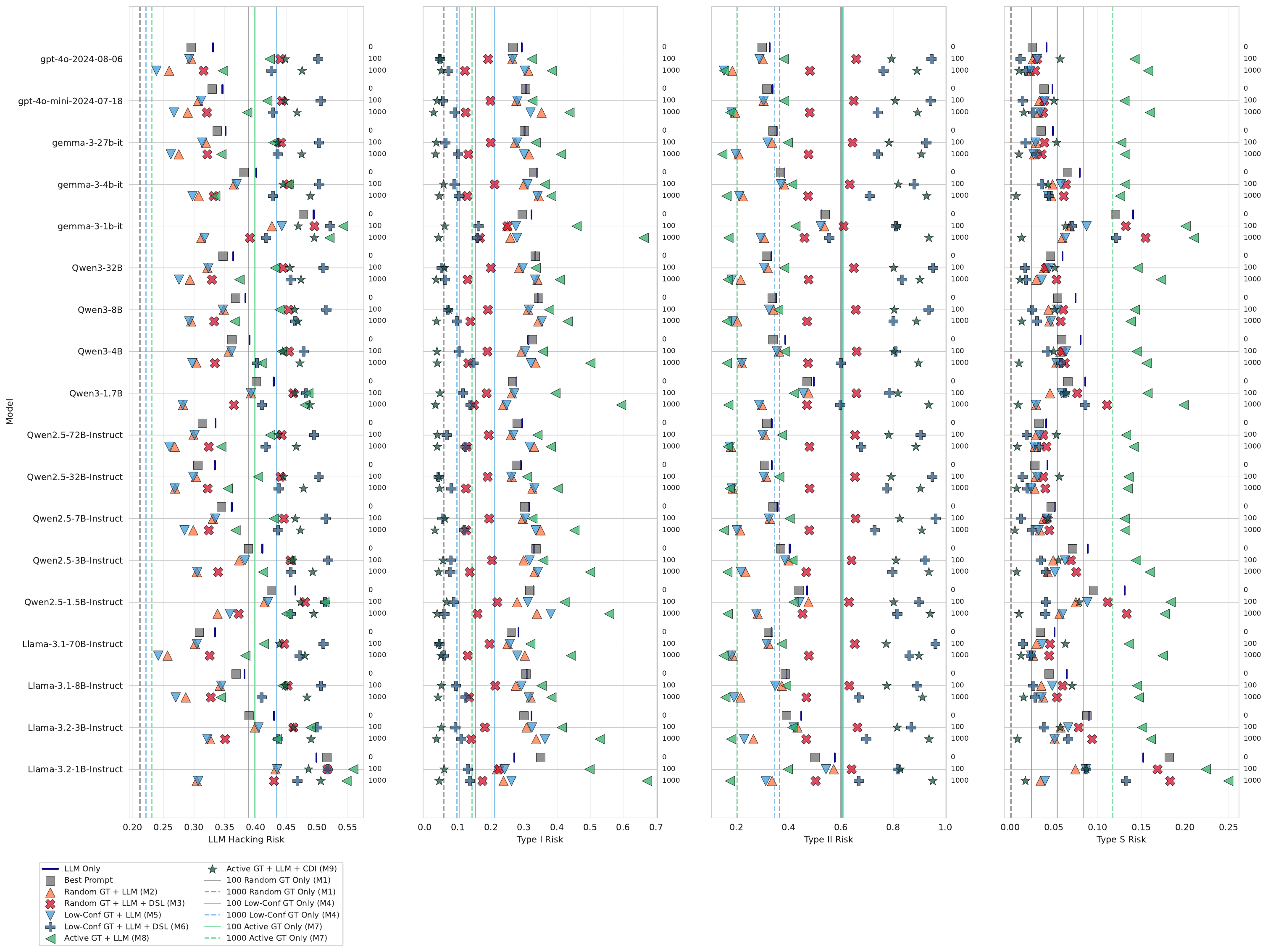}
    \caption{
    Effectiveness of mitigation strategies across models and error types. Each subplot shows a different risk metric with models on the y-axis and risk values on the x-axis. Points are vertically offset by the number of ground truth samples, as indicated by the numbers on the right (0, 100, or 1,000 samples).
    }
    \label{fig:llm_hacking_mitigation_techniques_ALL}
\end{figure*}

While Section~\ref{ssec:MitigatingLLMhackingrisk} focuses on the most distinct mitigation strategies, Figure~\ref{fig:llm_hacking_mitigation_techniques_ALL} reveals that the choice of sampling strategy (random vs.\ low confidence vs.\ active) has minimal impact on overall effectiveness.
The differences between M1, M4, and M7 (all using ground truth only) are negligible, as are the differences between M2, M5, and M8 (all using ground truth + LLM).
Hence, sophisticated sampling strategies provide limited benefit over simple random sampling when the number of human annotations is fixed.
Notice that results for $n_{\text{human}}=1,000$ do not contain tasks with very few datapoints, which means that the results may not be perfectly comparable to the results with $n_{\text{human}}=1,000$.

\begin{figure*}[htb]
    \centering
    \includegraphics[width=0.8\textwidth]{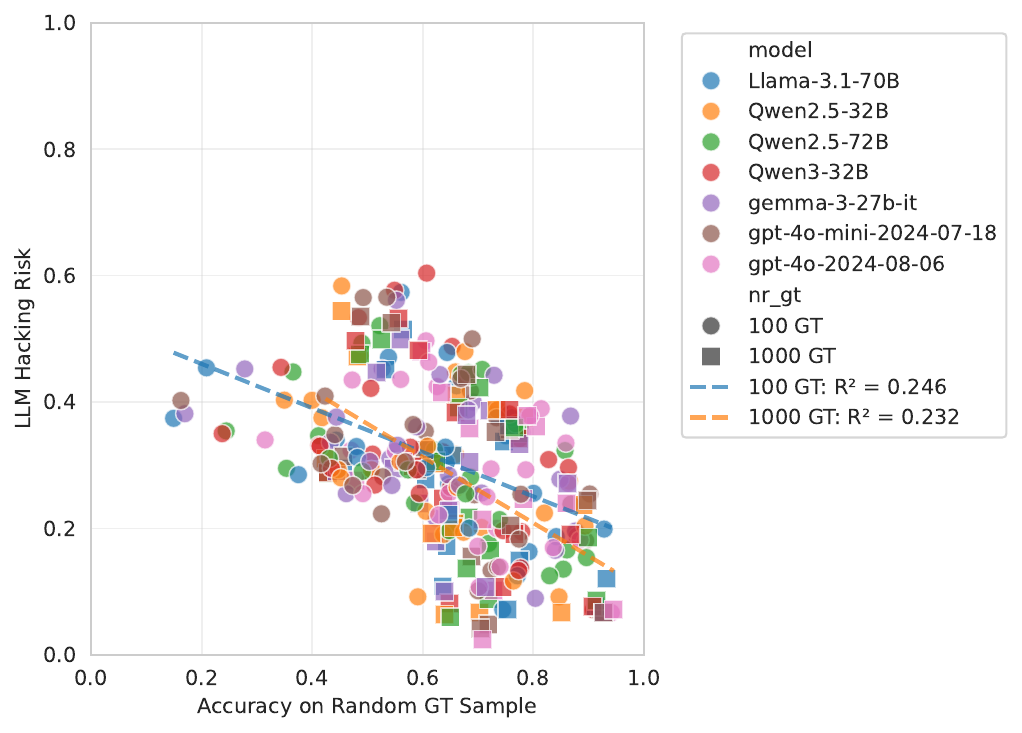}
    \caption{Significant negative correlation between LLM hacking risk and model accuracy derived from small samples of known ground truth data ($n_{\text{human}} \in \{100, 1000\}$). Each dot and square corresponds to one task-model-$n_{\text{human}}$ run.}
    \label{fig:accuracy_vs_risk_correlation_all_tasks}
\end{figure*}

\subsubsection{Prompt performance variation}
\label{app:prompt_performance_variation}

The ``best prompt'' results shown throughout our analysis represent an optimistic scenario where researchers have access to enough ground truth labels to select the optimal prompt.
This provides an upper bound on what careful prompt engineering might achieve.
In practice, the baseline results (averaged across all reasonable prompts) better represent expected performance when researchers lack extensive validation data.

Some tasks show a big gap between best-prompt and baseline performance of up to 5 percentage points and more.
Hence, prompt sensitivity is itself a task characteristic that researchers should consider when considering using LLM annotations.

\subsubsection{Model selection}
\label{app:model_selection_best_models}

\begin{figure*}[htb]
    \centering
    \includegraphics[width=\textwidth]{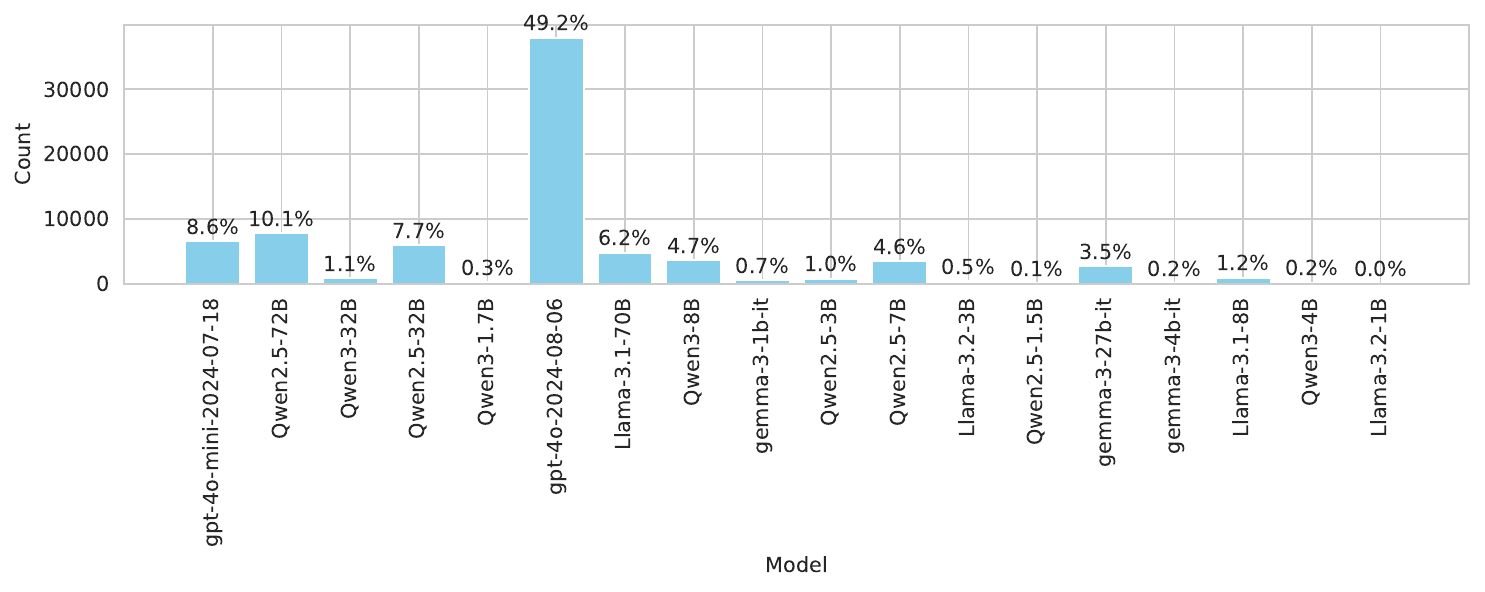}
    \caption{Frequency of model selection when using best-performing strategy. In case of ties, models are selected in order of increasing LLM hacking risk, i.e., GPT-4o first, see Table~\ref{tab:PAPER_llm_hacking_risk_by_model_task_averaged}.}
    \label{fig:model_selection_best_models}
\end{figure*}

Figure~\ref{fig:model_selection_best_models} reveals the gains from sophisticated model selection (see Figure~\ref{fig:model_selection_comparison} in Section~\ref{ssec:MitigatingLLMhackingrisk}): GPT-4o is only the best selected in about 49\% of cases.

Figure~\ref{fig:accuracy_vs_risk_correlation_all_tasks} shows that the negative relationship between performance and LLM hacking risk is significant even for a performance proxy measured only on the small human annotation sample.

\subsection{Robustness checks for ground truth uncertainty}
\label{app:outcome_of_interest}

Our primary analysis evaluates whether conclusions drawn from LLM-based annotations differ from those using ground truth, operationalized as differences in statistical significance at $\alpha = 0.05$ (see Section~\ref{ssec:Downstream_statistical_analysis}).
To prevent conflating sampling noise with systematic differences, we conduct two robustness checks: (1) testing whether LLM and ground truth effect estimates are statistically distinguishable, and (2) accounting for sampling uncertainty in ground truth tests using empirical Bayes methods.

\subsubsection{Testing statistical distinguishability of effect estimates}

For each hypothesis $h$, we assess whether the two effects (from the ground truth regression and the LLM-informed regression) are statistically distinguishable by fitting a pooled logistic model on a stacked dataset,
\begin{equation}
\text{logit}(P(y=1\mid x,s))=\alpha+\beta x+\gamma s+\delta (x\times s),
\end{equation}
where $s\in\{0,1\}$ indicates the label source (human ground truth vs.\ LLM) and standard errors are clustered by observation to account for the correlation between ground truth and LLM annotations of the same text.
The null $\text{H}_0\!:\delta=0$ tests equality of the ground truth and LLM effects of $x$, independent of significance thresholds.
Accordingly, our Type I/II/S error terminology is descriptive of divergences in conclusions \emph{relative to the ground truth regression} and is not a claim about false discovery or power in the population.

Table~\ref{tab:statistically_distinguishable_estimators} shows that LLM and ground truth regression coefficients are statistically distinguishable in 35-42\% of all hypotheses across models.
Critically, this distinguishability varies systematically with error type.
When LLMs correctly match ground truth conclusions, only 21-28\% of coefficients are significantly different.
In contrast, when LLM-informed regression conclusions differ, the coefficients are much more likely to be distinguishable: 58-66\% for LLM hacking cases overall.
Type S errors show the highest distinguishability rates, confirming these represent genuine estimation differences rather than borderline significance decisions.

These results validate that our error classification captures substantive differences in how LLMs process information compared to human annotations.
The high distinguishability rates for error cases demonstrate that LLM failures typically stem from systematically different effect estimates rather than minor perturbations around significance thresholds.
Conversely, the lower distinguishability rates for correct classifications suggest that when LLMs-informed regression conclusions are correct, the estimates might still be off, explaining the high Type M risk (see Table~\ref{tab:PAPER_llm_hacking_risk_by_model_task_averaged}).

\input{tables/statistically_distinguishable_estimators}

\subsubsection{Accounting for sampling uncertainty in ground truth regressions}

Our primary analysis treats sample-based regressions as ground truth, but these regressions are themselves subject to sampling error.
To quantify how this uncertainty affects our risk estimates, we use an empirical Bayes approach to estimate the posterior probability that each effect truly exists in the population.

For all hypotheses across all tasks, we apply the local false discovery rate (local FDR) method~\citep{Efron2004,efron2005local} to the $z$-statistics from ground truth regressions.
Using the R package \texttt{locfdr}, we estimate the empirical null distribution and compute 
\begin{equation}
\Pr(\text{effect exists} \mid z_h) = 1 - \text{localFDR}(z_h)
\end{equation}
for each hypothesis $h$.
This provides a probabilistic assessment of whether population effects exist, rather than binary classifications $S^{\text{GT}}_h$ based on $p < 0.05$.

\begin{figure*}[htb]
    \centering
    \includegraphics[height=0.9\textheight]{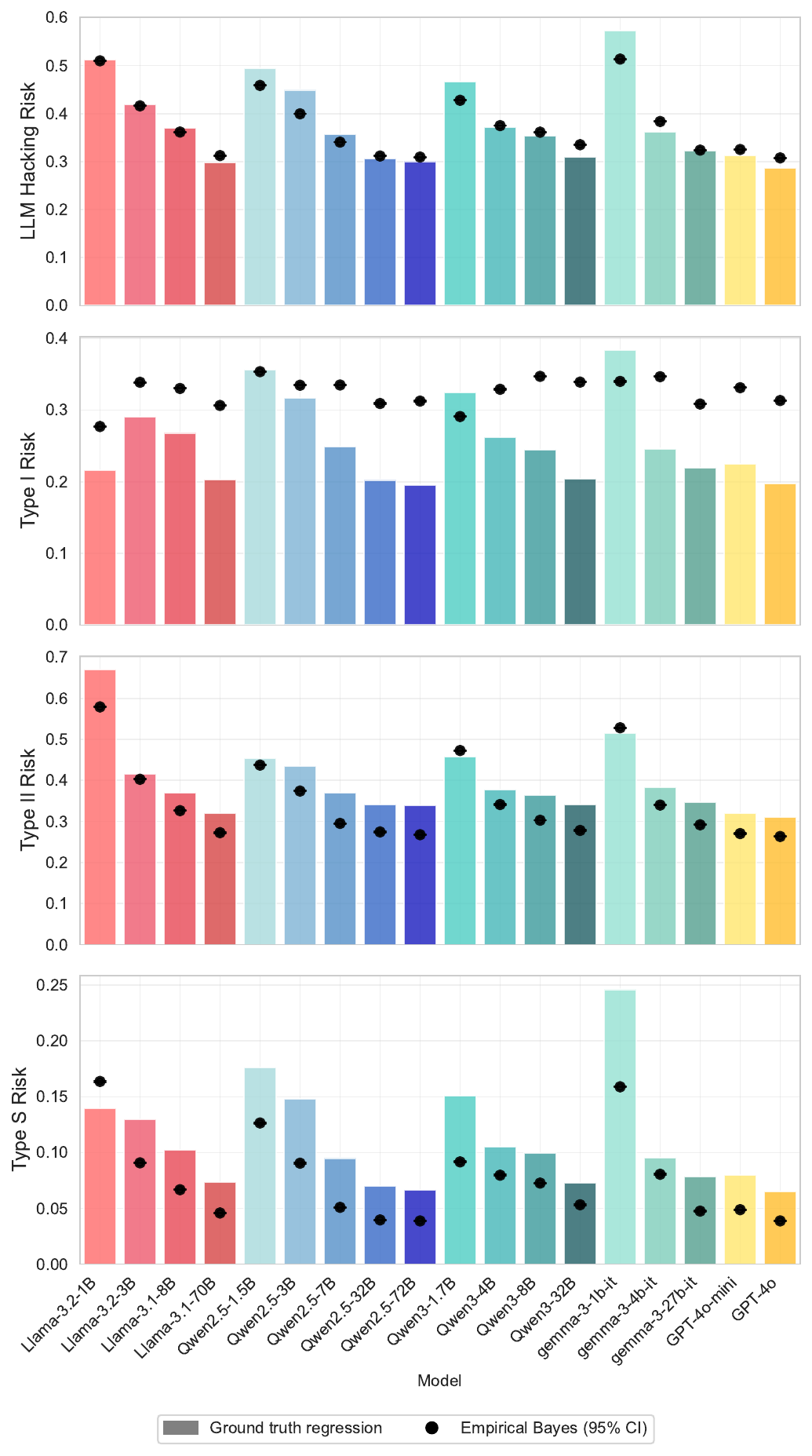}
    \caption{Risk estimates with uncertainty quantification via empirical Bayes.
    Bars show original risk estimates treating sample-based regressions as ground truth (as reported in Table~\ref{tab:PAPER_llm_hacking_risk_by_model_task_averaged}).
    Points with 95\% confidence intervals show risk estimates when accounting for uncertainty in ground truth regressions using local FDR to estimate posterior probabilities of effect existence, averaged across $1$,$000$ Monte Carlo samples.
    }
    \label{fig:risk_estimates_with_uncertainty_1000_bootstraps}
\end{figure*}

We then conduct a Monte Carlo simulation with $1$,$000$ iterations. In each iteration, we:
\begin{enumerate}
    \item Sample the `true' effect for each hypothesis from $\text{Bernoulli}(\Pr(\text{effect exists} \mid z_h))$.
    \item Update ground truth conclusions accordingly: hypotheses sampled as having no effect are labeled ``no difference,'' while those with effects retain their original direction.
    \item Recalculate Type I, II, S, and LLM hacking risk against these probabilistically-sampled ground truths.
\end{enumerate}
Figure~\ref{fig:risk_estimates_with_uncertainty_1000_bootstraps} shows that our original LLM hacking risk estimates, on average, are mostly robust to ground truth uncertainty effects.
However, with the empirical Bayes approach, the Type I risk increases, whereas
Type II and Type S risks decrease.
For example, Type I risk increases from $19.7\%$ to $31.3\%$ for GPT-4o.
This happens since the local FDR method produces more conservative probability estimates $\Pr(\text{effect exists})$ for hypotheses with ground truth $p$ value slightly below $0.05$.
If many effects classified as ``significant'' at $\alpha = 0.05$ have a low posterior probability of representing true population effects (i.e., are likely false positives in the original ground truth regression analysis), these now appear as Type 1 errors in the empirical Bayes approach.
At the same time, this also means there are fewer true effects to miss, which reduces Type II and Type S risk.

\clearpage

\input{tables/datasets_overview.tex}

\input{appendix_dataset_and_task_overview/appendix_datasets_tasks.tex}

\end{document}

%% file: macros.tex
\usepackage{xcolor}

\definecolor{DarkGreen}{rgb}{0.2,0.63,0.17}
\definecolor{DarkRed}{rgb}{0.89,0.1,0.11}
\definecolor{DarkBlue}{rgb}{0.12,0.47,0.71}
\definecolor{Gray}{rgb}{0.2,0.2,0.2}

\definecolor{DarkGreen}{RGB}{98,179,149}
\definecolor{DarkRed}{RGB}{206,86,69}
\definecolor{SABcolor}{RGB}{214,232,213}

\definecolor{bestcolor}{RGB}{214,232,213}
\definecolor{worstcolor}{RGB}{252,214,214}

\newcommand{\eat}[1]{}

%% file: tables/tasks_overview.tex
\begin{table}
\footnotesize
\caption{Overview of annotation tasks used in our experiments. $K.\alpha$ denotes Krippendorff's alpha ('-' indicates that $K.\alpha$ is unknown). See Appendix~\ref{app:Dataset and Task Overview} for detailed task descriptions.}
\label{tab:tasks_overview}
\renewcommand{\arraystretch}{1.2}
\begin{adjustbox}{max width=\textwidth}
\begin{tabular}{p{2.7cm}p{3.5cm}p{1.9cm}p{1.5cm}p{2.68cm}p{2cm}p{2cm}p{1cm}}
\toprule
Dataset Name & Task Name & Nr. of Datapoints: Used/Total & Nr. of Ground Truth Classes & Nr. of Prompts (\% paraphrases) (\% zero-shot) & Nr. of Groupings Total (\% original) & Nr. of Ground Truth Annotators Per Datapoint & $K.\alpha$ \\
\midrule
\multirow{6}{*}{\hyperref[app:datasetname:essay]{essay}~\citep{UniversityofLondonNational_Child_Development_Study2024}} & \hyperref[app:taskname:essaydomestic]{essay\_domestic} & 489/489 & 8 & 5 (80\%) (100\%) & 128 (57.0\%) & 1 & - \\
 & \hyperref[app:taskname:essayhousewife]{essay\_housewife} & 489/489 & 2 & 5 (80\%) (100\%) & 92 (79.3\%) & 1 & - \\
 & \hyperref[app:taskname:essayliving]{essay\_living} & 489/489 & 9 & 5 (80\%) (100\%) & 131 (55.7\%) & 1 & - \\
 & \hyperref[app:taskname:essaylocation]{essay\_location} & 489/489 & 5 & 5 (80\%) (100\%) & 110 (66.4\%) & 1 & - \\
 & \hyperref[app:taskname:essaynarrative]{essay\_narrative} & 489/489 & 3 & 5 (80\%) (100\%) & 98 (74.5\%) & 1 & - \\
 & \hyperref[app:taskname:essayworksocial]{essay\_worksocial} & 489/489 & 2 & 5 (80\%) (100\%) & 92 (79.3\%) & 1 & - \\
\myhdashline
\hyperref[app:datasetname:ideologynews]{ideology\_news}~\citep{baly-etal-2020-detect} & \hyperref[app:taskname:ideologynews]{ideology\_news} & 9968/37554 & 3 & 5 (60\%) (100\%) & 58 (56.9\%) & 1 & - \\
\myhdashline
\multirow{3}{*}{\hyperref[app:datasetname:manifestosuk]{manifestos\_uk}~\citep{benoit2016crowd}} & \hyperref[app:taskname:manifestoseconideology]{manifestos\_econ\_ideology} & 2463/2463 & 2 & 5 (40\%) (100\%) & 28 (32.1\%) & 5.6 & 0.71 \\
 & \hyperref[app:taskname:manifestosissue]{manifestos\_issue} & 9983/11620 & 2 & 5 (40\%) (80\%) & 28 (32.1\%) & 6.37 & 0.86 \\
 & \hyperref[app:taskname:manifestossocialideology]{manifestos\_social\_ideology} & 2091/2091 & 2 & 5 (20\%) (80\%) & 28 (32.1\%) & 4.98 & 0.75 \\
\myhdashline
\hyperref[app:datasetname:tone]{tone}~\citep{carlson2017pairwise} & \hyperref[app:taskname:tone]{tone} & 859/859 & 3 & 5 (60\%) (80\%) & 25 (0\%) & 1 & - \\
\myhdashline
\hyperref[app:datasetname:politenessstack]{politeness\_stack}~\citep{danescu-niculescu-mizil-etal-2013-computational} & \hyperref[app:taskname:politenessstack]{politeness\_stack} & 3298/3298 & 2 & 5 (60\%) (100\%) & 49 (61.2\%) & 5 & 0.26 \\
\myhdashline
\hyperref[app:datasetname:politenesswiki]{politeness\_wiki}~\citep{danescu-niculescu-mizil-etal-2013-computational} & \hyperref[app:taskname:politenesswiki]{politeness\_wiki} & 2171/2171 & 2 & 5 (60\%) (100\%) & 44 (56.8\%) & 5 & 0.38 \\
\myhdashline
\hyperref[app:datasetname:emotion]{emotion}~\citep{demszky-etal-2020-goemotions} & \hyperref[app:taskname:emotion]{emotion} & 7276/7276 & 26 & 5 (60\%) (100\%) & 123 (37.4\%) & 3.64 & 1 \\
\myhdashline
\hyperref[app:datasetname:topic]{topic}~\citep{egami2023using} & \hyperref[app:taskname:topic]{topic} & 10000/10000 & 2 & 5 (20\%) (60\%) & 33 (42.4\%) & 1 & - \\
\myhdashline
\multirow{3}{*}{\hyperref[app:datasetname:hatespeech]{hatespeech}~\citep{elsherief-etal-2021-latent}} & \hyperref[app:taskname:hatespeechexplicit]{hatespeech\_explicit} & 9998/21476 & 3 & 5 (40\%) (80\%) & 25 (0\%) & 3 & 1 \\
 & \hyperref[app:taskname:hatespeechimplicit]{hatespeech\_implicit} & 6135/6135 & 6 & 5 (40\%) (80\%) & 43 (0\%) & 1-3 & 1 \\
 & \hyperref[app:taskname:hatespeechtarget]{hatespeech\_target} & 5475/5475 & 26 & 5 (40\%) (80\%) & 94 (21.3\%) & 2 & 1 \\
\myhdashline
\hyperref[app:datasetname:issuesurvey]{issue\_survey}~\citep{fieldhouse2024british} & \hyperref[app:taskname:issuesurvey]{issue\_survey} & 9922/101930 & 48 & 5 (20\%) (80\%) & 364 (78.8\%) & 1-2 & 1 \\
\myhdashline
\hyperref[app:datasetname:misinfo]{misinfo}~\citep{gabriel-etal-2022-misinfo} & \hyperref[app:taskname:misinfo]{misinfo} & 9991/25164 & 2 & 5 (80\%) (100\%) & 35 (45.7\%) & 1 & - \\
\myhdashline
\multirow{2}{*}{\hyperref[app:datasetname:news]{news}~\citep{gilardi2023pnas}} & \hyperref[app:taskname:framesInews]{framesI\_news} & 326/326 & 3 & 7 (0\%) (85.7\%) & 25 (0\%) & 2 & 1 \\
 & \hyperref[app:taskname:relevancenews]{relevance\_news} & 1599/1599 & 2 & 7 (0\%) (85.7\%) & 19 (0\%) & 2 & 1 \\
\myhdashline
\multirow{2}{*}{\hyperref[app:datasetname:tweets17]{tweets17}~\citep{gilardi2023pnas}} & \hyperref[app:taskname:framesIItweets17]{framesII\_tweets17} & 569/569 & 15 & 5 (0\%) (80\%) & 74 (0\%) & 2 & 1 \\
 & \hyperref[app:taskname:relevancetweets17]{relevance\_tweets17} & 1856/1856 & 2 & 5 (0\%) (80\%) & 19 (0\%) & 2 & 1 \\
\myhdashline
\multirow{2}{*}{\hyperref[app:datasetname:tweets23]{tweets23}~\citep{gilardi2023pnas}} & \hyperref[app:taskname:framesItweets23]{framesI\_tweets23} & 150/150 & 3 & 7 (0\%) (85.7\%) & 25 (0\%) & 2 & 1 \\
 & \hyperref[app:taskname:relevancetweets23]{relevance\_tweets23} & 411/411 & 2 & 7 (0\%) (85.7\%) & 19 (0\%) & 2 & 1 \\
\myhdashline
\multirow{5}{*}{\hyperref[app:datasetname:tweets]{tweets}~\citep{gilardi2023pnas}} & \hyperref[app:taskname:framesIItweets]{framesII\_tweets} & 311/311 & 10 & 5 (0\%) (80\%) & 64 (0\%) & 2 & 1 \\
 & \hyperref[app:taskname:framesItweets]{framesI\_tweets} & 593/593 & 3 & 7 (0\%) (85.7\%) & 25 (0\%) & 2 & 1 \\
 & \hyperref[app:taskname:relevancetweets]{relevance\_tweets} & 1934/1934 & 2 & 7 (0\%) (85.7\%) & 19 (0\%) & 2 & 1 \\
 & \hyperref[app:taskname:stancetweets]{stance\_tweets} & 557/557 & 3 & 7 (0\%) (85.7\%) & 25 (0\%) & 2 & 1 \\
 & \hyperref[app:taskname:topictweets]{topic\_tweets} & 611/611 & 5 & 5 (0\%) (80\%) & 37 (0\%) & 2 & 1 \\
\myhdashline
\hyperref[app:datasetname:stanceclimate]{stance\_climate}~\citep{luo-etal-2020-detecting} & \hyperref[app:taskname:stanceclimate]{stance\_climate} & 1793/1793 & 3 & 5 (60\%) (100\%) & 25 (0\%) & 8 & - \\
\myhdashline
\hyperref[app:datasetname:manifestos]{manifestos}~\citep{merz2016manifesto,Lehmann2024} & \hyperref[app:taskname:manifestosissuesdetailed]{manifestos\_issues\_detailed} & 9986/1239149 & 46 & 5 (40\%) (100\%) & 212 (64.6\%) & 1 & - \\
\myhdashline
\hyperref[app:datasetname:factuality]{factuality}~\citep{min-etal-2023-factscore} & \hyperref[app:taskname:factuality]{factuality} & 9965/15225 & 3 & 5 (20\%) (80\%) & 48 (47.9\%) & 1-2 & 1 \\
\myhdashline
\hyperref[app:datasetname:fakenews]{fakenews}~\citep{shu2020fakenewsnet,Wu2024SheepDog} & \hyperref[app:taskname:fakenews]{fakenews} & 7954/7954 & 2 & 5 (80\%) (100\%) & 34 (44.1\%) & 1 & - \\
\myhdashline
\hyperref[app:datasetname:ideologytweets]{ideology\_tweets}~\citep{tornberg2024large} & \hyperref[app:taskname:ideologytweets]{ideology\_tweets} & 4099/4099 & 2 & 5 (60\%) (100\%) & 41 (53.7\%) & 1 & - \\
\myhdashline
\hyperref[app:datasetname:humor]{humor}~\citep{weller-seppi-2019-humor} & \hyperref[app:taskname:humor]{humor} & 9999/15817 & 2 & 5 (60\%) (100\%) & 22 (13.6\%) & 1 & - \\
\midrule
\textbf{21 datasets} & \textbf{37 tasks} & \textbf{145277/1533400} & \textbf{266} & \textbf{199 (36.2\%) (89.4\%)} & \textbf{2361 (48.1\%)} & \textbf{2.2 on average} & \textbf{0.91 on average} \\
\bottomrule
\end{tabular}
\end{adjustbox}
\end{table}

%% file: tables/PAPER_llm_hacking_risk_by_model_task_averaged.tex
% Requires: \usepackage{array,booktabs,arydshln}
% \begin{table}[htbp]
\small
\centering
\caption{Empirical LLM hacking risk and error types by model (lower is better). Highlights indicate \colorbox{bestcolor}{best} and \colorbox{worstcolor}{worst} risk values.}
\begin{tabular}{lrrrrr}
\toprule
\textbf{Model} & \textbf{LLM Hacking Risk} & \textbf{Type I Risk} & \textbf{Type II Risk} & \textbf{Type S Risk} & \textbf{Type M Risk} \\
\midrule
Llama-3.2-1B & \colorbox{worstcolor}{0.503} & 0.258 & \colorbox{worstcolor}{0.591} & \colorbox{worstcolor}{0.157} & \colorbox{worstcolor}{0.771} \\
Llama-3.2-3B & 0.422 & 0.310 & 0.438 & 0.095 & 0.572 \\
Llama-3.1-8B & 0.366 & 0.293 & 0.370 & 0.069 & 0.482 \\
Llama-3.1-70B & 0.316 & \colorbox{bestcolor}{0.257} & 0.324 & 0.050 & 0.415 \\
\hdashline
Qwen2.5-1.5B & 0.459 & \colorbox{worstcolor}{0.327} & 0.458 & 0.132 & 0.695 \\
Qwen2.5-3B & 0.405 & 0.308 & 0.409 & 0.094 & 0.601 \\
Qwen2.5-7B & 0.350 & 0.299 & 0.344 & 0.056 & 0.488 \\
Qwen2.5-32B & 0.315 & 0.263 & 0.324 & 0.043 & 0.447 \\
Qwen2.5-72B & 0.318 & 0.269 & 0.324 & \colorbox{bestcolor}{0.042} & 0.422 \\
\hdashline
Qwen3-1.7B & 0.429 & 0.267 & 0.499 & 0.091 & 0.585 \\
Qwen3-4B & 0.376 & 0.293 & 0.378 & 0.081 & 0.612 \\
Qwen3-8B & 0.369 & 0.314 & 0.349 & 0.076 & 0.502 \\
Qwen3-32B & 0.346 & 0.303 & 0.333 & 0.056 & 0.473 \\
\hdashline
Gemma-3-1b-it & 0.502 & 0.314 & 0.533 & \colorbox{worstcolor}{0.157} & 0.725 \\
Gemma-3-4b-it & 0.385 & 0.314 & 0.376 & 0.081 & 0.566 \\
Gemma-3-27b-it & 0.332 & 0.268 & 0.345 & 0.050 & 0.449 \\
\hdashline
GPT-4o-mini & 0.331 & 0.287 & 0.321 & 0.053 & 0.494 \\
GPT-4o & \colorbox{bestcolor}{0.312} & 0.263 & \colorbox{bestcolor}{0.317} & 0.043 & \colorbox{bestcolor}{0.405} \\
\midrule
Baseline 1 (random conclusions)$\!\!\!\!\!\!\!\!\!\!\!\!\!\!\!\!\!\!\!\!\!\!\!\!\!\!\!\!\!\!\!\!\!$ & 0.625 & 0.499 & 0.500 & 0.250 & - \\
Baseline 2 (random labels)$\!\!\!\!\!\!\!\!\!\!\!\!\!\!\!\!\!\!\!\!\!\!\!\!\!\!\!\!\!\!\!\!\!$ & 0.526 & 0.080 & 0.930 & 0.043 & - \\
\bottomrule
\end{tabular}
% \end{table}

%% file: tables/literature_review_table.tex
{{\scriptsize
\begin{longtable}{lcccll}
\caption{93 relevant papers from a systematic review of LLM usage recommendations in computational social science literature. 
Papers are sorted by their recommendation type (LLM Rec.). 
\CIRCLE~indicates explicit recommendation to use LLMs (\LEFTcircle~indicates conditional recommendation), 
\Circle~indicates recommendation against using LLMs, 
\cmark~indicates yes/present, 
\xmark~indicates no/absent, 
and `--' indicates not applicable or not reported. We omit `instruct' indications for model names. We use the following abbreviations for prompt types: zero-shot ($0$), few-shot (n), chain-of-thought (COT).
}
\label{tab:literature_review} \\
\toprule
\textbf{Paper} & \textbf{LLM Rec.} & \textbf{Validation} & \textbf{Dataset} & \textbf{Models Tested} & \textbf{Prompt Type} \\
\midrule
\endfirsthead

\multicolumn{6}{c}%
{{\tablename\ \thetable{} -- continued from previous page}} \\
\toprule
\textbf{Paper} & \textbf{LLM Rec.} & \textbf{Validation} & \textbf{Dataset} & \textbf{Models tested} & \textbf{Prompt Type} \\
\midrule
\endhead

\midrule
\multicolumn{6}{r}{{Continued on next page}} \\
\endfoot

\bottomrule
\endlastfoot

\citep{gilardi2023pnas} & \CIRCLE & \xmark & \cmark & gpt-3.5-turbo & $0$ \\
\citep{Mizumoto2023} & \CIRCLE & \cmark & \cmark & text-davinci-003 & $0$ \\
\citep{chiang-lee-2023-large} & \CIRCLE & \cmark & \cmark & text-curie-001,text-davinci-003,ChatGPT,t0 & $0$ \\
\citep{wu2023large} & \CIRCLE & \xmark & \xmark & ChatGPT & $0$ \\
\citep{Wang2024Human} & \CIRCLE & \cmark & \cmark & text-davinci-003 & n \\
\citep{chae2023large} & \CIRCLE & \cmark & \cmark & many different models & $0$,n \\
\citep{chew2023llm} & \CIRCLE & \cmark & \cmark & gpt-3.5 & $0$ \\
\citep{Rathje2024pnas} & \CIRCLE & \cmark & \cmark & GPT-3.5 Turbo,GPT-4,GPT-4 Turbo  & $0$,n \\
\citep{krugmann2024sentiment} & \CIRCLE & \xmark & \cmark & GPT-3.5,GPT-4,Llama 2 & $0$,n \\
\citep{LeMenspnas2309350120} & \CIRCLE & \cmark & \cmark & GPT-3,GPT-3.5Turbo & $0$ \\
\citep{Linegar2023} & \CIRCLE & \cmark & \xmark & -- & -- \\
\citep{Lu_2024} & \CIRCLE & \xmark & \xmark & Llama3 & -- \\
\citep{NEURIPS2024_937ae0e8} & \CIRCLE & \xmark & \cmark & many different models & $0$ \\
\citep{Gu_Zhu_Ye_2024} & \CIRCLE & \xmark & \cmark & many different models & $0$,n \\
\citep{hussain2024tutorial} & \CIRCLE & \cmark & \xmark & -- & -- \\
\citep{chhun-etal-2022-human} & \CIRCLE & \xmark & \cmark & GPT-2 & -- \\
\citep{tornberg2023use} & \CIRCLE & \cmark & \xmark & GPT-4 & $0$,COT \\
\citep{pmlr-v162-zhong22a} & \CIRCLE & \xmark & \cmark & GPT-3 Curie (13B),GPT-3 Davinci (175B) & $0$ \\
\citep{bhattacharyya-etal-2023-video} & \CIRCLE & \xmark & \cmark & GPT-3.5,Flan-t5 & $0$ \\
\citep{Li2024HOT} & \CIRCLE & \xmark & \cmark & gpt-3.5-turbo & $0$ \\
\citep{roy-etal-2022-towards} & \CIRCLE & \xmark & \cmark & GPT-J-6B & n \\
\citep{Törnberg2024Large} & \CIRCLE & \cmark & \cmark & gpt-4 & $0$ \\
\citep{laskar2023cqsumdp} & \CIRCLE & \xmark & \cmark & ChatGPT & $0$ \\
\citep{karjus2024evolving} & \CIRCLE & \xmark & \xmark & GPT4 & $0$ \\
\citep{peskine-etal-2023-definitions} & \CIRCLE & \xmark & \cmark & GPT-3 & $0$ \\
\citep{Choi2024Automated} & \CIRCLE & \xmark & \cmark & GPT-3.5-Turbo,GPT-4,Llama 2 & $0$,n,COT \\
\citep{Wan2024TnT} & \CIRCLE & \cmark & \cmark & GPT-4,GPT-3.5 & -- \\
\citep{Heseltine2024} & \CIRCLE & \xmark & \cmark & GPT-4 & $0$ \\
\citep{Mets2024} & \CIRCLE & \xmark & \cmark & GPT-3.5 & -- \\
\citep{Mellon2024Issue} & \CIRCLE & \cmark & \cmark & Claude 1.3,Claude 2,GPT-4,GPT-3.5-turbo,PaLM-2,Llama 2 & -- \\
\citep{alizadeh2025open} & \CIRCLE & \cmark & \cmark & gpt-4,gpt-3.5-turbo,LLaMA-1,LLaMA-2 & $0$,n,COT \\
\citep{Fatemi2023} & \CIRCLE & \xmark & \cmark & gpt-3.5-Turbo & $0$ \\
\citep{rouzegar-makrehchi-2024-enhancing} & \CIRCLE & \cmark & \cmark & GPT-3.5 & $0$,n \\
\citep{He2024If} & \CIRCLE & \xmark & \cmark & GPT4 & $0$ \\
\citep{li2024political} & \CIRCLE & \xmark & \xmark & -- & -- \\
\citep{hoes2023leveraging} & \CIRCLE & \xmark & \cmark & gpt-3.5-turbo & $0$ \\
\citep{Briggs2025} & \CIRCLE & \xmark & \cmark & GPT-4o & $0$ \\
\citep{lupo2023towards} & \CIRCLE & \xmark & \cmark & text-da-vinci-003,GPT-4,Llama-2-13b,Mixtral-8x7b,GPT-SW3-20b & $0$ \\
\citep{pangakis2025keeping} & \CIRCLE & \cmark & \cmark & GPT-4 & n \\
\citep{li-etal-2023-coannotating} & \CIRCLE & \cmark & \cmark & gpt-3.5-turbo & $0$ \\
\citep{feuerriegel2025using} & \CIRCLE & \cmark & \xmark & -- & -- \\
\citep{Halterman_Keith_2025} & \CIRCLE & \cmark & \cmark & Mistral (7B \& NeMo),Llama-3.1-8B,OLMo-7B & $0$ \\
\citep{Zhou_Xu_Wang_Lu_Gao_Ai_2025} & \CIRCLE & \xmark & \cmark & ChatGPT & $0$ \\
\citep{Fu2024} & \CIRCLE & \xmark & \cmark & GPT-3.5,GPT-4 & $0$ \\
\citep{dosSantos2024} & \CIRCLE & \xmark & \cmark & GPT-3.5-turbo & $0$,n \\
\citep{qiu2024interactive} & \CIRCLE & \xmark & \cmark & GPT4 & $0$ \\
\citep{zhang-etal-2023-mitigating} & \CIRCLE & \cmark & \cmark & GPT-3,ChatGPT,text-davinci-003 & $0$,n \\
\citep{egami2023using} & \LEFTcircle & \cmark & \cmark & flan-ul2 & $0$,n \\
\citep{weber2023evaluation} & \LEFTcircle & \cmark & \cmark & neural-chat-7b-v3-2,Starling-LM-7B-alpha,openchat\_3.5,zephyr & $0$,n,COT \\
\citep{pavlovic-poesio-2024-effectiveness} & \LEFTcircle & \xmark & \cmark & GPT 3.5-turbo & $0$ \\
\citep{Kholodna2024} & \LEFTcircle & \xmark & \cmark & many different models & n \\
\citep{mu-etal-2024-navigating} & \LEFTcircle & \cmark & \cmark & GPT-3.5-turbo,LLaMA-OA & $0$,n \\
\citep{khondaker-etal-2023-gptaraeval} & \LEFTcircle & \xmark & \cmark & BLOOMZ,ChatGPT,MARBERT & n \\
\citep{karjus2025machine} & \LEFTcircle & \cmark & \xmark & GPT-3.5,GPT-4 & $0$ \\
\citep{Kumar_AbuHashem_Durumeric_2024} & \LEFTcircle & \cmark & \cmark & GPT-3,GPT-3.5,GPT-4,Gemini Pro,LLAMA 2 & $0$,COT \\
\citep{moller-etal-2024-parrot} & \LEFTcircle & \cmark & \cmark &  GPT-4,Llama-2 70B & $0$ \\
\citep{pangakis2023automated} & \LEFTcircle & \cmark & \cmark & GPT-4 & -- \\
\citep{liang2023holistic} & \LEFTcircle & \cmark & \cmark & many different models & n \\
\citep{ziems-etal-2024-large} & \LEFTcircle & \cmark & \cmark & many different models & $0$,n \\
\citep{laskar-etal-2023-systematic} & \LEFTcircle & \cmark & \cmark & gpt-3.5-turbo & $0$,n \\
\citep{yu2023open} & \LEFTcircle & \cmark & \cmark & Llama 2 (13b \& 70b),gpt-3.5-turbo,gpt-4 & $0$,n \\
\citep{davidson2024start} & \LEFTcircle & \xmark & \xmark & -- & -- \\
\citep{OrnsteinBlasingameTruscott2025} & \LEFTcircle & \cmark & \cmark & GPT-3,GPT-4 & $0$,n \\
\citep{Jiang2024Disinformation} & \LEFTcircle & \xmark & \cmark & GPT-4,GPT3.5 & $0$ \\
\citep{maehlum-etal-2024-difficult} & \LEFTcircle & \cmark & \cmark & ChatNorT5,NorMistral & $0$,n \\
\citep{kim-etal-2024-meganno} & \LEFTcircle & \cmark & \xmark & -- & -- \\
\citep{movva-etal-2024-annotation} & \LEFTcircle & \cmark & \cmark & GPT-4 & $0$ \\
\citep{barrie2024replication} & \LEFTcircle & \cmark & \cmark & GPT4,Gemini,Llama & $0$ \\
\citep{moghimifar2024modelling} & \LEFTcircle & \xmark & \cmark & GPT-4 & $0$ \\
\citep{Rogers2024} & \LEFTcircle & \cmark & \xmark & GPT-3.5,Claude-2 & $0$ \\
\citep{gligoric-etal-2025-unconfident} & \LEFTcircle & \cmark & \cmark & GPT-4o,GPT-3.5 & $0$,n \\
\citep{MISIEJUK2024100216} & \LEFTcircle & \cmark & \cmark & text-davinci-003,GPT4 & $0$,n \\
\citep{XU2024103665} & \LEFTcircle & \xmark & \xmark & -- & -- \\
\citep{Barrio2023Framing} & \LEFTcircle & \xmark & \cmark & GPT-3 & -- \\
\citep{tornberg2024best} & \LEFTcircle & \cmark & \xmark & -- & -- \\
\citep{jin-etal-2024-mm} & \LEFTcircle & \cmark & \cmark & LLaVA-v1.5,BLIP2,InstructBLIP,LLaMA-Adapter-v2 & $0$ \\
\citep{roy-etal-2023-probing} & \LEFTcircle & \cmark & \cmark & GPT-3.5,text-davinci,flan-t5-large & $0$,n \\
\citep{ni-etal-2024-afacta} & \LEFTcircle & \cmark & \cmark & GPT-3.5,GPT-4 & $0$ \\
\citep{Mishra2024} & \LEFTcircle & \xmark & \cmark & GPT-3.5-turbo & $0$ \\
\citep{Liyanage2024} & \LEFTcircle & \cmark & \cmark & GPT-4 & $0$,n,COT \\
\citep{Stuhler2025} & \LEFTcircle & \cmark & \cmark & Llama-3 (8B \& 70B) & $0$,n \\
\citep{yao-etal-2023-beyond} & \LEFTcircle & \cmark & \cmark & FLAN-T5-XL & n \\
\citep{barrie2024prompt} & \LEFTcircle & \cmark & \xmark & gpt-3.5-turbo & $0$ \\
\citep{egami2024using} & \LEFTcircle & \cmark & \cmark & gpt-4o,gpt-3.5-turbo,DeepSeek-r1 & $0$ \\
\citep{li-2024-human} & \LEFTcircle & \xmark & \cmark & GPT-4,gemini-1.5-pro & -- \\
\citep{NEURIPS2023_8b8a7960} & \Circle & \xmark & \cmark & ChatGPT,GPT4,Vicuna 33B & COT \\
\citep{chan2024chateval} & \Circle & \cmark & \cmark & GPT-4. GPT-3.5-turbo & -- \\
\citep{liu2023trustworthy} & \Circle & \cmark & \xmark & many different models & -- \\
\citep{weerasooriya-etal-2023-vicarious} & \Circle & \cmark & \cmark & gpt3.5turbo,v3.5 & $0$ \\
\citep{Ashwin2025} & \Circle & \cmark & \cmark & gpt-4o-mini,gpt-3.5-turbo,llama 3 8b,llama 2 13b & n,COT \\
\citep{bucher2024fine} & \Circle & \xmark & \cmark & GPT-3.5,GPT-4,Claude Opus 3,BART-LRG & -- \\
\citep{Wu2024Fake} & \Circle & \xmark & \cmark & GPT3.5,Llama2-13B,InstructGPT & $0$ \\
\citep{felkner-etal-2024-gpt} & \Circle & \cmark & \cmark & GPT-3.5-Turbo & $0$ \\
\end{longtable}
}}

%% file: tables/PAPER_llm_hacking_risk_by_task_model.tex
% Requires: \usepackage{graphicx,array,booktabs,arydshln}
% \begin{table}[htbp]
\small
\centering
\caption{LLM hacking risk by task and model. `-' indicates that none of the prompts passed our threshold of producing valid annotations in at least 99\% of cases.}
\renewcommand{\arraystretch}{1.2}
\begin{adjustbox}{max width=\textwidth}
\rowcolors{2}{gray!20}{white}
\begin{tabular}{lrrrr:rrrrr:rrrr:rrr:rr}
\toprule
Task & \rotatebox{90}{Llama-3.2-1B} & \rotatebox{90}{Llama-3.2-3B} & \rotatebox{90}{Llama-3.1-8B} & \rotatebox{90}{Llama-3.1-70B} & \rotatebox{90}{Qwen2.5-1.5B} & \rotatebox{90}{Qwen2.5-3B} & \rotatebox{90}{Qwen2.5-7B} & \rotatebox{90}{Qwen2.5-32B} & \rotatebox{90}{Qwen2.5-72B} & \rotatebox{90}{Qwen3-1.7B} & \rotatebox{90}{Qwen3-4B} & \rotatebox{90}{Qwen3-8B} & \rotatebox{90}{Qwen3-32B} & \rotatebox{90}{Gemma-3-1b-it} & \rotatebox{90}{Gemma-3-4b-it} & \rotatebox{90}{Gemma-3-27b-it} & \rotatebox{90}{GPT-4o-mini} & \rotatebox{90}{GPT-4o} \\
\midrule
politeness\_wiki & 0.39 & 0.25 & 0.18 & 0.22 & 0.27 & 0.28 & 0.30 & 0.26 & 0.18 & 0.35 & 0.21 & 0.29 & 0.25 & 0.28 & 0.21 & 0.17 & 0.16 & 0.18 \\
politeness\_stack & 0.46 & 0.54 & 0.36 & 0.28 & 0.48 & 0.37 & 0.35 & 0.33 & 0.29 & 0.44 & 0.34 & 0.32 & 0.32 & 0.40 & 0.43 & 0.22 & 0.27 & 0.32 \\
stance\_climate & 0.58 & 0.39 & 0.33 & 0.30 & 0.39 & 0.32 & 0.22 & 0.24 & 0.22 & 0.39 & 0.30 & 0.33 & 0.31 & 0.46 & 0.30 & 0.33 & 0.17 & 0.23 \\
ideology\_news & 0.58 & 0.56 & 0.54 & 0.47 & 0.53 & 0.53 & 0.52 & 0.48 & 0.50 & 0.59 & 0.55 & 0.51 & 0.54 & 0.59 & 0.57 & 0.44 & 0.57 & 0.49 \\
fakenews & 0.52 & 0.46 & 0.42 & 0.42 & 0.37 & 0.37 & 0.47 & 0.46 & 0.46 & 0.44 & 0.49 & 0.46 & 0.51 & - & 0.28 & 0.41 & 0.45 & 0.47 \\
factuality & 0.57 & 0.61 & 0.66 & 0.62 & 0.66 & 0.57 & 0.60 & 0.59 & 0.56 & 0.59 & 0.65 & 0.63 & 0.63 & 0.62 & 0.62 & 0.60 & 0.59 & 0.44 \\
hatespeech\_explicit & 0.46 & 0.44 & 0.30 & 0.33 & 0.43 & 0.42 & 0.38 & 0.30 & 0.33 & 0.40 & 0.38 & 0.43 & 0.33 & 0.56 & 0.41 & 0.33 & 0.37 & 0.26 \\
hatespeech\_implicit & 0.51 & 0.46 & 0.40 & 0.36 & 0.54 & 0.43 & 0.40 & 0.39 & 0.35 & 0.50 & 0.41 & 0.39 & 0.34 & 0.57 & 0.38 & 0.34 & 0.36 & 0.34 \\
hatespeech\_target & 0.51 & 0.47 & 0.40 & 0.38 & 0.48 & 0.50 & 0.36 & 0.37 & 0.39 & 0.44 & 0.42 & 0.42 & 0.38 & 0.55 & 0.40 & 0.42 & 0.38 & 0.39 \\
emotion & 0.60 & 0.38 & 0.36 & 0.30 & 0.47 & 0.40 & 0.34 & 0.32 & 0.33 & 0.39 & 0.35 & 0.33 & 0.32 & 0.59 & 0.35 & 0.32 & 0.32 & 0.31 \\
essay\_living & 0.52 & 0.46 & 0.35 & 0.37 & 0.54 & 0.44 & 0.39 & 0.29 & 0.29 & 0.50 & 0.44 & 0.36 & 0.32 & 0.61 & 0.44 & 0.33 & 0.40 & 0.35 \\
essay\_housewife & 0.53 & 0.47 & 0.43 & 0.31 & 0.53 & 0.38 & 0.36 & 0.35 & 0.31 & 0.57 & 0.33 & 0.33 & 0.36 & 0.55 & 0.44 & 0.38 & 0.33 & 0.35 \\
essay\_worksocial & 0.57 & 0.36 & 0.48 & 0.26 & 0.47 & 0.41 & 0.25 & 0.32 & 0.35 & 0.43 & 0.32 & 0.30 & 0.33 & 0.46 & 0.31 & 0.30 & 0.33 & 0.36 \\
essay\_location & 0.53 & 0.50 & 0.47 & 0.53 & 0.53 & 0.43 & 0.50 & 0.55 & 0.50 & 0.45 & 0.44 & 0.37 & 0.47 & 0.59 & 0.49 & 0.49 & 0.47 & 0.52 \\
essay\_domestic & 0.54 & 0.45 & 0.45 & 0.30 & 0.49 & 0.50 & 0.38 & 0.28 & 0.27 & 0.44 & 0.42 & 0.50 & 0.32 & 0.58 & 0.43 & 0.31 & 0.36 & 0.28 \\
essay\_narrative & 0.51 & 0.38 & 0.41 & 0.41 & 0.47 & 0.38 & 0.41 & 0.35 & 0.38 & 0.41 & 0.39 & 0.44 & 0.38 & 0.50 & 0.44 & 0.50 & 0.37 & 0.40 \\
issue\_survey & 0.50 & 0.35 & 0.29 & 0.23 & 0.45 & 0.42 & 0.28 & 0.23 & 0.22 & 0.36 & 0.30 & 0.30 & 0.23 & 0.60 & 0.26 & 0.21 & 0.24 & 0.21 \\
humor & 0.37 & 0.33 & 0.12 & 0.04 & 0.13 & 0.14 & 0.10 & 0.03 & 0.15 & 0.21 & 0.10 & 0.09 & 0.16 & 0.35 & 0.19 & 0.05 & 0.07 & 0.02 \\
topic & 0.51 & 0.29 & 0.19 & 0.08 & 0.49 & 0.30 & 0.20 & 0.04 & 0.12 & 0.42 & 0.30 & 0.21 & 0.20 & 0.43 & 0.26 & 0.11 & 0.12 & 0.07 \\
ideology\_tweets & - & 0.58 & - & 0.39 & 0.58 & 0.63 & 0.44 & 0.28 & 0.50 & 0.53 & 0.64 & 0.54 & 0.66 & 0.66 & 0.50 & 0.44 & 0.49 & 0.30 \\
misinfo & 0.35 & 0.35 & 0.39 & 0.35 & 0.38 & 0.37 & 0.40 & 0.40 & 0.35 & 0.36 & 0.36 & 0.38 & 0.34 & 0.40 & 0.37 & 0.35 & 0.36 & 0.39 \\
tone & 0.25 & 0.20 & 0.23 & 0.24 & 0.18 & 0.20 & 0.21 & 0.25 & 0.21 & 0.21 & 0.19 & 0.21 & 0.28 & 0.31 & 0.28 & 0.23 & 0.21 & 0.25 \\
manifestos\_issue & 0.41 & 0.37 & 0.43 & 0.36 & 0.37 & 0.33 & 0.42 & 0.39 & 0.37 & 0.34 & 0.40 & 0.38 & 0.38 & 0.43 & 0.44 & 0.39 & 0.37 & 0.38 \\
manifestos\_econ\_ideology & 0.56 & 0.38 & 0.17 & 0.07 & 0.42 & 0.45 & 0.20 & 0.25 & 0.24 & 0.48 & 0.26 & 0.10 & 0.22 & 0.35 & 0.29 & 0.26 & 0.10 & 0.14 \\
manifestos\_social\_ideology & 0.31 & 0.34 & 0.30 & 0.21 & 0.47 & 0.39 & 0.27 & 0.23 & 0.21 & 0.34 & 0.39 & 0.26 & 0.21 & 0.29 & 0.30 & 0.28 & 0.27 & 0.30 \\
manifestos\_issues\_detailed & 0.58 & 0.55 & 0.47 & 0.40 & 0.60 & 0.54 & 0.48 & 0.42 & 0.41 & 0.58 & 0.49 & 0.45 & 0.42 & 0.65 & 0.45 & 0.42 & 0.43 & 0.41 \\
relevance\_tweets & 0.60 & 0.55 & 0.43 & 0.35 & 0.55 & 0.31 & 0.36 & 0.33 & 0.27 & 0.45 & 0.32 & 0.39 & 0.36 & 0.55 & 0.34 & 0.40 & 0.38 & 0.47 \\
framesI\_tweets & 0.58 & 0.40 & 0.38 & 0.34 & 0.53 & 0.46 & 0.28 & 0.31 & 0.32 & 0.52 & 0.48 & 0.40 & 0.32 & 0.57 & 0.46 & 0.39 & 0.34 & 0.27 \\
framesII\_tweets & 0.50 & 0.52 & 0.59 & 0.53 & 0.55 & 0.59 & 0.57 & 0.54 & 0.52 & 0.56 & 0.55 & 0.64 & 0.52 & 0.59 & 0.67 & 0.56 & 0.57 & 0.48 \\
stance\_tweets & 0.52 & 0.38 & 0.30 & 0.22 & 0.52 & 0.42 & 0.31 & 0.26 & 0.28 & 0.31 & 0.42 & 0.32 & 0.30 & 0.55 & 0.36 & 0.32 & 0.28 & 0.25 \\
topic\_tweets & 0.49 & 0.41 & 0.29 & 0.27 & 0.33 & 0.45 & 0.23 & 0.12 & 0.20 & 0.39 & 0.33 & - & 0.19 & 0.54 & 0.37 & 0.21 & 0.28 & 0.25 \\
relevance\_tweets23 & 0.58 & 0.42 & 0.32 & 0.18 & 0.57 & 0.24 & 0.31 & 0.19 & 0.16 & 0.42 & 0.22 & 0.35 & 0.32 & 0.47 & 0.24 & 0.14 & 0.22 & 0.24 \\
framesI\_tweets23 & 0.50 & 0.54 & 0.45 & 0.48 & 0.51 & 0.49 & 0.45 & 0.48 & 0.54 & 0.53 & 0.54 & 0.52 & 0.51 & 0.54 & 0.46 & 0.48 & 0.48 & 0.51 \\
relevance\_tweets17 & 0.36 & 0.32 & 0.27 & 0.18 & 0.34 & 0.32 & 0.35 & 0.17 & 0.19 & 0.31 & 0.14 & 0.28 & 0.30 & 0.31 & 0.36 & 0.34 & 0.27 & 0.26 \\
framesII\_tweets17 & 0.55 & 0.53 & 0.47 & 0.40 & 0.62 & 0.63 & 0.45 & 0.36 & 0.37 & 0.57 & 0.56 & 0.50 & 0.36 & 0.56 & 0.59 & 0.33 & 0.39 & 0.32 \\
relevance\_news & 0.57 & 0.16 & 0.11 & 0.07 & 0.26 & 0.20 & 0.08 & 0.09 & 0.07 & 0.18 & 0.12 & 0.10 & 0.07 & 0.48 & 0.10 & 0.07 & 0.08 & 0.07 \\
framesI\_news & 0.60 & 0.45 & 0.44 & 0.41 & 0.50 & 0.39 & 0.34 & 0.39 & 0.36 & 0.47 & 0.39 & 0.48 & 0.35 & 0.54 & 0.45 & 0.39 & 0.38 & 0.24 \\
\bottomrule
\end{tabular}
\end{adjustbox}
% \end{table}

%% file: tables/PAPER_fraction_na_table.tex
% \begin{table}[htbp]
% \tiny
\centering
\caption{Average fraction of non-responses (NA) by task and model, indicating instruction-following failure rates where models produced outputs that could not be mapped to valid annotation categories.}
\renewcommand{\arraystretch}{1.2}
\begin{adjustbox}{max width=\textwidth}
\rowcolors{2}{gray!20}{white}
\begin{tabular}{lrrrr:rrrrr:rrrr:rrr:rr}
\toprule
 & \rotatebox{90}{Llama-3.2-1B} & \rotatebox{90}{Llama-3.2-3B} & \rotatebox{90}{Llama-3.1-8B} & \rotatebox{90}{Llama-3.1-70B} & \rotatebox{90}{Qwen2.5-1.5B} & \rotatebox{90}{Qwen2.5-3B} & \rotatebox{90}{Qwen2.5-7B} & \rotatebox{90}{Qwen2.5-32B} & \rotatebox{90}{Qwen2.5-72B} & \rotatebox{90}{Qwen3-1.7B} & \rotatebox{90}{Qwen3-4B} & \rotatebox{90}{Qwen3-8B} & \rotatebox{90}{Qwen3-32B} & \rotatebox{90}{Gemma-3-1b-it} & \rotatebox{90}{Gemma-3-4b-it} & \rotatebox{90}{Gemma-3-27b-it} & \rotatebox{90}{GPT-4o-mini} & \rotatebox{90}{GPT-4o}6 \\
Task &  &  &  &  &  &  &  &  &  &  &  &  &  &  &  &  &  &  \\
\midrule
politeness\_wiki & 0 & 0.014 & 0.003 & 0.002 & 0 & 0 & 0 & 0 & 0 & 0 & 0 & 0 & 0 & 0 & 0.003 & 0.070 & 0.001 & 0.003 \\
politeness\_stack & 0.004 & 0.029 & 0.007 & 0.003 & 0 & 0 & 0 & 0 & 0 & 0 & 0 & 0 & 0 & 0.001 & 0.028 & 0.076 & 0 & 0.002 \\
stance\_climate & 0.019 & 0.002 & 0 & 0 & 0 & 0 & 0 & 0 & 0 & 0 & 0.001 & 0 & 0 & 0 & 0 & 0 & 0 & 0 \\
ideology\_news & 0.091 & 0 & 0 & 0 & 0.001 & 0 & 0.001 & 0 & 0 & 0 & 0 & 0 & 0 & 0 & 0 & 0.001 & 0 & 0 \\
fakenews & 0.005 & 0 & 0 & 0 & 0 & 0 & 0 & 0.008 & 0 & 0 & 0 & 0 & 0 & 0.344 & 0.001 & 0.001 & 0 & 0 \\
factuality & 0 & 0 & 0 & 0.001 & 0 & 0 & 0 & 0 & 0 & 0 & 0 & 0 & 0 & 0.009 & 0 & 0 & 0 & 0 \\
hatespeech\_explicit & 0.092 & 0.102 & 0 & 0 & 0 & 0 & 0 & 0 & 0 & 0 & 0 & 0 & 0 & 0.001 & 0 & 0 & 0 & 0 \\
hatespeech\_implicit & 0.184 & 0.242 & 0.006 & 0.008 & 0.090 & 0.009 & 0.001 & 0.003 & 0.002 & 0.081 & 0 & 0 & 0 & 0.100 & 0.005 & 0.003 & 0.001 & 0.001 \\
hatespeech\_target & 0 & 0 & 0 & 0 & 0 & 0 & 0 & 0 & 0 & 0 & 0 & 0 & 0 & 0 & 0 & 0 & 0 & 0 \\
emotion & 0 & 0 & 0 & 0 & 0 & 0 & 0 & 0 & 0 & 0 & 0 & 0 & 0 & 0 & 0 & 0 & 0 & 0 \\
essay\_living & 0.005 & 0 & 0 & 0 & 0 & 0 & 0.001 & 0 & 0 & 0 & 0 & 0 & 0 & 0.002 & 0 & 0 & 0 & 0 \\
essay\_housewife & 0 & 0 & 0 & 0 & 0 & 0 & 0 & 0 & 0 & 0 & 0 & 0 & 0 & 0.006 & 0 & 0 & 0 & 0 \\
essay\_worksocial & 0 & 0 & 0 & 0 & 0 & 0 & 0 & 0 & 0 & 0 & 0 & 0 & 0 & 0.093 & 0 & 0 & 0 & 0 \\
essay\_location & 0.033 & 0 & 0.010 & 0.002 & 0 & 0 & 0 & 0 & 0 & 0 & 0.014 & 0 & 0 & 0 & 0 & 0 & 0 & 0 \\
essay\_domestic & 0.027 & 0 & 0 & 0 & 0 & 0.013 & 0 & 0 & 0 & 0 & 0 & 0 & 0 & 0 & 0 & 0 & 0 & 0.001 \\
essay\_narrative & 0.002 & 0 & 0 & 0 & 0 & 0 & 0 & 0 & 0 & 0 & 0 & 0 & 0 & 0.005 & 0 & 0 & 0 & 0 \\
issue\_survey & 0.062 & 0.019 & 0.191 & 0 & 0.106 & 0.057 & 0.027 & 0.001 & 0.013 & 0.093 & 0.004 & 0.008 & 0.006 & 0.289 & 0.024 & 0.007 & 0.001 & 0 \\
humor & 0 & 0.001 & 0 & 0 & 0 & 0 & 0 & 0 & 0 & 0 & 0 & 0 & 0 & 0.319 & 0 & 0 & 0 & 0 \\
topic & 0 & 0 & 0.002 & 0 & 0.072 & 0.014 & 0 & 0 & 0 & 0 & 0 & 0 & 0 & 0.001 & 0 & 0 & 0 & 0 \\
ideology\_tweets & 0.055 & 0.137 & 0.131 & 0.085 & 0.085 & 0.068 & 0.075 & 0.146 & 0.091 & 0.226 & 0.089 & 0.115 & 0.078 & 0.144 & 0.057 & 0.101 & 0.093 & 0.098 \\
misinfo & 0 & 0 & 0 & 0 & 0 & 0 & 0 & 0 & 0 & 0 & 0 & 0 & 0 & 0 & 0.003 & 0 & 0 & 0 \\
tone & 0.036 & 0.001 & 0.013 & 0.015 & 0 & 0 & 0.007 & 0.005 & 0.014 & 0.001 & 0.001 & 0.001 & 0.002 & 0 & 0.002 & 0.011 & 0 & 0 \\
manifestos\_issue & 0.101 & 0.027 & 0.046 & 0.035 & 0.010 & 0.012 & 0.028 & 0.042 & 0.010 & 0.019 & 0.004 & 0.007 & 0.010 & 0.193 & 0.049 & 0.015 & 0.003 & 0.009 \\
manifestos\_econ\_ideology & 0.022 & 0.002 & 0.038 & 0.063 & 0.156 & 0.153 & 0.139 & 0.159 & 0.148 & 0.115 & 0.127 & 0.166 & 0.028 & 0.023 & 0.007 & 0.031 & 0.100 & 0.082 \\
manifestos\_social\_ideology & 0.028 & 0.061 & 0.096 & 0.032 & 0 & 0.049 & 0.031 & 0.095 & 0.068 & 0.023 & 0.018 & 0.023 & 0.005 & 0.062 & 0.013 & 0.016 & 0.017 & 0.030 \\
manifestos\_issues\_detailed & 0.046 & 0.095 & 0.072 & 0.011 & 0.042 & 0.029 & 0.012 & 0.005 & 0.009 & 0.056 & 0.017 & 0.016 & 0.018 & 0.217 & 0.008 & 0.001 & 0.006 & 0.005 \\
relevance\_tweets & 0.341 & 0.002 & 0.286 & 0.144 & 0.286 & 0.287 & 0.287 & 0.286 & 0.144 & 0.143 & 0.143 & 0.144 & 0.286 & 0.286 & 0.286 & 0.287 & 0.286 & 0.286 \\
framesI\_tweets & 0.144 & 0 & 0 & 0 & 0.119 & 0.206 & 0.079 & 0.124 & 0.051 & 0.267 & 0.043 & 0.033 & 0.066 & 0 & 0 & 0 & 0.006 & 0.001 \\
framesII\_tweets & 0.161 & 0.003 & 0.022 & 0.052 & 0.035 & 0.023 & 0.044 & 0.074 & 0.082 & 0.003 & 0.008 & 0.030 & 0.060 & 0.021 & 0.044 & 0.059 & 0.083 & 0.161 \\
stance\_tweets & 0.084 & 0.004 & 0.002 & 0.001 & 0.136 & 0.124 & 0.057 & 0.064 & 0.027 & 0.053 & 0.051 & 0.011 & 0.060 & 0.018 & 0 & 0 & 0.092 & 0.001 \\
topic\_tweets & 0.371 & 0.064 & 0.032 & 0.012 & 0.155 & 0.047 & 0.078 & 0.306 & 0.031 & 0.218 & 0.117 & 0.074 & 0.109 & 0.002 & 0.005 & 0.001 & 0.062 & 0.008 \\
relevance\_tweets23 & 0.041 & 0.002 & 0 & 0 & 0.041 & 0.018 & 0.014 & 0.043 & 0.013 & 0.026 & 0.024 & 0.010 & 0.035 & 0.015 & 0.001 & 0.002 & 0.007 & 0.002 \\
framesI\_tweets23 & 0.089 & 0.002 & 0 & 0 & 0.165 & 0.236 & 0.132 & 0.144 & 0.065 & 0.271 & 0.038 & 0.027 & 0.070 & 0.004 & 0 & 0 & 0.011 & 0.001 \\
relevance\_tweets17 & 0.011 & 0 & 0 & 0 & 0.076 & 0.003 & 0.004 & 0.007 & 0.007 & 0.021 & 0.007 & 0.001 & 0.029 & 0.001 & 0 & 0.002 & 0.002 & 0.001 \\
framesII\_tweets17 & 0.096 & 0.001 & 0.008 & 0.007 & 0.032 & 0.009 & 0.016 & 0.022 & 0.035 & 0.051 & 0 & 0.003 & 0.022 & 0.034 & 0.005 & 0.017 & 0.018 & 0.047 \\
relevance\_news & 0.063 & 0 & 0 & 0 & 0.004 & 0.039 & 0.002 & 0.065 & 0 & 0.014 & 0.001 & 0.001 & 0 & 0.001 & 0 & 0 & 0.004 & 0 \\
framesI\_news & 0.134 & 0.003 & 0.003 & 0 & 0.057 & 0.241 & 0.085 & 0.098 & 0.025 & 0.285 & 0.052 & 0.003 & 0.001 & 0 & 0.001 & 0.001 & 0.015 & 0 \\
\bottomrule
\end{tabular}
\end{adjustbox}
% \end{table}

%% file: tables/regression_llm_hacking.tex
% \begin{table}[!htbp] \centering 
%   \caption{OLS Regression Results: LLM Hacking Risk by Prompt Type, Prompt Detail, F1 Score (weighted) with Task (T) and Model (M) Interactions, and Significance-Related Predictors} 
%   \label{} 
% \footnotesize 
% \begin{tabular}{@{\extracolsep{1pt}}lc} 
% \\[-1.8ex]\hline 
% \hline \\[-1.8ex] 
%  & \multicolumn{1}{c}{\textit{Dependent variable:}} \\ 
% \cline{2-2} 
% \\[-1.8ex] & llm\_hacking\_binary \\ 
% \hline \\[-1.8ex] 
 (shot\_type)zero\_shot & 0.003$^{***}$ (0.001) \\ 
  (detail\_level)short & $-$0.002$^{***}$ (0.001) \\ 
  Weighted F1 Score & $-$0.102$^{***}$ (0.017) \\ 
  (T)essay\_domestic & 0.030$^{***}$ (0.007) \\ 
  (T)essay\_housewife & $-$0.059$^{***}$ (0.013) \\ 
  (T)essay\_living & 0.001 (0.007) \\ 
  (T)essay\_location & $-$0.009 (0.007) \\ 
  (T)essay\_narrative & 0.063$^{***}$ (0.010) \\ 
  (T)essay\_worksocial & 0.088$^{***}$ (0.011) \\ 
  (T)factuality & 0.364$^{***}$ (0.016) \\ 
  (T)fakenews & 0.264$^{***}$ (0.066) \\ 
  (T)framesII\_tweets & 0.067$^{***}$ (0.011) \\ 
  (T)framesII\_tweets17 & 0.018$^{*}$ (0.009) \\ 
  (T)framesI\_news & 0.302$^{***}$ (0.014) \\ 
  (T)framesI\_tweets & 0.463$^{***}$ (0.014) \\ 
  (T)framesI\_tweets23 & 0.080$^{***}$ (0.015) \\ 
  (T)hatespeech\_explicit & 0.269$^{***}$ (0.014) \\ 
  (T)hatespeech\_implicit & 0.342$^{***}$ (0.010) \\ 
  (T)hatespeech\_target & 0.076$^{***}$ (0.007) \\ 
  (T)humor & $-$0.041 (0.027) \\ 
  (T)ideology\_news & 0.317$^{***}$ (0.014) \\ 
  (T)ideology\_tweets & 0.478$^{***}$ (0.047) \\ 
  (T)issue\_survey & $-$0.019$^{***}$ (0.005) \\ 
  (T)manifestos\_econ\_ideology & 1.022$^{***}$ (0.044) \\ 
  (T)manifestos\_issue & 0.422$^{***}$ (0.053) \\ 
  (T)manifestos\_issues\_detailed & 0.052$^{***}$ (0.005) \\ 
  (T)manifestos\_social\_ideology & 0.467$^{***}$ (0.041) \\ 
  (T)misinfo & $-$0.0004 (0.073) \\ 
  (T)politeness\_stack & 0.554$^{***}$ (0.042) \\ 
  (T)politeness\_wiki & 0.876$^{***}$ (0.040) \\ 
  (T)relevance\_news & 0.603$^{***}$ (0.028) \\ 
  (T)relevance\_tweets & 0.829$^{***}$ (0.031) \\ 
  (T)relevance\_tweets17 & 0.431$^{**}$ (0.208) \\ 
  (T)relevance\_tweets23 & 1.070$^{***}$ (0.047) \\ 
  (T)stance\_climate & 0.407$^{***}$ (0.020) \\ 
  (T)stance\_tweets & 0.383$^{***}$ (0.021) \\ 
  (T)tone & 0.355$^{***}$ (0.045) \\ 
  (T)topic & 0.709$^{***}$ (0.026) \\ 
  (T)topic\_tweets & 0.299$^{***}$ (0.020) \\ 
  (M)Qwen2.5-32B & 0.002 (0.007) \\ 
  (M)Qwen2.5-3B & 0.011$^{*}$ (0.006) \\ 
  (M)Qwen2.5-72B & 0.003 (0.007) \\ 
  (M)Qwen2.5-7B & $-$0.008 (0.006) \\ 
  (M)Qwen3-1.7B & $-$0.001 (0.005) \\ 
  (M)Qwen3-32B & $-$0.003 (0.007) \\ 
  (M)Qwen3-4B & 0.001 (0.005) \\ 
  (M)Qwen3-8B & $-$0.00001 (0.007) \\ 
  (M)Gemma-3-1b-it & 0.027$^{***}$ (0.005) \\ 
  (M)Gemma-3-27b-it & 0.001 (0.006) \\ 
  (M)Gemma-3-4b-it & $-$0.005 (0.006) \\ 
  (M)GPT-4o & 0.009 (0.007) \\ 
  (M)GPT-4o-mini & 0.007 (0.007) \\ 
  (M)Llama-3.1-70B & $-$0.00000 (0.007) \\ 
  (M)Llama-3.1-8B & $-$0.005 (0.005) \\ 
  (M)Llama-3.2-1B & 0.039$^{***}$ (0.005) \\ 
  (M)Llama-3.2-3B & $-$0.004 (0.005) \\ 
  significant\_difference\_found & 0.324$^{***}$ (0.001) \\ 
  normalized\_distance\_from\_significance\_treshold & $-$0.220$^{***}$ (0.001) \\ 
  Weighted F1 Score:(T)essay\_domestic & $-$0.042$^{***}$ (0.014) \\ 
  Weighted F1 Score:(T)essay\_housewife & 0.159$^{***}$ (0.019) \\ 
  Weighted F1 Score:(T)essay\_living & $-$0.009 (0.013) \\ 
  Weighted F1 Score:(T)essay\_location & 0.097$^{***}$ (0.015) \\ 
  Weighted F1 Score:(T)essay\_narrative & $-$0.012 (0.024) \\ 
  Weighted F1 Score:(T)essay\_worksocial & $-$0.044$^{**}$ (0.018) \\ 
  Weighted F1 Score:(T)factuality & 0.064$^{*}$ (0.035) \\ 
  Weighted F1 Score:(T)fakenews & $-$0.177 (0.108) \\ 
  Weighted F1 Score:(T)framesII\_tweets & 0.062$^{*}$ (0.033) \\ 
  Weighted F1 Score:(T)framesII\_tweets17 & $-$0.086$^{***}$ (0.024) \\ 
  Weighted F1 Score:(T)framesI\_news & $-$0.268$^{***}$ (0.029) \\ 
  Weighted F1 Score:(T)framesI\_tweets & $-$0.576$^{***}$ (0.029) \\ 
  Weighted F1 Score:(T)framesI\_tweets23 & 0.217$^{***}$ (0.031) \\ 
  Weighted F1 Score:(T)hatespeech\_explicit & $-$0.351$^{***}$ (0.027) \\ 
  Weighted F1 Score:(T)hatespeech\_implicit & $-$0.646$^{***}$ (0.028) \\ 
  Weighted F1 Score:(T)hatespeech\_target & $-$0.128$^{***}$ (0.016) \\ 
  Weighted F1 Score:(T)humor & $-$0.129$^{***}$ (0.043) \\ 
  Weighted F1 Score:(T)ideology\_news & $-$0.135$^{***}$ (0.033) \\ 
  Weighted F1 Score:(T)ideology\_tweets & $-$0.328$^{***}$ (0.074) \\ 
  Weighted F1 Score:(T)issue\_survey & $-$0.007 (0.011) \\ 
  Weighted F1 Score:(T)manifestos\_econ\_ideology & $-$1.522$^{***}$ (0.066) \\ 
  Weighted F1 Score:(T)manifestos\_issue & $-$0.517$^{***}$ (0.076) \\ 
  Weighted F1 Score:(T)manifestos\_issues\_detailed & $-$0.107$^{***}$ (0.013) \\ 
  Weighted F1 Score:(T)manifestos\_social\_ideology & $-$0.581$^{***}$ (0.059) \\ 
  Weighted F1 Score:(T)misinfo & $-$0.033 (0.100) \\ 
  Weighted F1 Score:(T)politeness\_stack & $-$0.754$^{***}$ (0.070) \\ 
  Weighted F1 Score:(T)politeness\_wiki & $-$1.217$^{***}$ (0.054) \\ 
  Weighted F1 Score:(T)relevance\_news & $-$0.780$^{***}$ (0.036) \\ 
  Weighted F1 Score:(T)relevance\_tweets & $-$0.891$^{***}$ (0.044) \\ 
  Weighted F1 Score:(T)relevance\_tweets17 & $-$0.386$^{*}$ (0.229) \\ 
  Weighted F1 Score:(T)relevance\_tweets23 & $-$1.388$^{***}$ (0.068) \\ 
  Weighted F1 Score:(T)stance\_climate & $-$0.598$^{***}$ (0.037) \\ 
  Weighted F1 Score:(T)stance\_tweets & $-$0.541$^{***}$ (0.039) \\ 
  Weighted F1 Score:(T)tone & $-$0.523$^{***}$ (0.056) \\ 
  Weighted F1 Score:(T)topic & $-$0.943$^{***}$ (0.038) \\ 
  Weighted F1 Score:(T)topic\_tweets & $-$0.380$^{***}$ (0.028) \\ 
  Weighted F1 Score:(M)Qwen2.5-32B & $-$0.044$^{***}$ (0.015) \\ 
  Weighted F1 Score:(M)Qwen2.5-3B & $-$0.044$^{***}$ (0.016) \\ 
  Weighted F1 Score:(M)Qwen2.5-72B & $-$0.046$^{***}$ (0.016) \\ 
  Weighted F1 Score:(M)Qwen2.5-7B & $-$0.024 (0.015) \\ 
  Weighted F1 Score:(M)Qwen3-1.7B & $-$0.027$^{*}$ (0.016) \\ 
  Weighted F1 Score:(M)Qwen3-32B & $-$0.040$^{**}$ (0.016) \\ 
  Weighted F1 Score:(M)Qwen3-4B & $-$0.041$^{***}$ (0.014) \\ 
  Weighted F1 Score:(M)Qwen3-8B & $-$0.039$^{**}$ (0.016) \\ 
  Weighted F1 Score:(M)Gemma-3-1b-it & $-$0.040$^{**}$ (0.019) \\ 
  Weighted F1 Score:(M)Gemma-3-27b-it & $-$0.048$^{***}$ (0.015) \\ 
  Weighted F1 Score:(M)Gemma-3-4b-it & $-$0.031$^{**}$ (0.015) \\ 
  Weighted F1 Score:(M)GPT-4o & $-$0.060$^{***}$ (0.016) \\ 
  Weighted F1 Score:(M)GPT-4o-mini & $-$0.056$^{***}$ (0.016) \\ 
  Weighted F1 Score:(M)Llama-3.1-70B & $-$0.047$^{***}$ (0.015) \\ 
  Weighted F1 Score:(M)Llama-3.1-8B & $-$0.031$^{**}$ (0.014) \\ 
  Weighted F1 Score:(M)Llama-3.2-1B & $-$0.041$^{**}$ (0.019) \\ 
  Weighted F1 Score:(M)Llama-3.2-3B & $-$0.025 (0.015) \\ 
  Constant & 0.296$^{***}$ (0.006) \\ 
 \midrule 
Observations & 1,429,925 \\ 
Accuracy (on binarized model predictions) & 0.830 \\ 
R$^{2}$ & 0.154 \\ 
Adjusted R$^{2}$ & 0.154 \\ 
Residual Std. Error & 0.352 (df = 1429813) \\ 
F Statistic & 2,352.722$^{***}$ (df = 111; 1429813) \\ 
\midrule
% \hline \\[-1.8ex] 
\textit{Note:}  & \multicolumn{1}{r}{$^{*}$p$<$0.1; $^{**}$p$<$0.05; $^{***}$p$<$0.01}
% \end{tabular} 
% \end{table} 

%% file: tables/PAPER_llm_hacking_f1_score_weighted_correlations_table.tex
\centering
\caption{Point-biserial correlations between LLM hacking occurrence and model F1 score weighted across different tasks. Significance levels: * p < 0.05, ** p < 0.01, *** p < 0.001.}
\label{tab:llm_hacking_f1_score_weighted_correlations}
\rowcolors{2}{gray!20}{white}
\begin{tabular}{lcccc}
\toprule
Task & Correlation & p-value & Significance & N samples \\
\midrule
politeness\_wiki & -0.292 & 0.000 & *** & 3,741 \\
politeness\_stack & -0.136 & 0.000 & *** & 3,936 \\
stance\_climate & -0.198 & 0.000 & *** & 6,525 \\
ideology\_news & -0.063 & 0.000 & *** & 11,968 \\
fakenews & -0.020 & 0.318 &  & 2,604 \\
factuality & -0.023 & 0.012 & * & 12,015 \\
hatespeech\_explicit & -0.175 & 0.000 & *** & 6,525 \\
hatespeech\_implicit & -0.152 & 0.000 & *** & 18,525 \\
hatespeech\_target & -0.068 & 0.000 & *** & 118,170 \\
emotion & -0.137 & 0.000 & *** & 154,440 \\
essay\_living & -0.115 & 0.000 & *** & 62,745 \\
essay\_housewife & -0.298 & 0.000 & *** & 7,565 \\
essay\_worksocial & -0.017 & 0.173 &  & 6,786 \\
essay\_location & -0.088 & 0.000 & *** & 35,856 \\
essay\_domestic & -0.081 & 0.000 & *** & 69,872 \\
essay\_narrative & -0.082 & 0.000 & *** & 22,704 \\
issue\_survey & -0.126 & 0.000 & *** & 751,502 \\
humor & -0.123 & 0.000 & *** & 1,914 \\
topic & -0.399 & 0.000 & *** & 2,871 \\
ideology\_tweets & -0.146 & 0.000 & *** & 1,120 \\
misinfo & -0.043 & 0.028 & * & 2,670 \\
tone & -0.108 & 0.000 & *** & 6,142 \\
manifestos\_issue & -0.117 & 0.000 & *** & 1,680 \\
manifestos\_econ\_ideology & -0.460 & 0.000 & *** & 1,680 \\
manifestos\_social\_ideology & -0.228 & 0.000 & *** & 1,428 \\
manifestos\_issues\_detailed & -0.101 & 0.000 & *** & 445,224 \\
relevance\_tweets & -0.313 & 0.000 & *** & 1,786 \\
framesI\_tweets & -0.193 & 0.000 & *** & 7,100 \\
framesII\_tweets & -0.085 & 0.000 & *** & 10,206 \\
stance\_tweets & -0.147 & 0.000 & *** & 6,674 \\
topic\_tweets & -0.257 & 0.000 & *** & 4,794 \\
relevance\_tweets23 & -0.398 & 0.000 & *** & 1,568 \\
framesI\_tweets23 & -0.019 & 0.238 &  & 4,059 \\
relevance\_tweets17 & -0.076 & 0.005 & ** & 1,406 \\
framesII\_tweets17 & -0.138 & 0.000 & *** & 21,634 \\
relevance\_news & -0.425 & 0.000 & *** & 2,185 \\
framesI\_news & -0.131 & 0.000 & *** & 7,125 \\
\bottomrule
\end{tabular}

%% file: tables/statistically_distinguishable_estimators.tex
\begin{table}[htbp]
\centering
\caption{Fraction of hypotheses with significantly different LLM vs.\ ground truth regression coefficients at $\alpha=0.05$. Results are aggregated across all hypotheses and tasks.}
\label{tab:statistically_distinguishable_estimators}
\begin{tabular}{l|cccccc}
\toprule
\textbf{Model} & \textbf{Overall} & \textbf{Correct} & \textbf{LLM hacking} & \textbf{Type I} & \textbf{Type II} & \textbf{Type S} \\
\midrule
Llama-3.2-1B & 42.4\% & 20.8\% & 65.7\% & 51.5\% & 60.6\% & 94.2\% \\
Llama-3.2-3B & 39.6\% & 25.9\% & 60.6\% & 49.6\% & 58.7\% & 82.9\% \\
Llama-3.1-8B & 37.2\% & 26.6\% & 59.9\% & 50.6\% & 58.7\% & 77.1\% \\
Llama-3.1-70B & 35.3\% & 26.3\% & 60.5\% & 42.5\% & 58.4\% & 61.8\% \\
\hdashline
Qwen2.5-1.5B & 42.4\% & 27.5\% & 61.9\% & 47.7\% & 59.7\% & 94.3\% \\
Qwen2.5-3B & 38.7\% & 27.3\% & 58.9\% & 51.1\% & 59.1\% & 80.0\% \\
Qwen2.5-7B & 36.7\% & 27.1\% & 58.6\% & 46.7\% & 57.4\% & 70.6\% \\
Qwen2.5-32B & 36.0\% & 26.2\% & 62.2\% & 48.6\% & 63.3\% & 73.5\% \\
Qwen2.5-72B & 34.6\% & 25.8\% & 59.4\% & 45.9\% & 59.0\% & 57.6\% \\
\hdashline
Qwen3-1.7B & 38.2\% & 24.7\% & 57.8\% & 48.5\% & 55.1\% & 86.1\% \\
Qwen3-4B & 37.9\% & 26.6\% & 60.7\% & 51.3\% & 59.3\% & 80.0\% \\
Qwen3-8B & 36.9\% & 27.2\% & 57.7\% & 48.4\% & 57.4\% & 77.8\% \\
Qwen3-32B & 36.3\% & 27.1\% & 58.9\% & 47.5\% & 57.7\% & 64.7\% \\
\hdashline
gemma-3-1b-it & 40.7\% & 21.5\% & 62.1\% & 50.2\% & 58.7\% & 88.9\% \\
gemma-3-4b-it & 37.6\% & 26.2\% & 60.5\% & 47.0\% & 59.3\% & 77.1\% \\
gemma-3-27b-it & 37.0\% & 27.8\% & 59.5\% & 47.4\% & 57.3\% & 67.6\% \\
\hdashline
GPT-4o-mini & 35.3\% & 26.8\% & 57.5\% & 47.3\% & 58.6\% & 63.9\% \\
GPT-4o & 36.6\% & 28.0\% & 61.4\% & 45.4\% & 60.8\% & 67.6\% \\
\bottomrule
\end{tabular}
\end{table}

%% file: tables/datasets_overview.tex
{{\tiny
\renewcommand{\arraystretch}{1.1}
\begin{longtable}{m{3cm}m{1.5cm}m{1.8cm}m{5.2cm}m{1.5cm}m{1.5cm}}
\caption{Overview of datasets used in experiments.} \label{tab:datasets_overview} \\
\toprule
\rowcolors{2}{gray!25}{white}
 & Dataset Name & Data Type & Dataset Description & Nr. of Datapoints & Nr. of Unique Datapoints \\
\midrule
\endfirsthead
\caption[]{Overview of datasets used in experiments} \\
\toprule
% \rowcolors{2}{gray!25}{white}
 & Dataset Name & Data Type & Dataset Description & Nr. of Datapoints & Nr. of Unique Datapoints \\
\midrule
\endhead
\midrule
\multicolumn{6}{r}{Continued on next page} \\
\midrule
\endfoot
\bottomrule
\endlastfoot
\citet{UniversityofLondonNational_Child_Development_Study2024} & \hyperref[app:datasetname:essay]{essay} & Essays & Essays written by eleven year old cohort members from the National Child Development Study (NCDS) in 1969 with the following instructions: "Imagine you are now 25 years old. Write about the life you are leading, your interests, your home life and your work at the age of 25. (You have 30 minutes to do this)." Additionally, they were given a short questionnaire to complete at school about their interests outside school, the school subjects they enjoyed most, and what they thought they were most likely to do when they left secondary school. & 489 & 489 \\
\rowcolor{gray!15}
\citet{baly-etal-2020-detect} & \hyperref[app:datasetname:ideologynews]{ideology\_news} & News articles & A set of articles from media sources covering the United States of America. Data was scraped from \url{https://www.allsides.com/}. & 37554 & 37554 \\
\citet{benoit2016crowd} & \hyperref[app:datasetname:manifestosuk]{manifestos\_uk} & Party Manifestos from the UK & Sentences from manifestos from UK parties, 1987 to 2010. & 11688 & 11620 \\
\rowcolor{gray!15}
\citet{carlson2017pairwise} & \hyperref[app:datasetname:tone]{tone} & Political ads & The data contains 935 political ads from US Senate Candidates. The data was constructed by~\citet{carlson2017pairwise} (accessed via~\citet{OrnsteinBlasingameTruscott2025}). & 859 & 859 \\
\citet{danescu-niculescu-mizil-etal-2013-computational} & \hyperref[app:datasetname:politenessstack]{politeness\_stack} & StackExchange requests & Requests from the StackExchange question-answering communities. These are often users commenting on existing posts requesting further information or proposing edits. & 3302 & 3298 \\
\rowcolor{gray!15}
\citet{danescu-niculescu-mizil-etal-2013-computational} & \hyperref[app:datasetname:politenesswiki]{politeness\_wiki} & Wikipedia requests & Requests from Wikipedia editor's on user talk pages. & 2178 & 2171 \\
\citet{demszky-etal-2020-goemotions} & \hyperref[app:datasetname:emotion]{emotion} & Reddit comments & Reddit comments from 2005-2019 selected from subreddits with 10k+ comments, filtered to remove deleted, non-English, and toxic content. & 7347 & 7276 \\
\rowcolor{gray!15}
\citet{egami2023using} & \hyperref[app:datasetname:topic]{topic} & Congressional bills & A balanced random sample of congressional bills. They are relating to economy (5k), or three other topics (5k). & 10000 & 10000 \\
\citet{elsherief-etal-2021-latent} & \hyperref[app:datasetname:hatespeech]{hatespeech} & Tweets & Filtered set of tweets, retweets, and replies from most prominent U.S hate groups between January 1, 2015 and December 31, 2017. & 21480 & 21476 \\
\rowcolor{gray!15}
\citet{fieldhouse2024british} & \hyperref[app:datasetname:issuesurvey]{issue\_survey} & Open ended survey responses & The British Election Study (BES) dataset contains responses from 118,597 respondents to survey questions from 29 Internet Panel waves, conducted between February 2014 and July 2024. & 562514 & 101930 \\
\citet{gabriel-etal-2022-misinfo} & \hyperref[app:datasetname:misinfo]{misinfo} & News headlines & The Misinfo Reaction Frames corpus is an dataset of 25k news headlines to articles that have been factchecked. The articles can be about Covid-19, Cancer or Climate Change. & 25164 & 25164 \\
\rowcolor{gray!15}
\citet{gilardi2023pnas} & \hyperref[app:datasetname:news]{news} & News articles & A random sample of 1,606 articles newspaper articles on content moderation published from January 2020 to April 2021, drawn from a dataset of 980k articles collected via LexisNexis. & 1606 & 1599 \\
\citet{gilardi2023pnas} & \hyperref[app:datasetname:tweets17]{tweets17} & Tweets & A random sample of 1,856 tweets posted by members of the US Congress from 2017 to 2022, drawn from a dataset of 20 million tweets. & 1856 & 1856 \\
\rowcolor{gray!15}
\citet{gilardi2023pnas} & \hyperref[app:datasetname:tweets23]{tweets23} & Tweets & This dataset is similar to the tweets dataset, but for January 2023. It includes a random sample of 500 tweets (of which 339 were in English) drawn from a dataset of 1.3 million tweets. & 411 & 411 \\
\citet{gilardi2023pnas} & \hyperref[app:datasetname:tweets]{tweets} & Tweets & A random sample of 2,382 tweets drawn from a dataset of 2.6 million tweets on content moderation posted from January 2020 to April 2021. & 2383 & 1934 \\
\rowcolor{gray!15}
\citet{luo-etal-2020-detecting} & \hyperref[app:datasetname:stanceclimate]{stance\_climate} & News article spans & Opinion spans extracted from global warming news articles, published from Jan. 1, 2000 to April 12, 2020 by various U.S. news sources. & 1996 & 1793 \\
\citet{merz2016manifesto, Lehmann2024} & \hyperref[app:datasetname:manifestos]{manifestos} & Political Party Manifestos & Political manifesto sentences from the manifestoproject corpus, 2000 to 2022. & 1305323 & 1239149 \\
\rowcolor{gray!15}
\citet{min-etal-2023-factscore} & \hyperref[app:datasetname:factuality]{factuality} & LLM outputs & LLM prompt-response pairs. The outputs were generated by three different models (InstructGPT~\citep{NEURIPS2022b1efde53}, ChatGPT\footnote{\url{https://chatgpt.com/}}, and search augmented PerplexityAI\footnote{\url{https://www.perplexity.ai/}}), using the prompt "Tell me a bio of <entity>", where entity corresponds to one of 183 people entities from Wikidata who have corresponding Wikipedia page (randomly sampled over 20 categories based on pageview frequency and nationality). & 16040 & 15225 \\
\citet{shu2020fakenewsnet, Wu2024SheepDog} & \hyperref[app:datasetname:fakenews]{fakenews} & News articles & News articles that have been fact-checked by PolitiFact or GossipCop. & 8366 & 7954 \\
\rowcolor{gray!15}
\citet{tornberg2024large} & \hyperref[app:datasetname:ideologytweets]{ideology\_tweets} & Tweets & Tweets made by politicians before elections. & 4100 & 4099 \\
\citet{weller-seppi-2019-humor} & \hyperref[app:datasetname:humor]{humor} & Reddit posts & 15910 posts from reddit/r/jokes. & 15910 & 15817 \\
\midrule
\textbf{Total} & \textbf{21 datasets} &  &  & \textbf{2040566} & \textbf{1511674} \\
\end{longtable}
}}

%% file: appendix_dataset_and_task_overview/appendix_datasets_tasks.tex
\section{Dataset and Task Overview}
\label{app:Dataset and Task Overview}

Here, we provide a detailed description of all datasets (\faDatabase) and annotation tasks (\faEdit) used in the experiments.

Table~\ref{tab:datasets_overview} concisely lists all datasets.

\subsection{\faDatabase $\;$ politeness\_wiki dataset}
\label{app:datasetname:politenesswiki}

\begin{itemize}
\item \textbf{Data Type:} Wikipedia requests
\item \textbf{Dataset description:} Requests from Wikipedia editor's on user talk pages.
\item \textbf{Dataset Links:} \url{http://www.cs.cornell.edu/~cristian/Politeness_files/Stanford_politeness_corpus.zip}
\item \textbf{Dataset citation:} \citet{danescu-niculescu-mizil-etal-2013-computational}
\end{itemize}

\subsubsection{\faEdit $\;$ politeness\_wiki task}
\label{app:taskname:politenesswiki}

\begin{itemize}
\item \textbf{Task description:} Indicate perceived politeness of wikipedia request.

\item \textbf{Prompts taken from}:~\citep{gligoric-etal-2025-unconfident, ziems-etal-2024-large}

\item \textbf{Ground truth annotators:} Amazon Mechanical Turk workers (US-based)

\item \textbf{Notes:} Following~\citet{gligoric-etal-2025-unconfident}, we consider perceived politeness as a binary classification task and omit the neutral class. This reduces the number of datapoints from 4338 to 2171.
\end{itemize}

\subsection{\faDatabase $\;$ politeness\_stack dataset}
\label{app:datasetname:politenessstack}

\begin{itemize}
\item \textbf{Data Type:} StackExchange requests
\item \textbf{Dataset description:} Requests from the StackExchange question-answering communities. These are often users commenting on existing posts requesting further information or proposing edits.
\item \textbf{Dataset Links:} \url{http://www.cs.cornell.edu/~cristian/Politeness_files/Stanford_politeness_corpus.zip}
\item \textbf{Dataset citation:} \citet{danescu-niculescu-mizil-etal-2013-computational}
\end{itemize}

\subsubsection{\faEdit $\;$ politeness\_stack task}
\label{app:taskname:politenessstack}

\begin{itemize}
\item \textbf{Task description:} Indicate perceived politeness of a StackExchange request.

\item \textbf{Prompts taken from}:~\citep{gligoric-etal-2025-unconfident, ziems-etal-2024-large}

\item \textbf{Ground truth annotators:} Amazon Mechanical Turk workers (US-based)

\item \textbf{Notes:} Following~\citet{gligoric-etal-2025-unconfident}, we consider perceived politeness as a binary classification task and omit the neutral class. This reduces the number of datapoints from 6594 to 3298.
\end{itemize}

\subsection{\faDatabase $\;$ stance\_climate dataset}
\label{app:datasetname:stanceclimate}

\begin{itemize}
\item \textbf{Data Type:} News article spans
\item \textbf{Dataset description:} Opinion spans extracted from global warming news articles, published from Jan. 1, 2000 to April 12, 2020 by various U.S. news sources.
\item \textbf{Dataset Links:} \url{https://github.com/yiweiluo/GWStance}
\item \textbf{Dataset citation:} \citet{luo-etal-2020-detecting}
\end{itemize}

\subsubsection{\faEdit $\;$ stance\_climate task}
\label{app:taskname:stanceclimate}

\begin{itemize}
\item \textbf{Task description:} Annotate the stance of a sentence in terms of agreeing, disagreeing, or being neutral with respect to the target opinion, "Climate change/-global warming is a serious concern."

\item \textbf{Prompts taken from}:~\citep{luo-etal-2020-detecting, gligoric-etal-2025-unconfident}

\item \textbf{Ground truth annotators:} Amazon Mechanical Turk workers (US-based)

\item \textbf{Notes:} We only include data where at least 5 out of 8 crowdworkers agree on the stance.
\end{itemize}

\subsection{\faDatabase $\;$ ideology\_news dataset}
\label{app:datasetname:ideologynews}

\begin{itemize}
\item \textbf{Data Type:} News articles
\item \textbf{Dataset description:} A set of articles from media sources covering the United States of America. Data was scraped from \url{https://www.allsides.com/}.
\item \textbf{Dataset Links:} \url{https://github.com/ramybaly/Article-Bias-Prediction}
\item \textbf{Dataset citation:} \citet{baly-etal-2020-detect}
\end{itemize}

\subsubsection{\faEdit $\;$ ideology\_news task}
\label{app:taskname:ideologynews}

\begin{itemize}
\item \textbf{Task description:} Labelling news articles according to Left, Right, and Centrist political bias. This task provides an entire news article as context.

\item \textbf{Prompts taken from}:~\citep{gligoric-etal-2025-unconfident, ziems-etal-2024-large}

\item \textbf{Ground truth annotators:} Ratings provided by AllSides.

\end{itemize}

\subsection{\faDatabase $\;$ fakenews dataset}
\label{app:datasetname:fakenews}

\begin{itemize}
\item \textbf{Data Type:} News articles
\item \textbf{Dataset description:} News articles that have been fact-checked by PolitiFact or GossipCop.
\item \textbf{Dataset Links:} \url{https://github.com/KaiDMML/FakeNewsNet}, \url{https://github.com/jiayingwu19/SheepDog}
\item \textbf{Dataset citation:} \citet{shu2020fakenewsnet, Wu2024SheepDog}
\end{itemize}

\subsubsection{\faEdit $\;$ fakenews task}
\label{app:taskname:fakenews}

\begin{itemize}
\item \textbf{Task description:} Labelling news articles as fake or real depending on whether they contain fake claims or not.

\item \textbf{Prompts taken from}:~\citep{Wu2024SheepDog}

\item \textbf{Ground truth annotators:} Journalists and domain experts

\item \textbf{Notes:} The dataset was originally released by \citet{shu2020fakenewsnet} with ground truth annotations and metadata. \citet{Wu2024SheepDog} subsequently provided the news article contents.
\end{itemize}

\subsection{\faDatabase $\;$ factuality dataset}
\label{app:datasetname:factuality}

\begin{itemize}
\item \textbf{Data Type:} LLM outputs
\item \textbf{Dataset description:} LLM prompt-response pairs. The outputs were generated by three different models (InstructGPT~\citep{NEURIPS2022b1efde53}, ChatGPT\footnote{\url{https://chatgpt.com/}}, and search augmented PerplexityAI\footnote{\url{https://www.perplexity.ai/}}), using the prompt "Tell me a bio of <entity>", where entity corresponds to one of 183 people entities from Wikidata who have corresponding Wikipedia page (randomly sampled over 20 categories based on pageview frequency and nationality).
\item \textbf{Dataset Links:} \url{https://drive.google.com/drive/folders/1kFey69z8hGXScln01mVxrOhrqgM62X7I?usp=sharing}
\item \textbf{Dataset citation:} \citet{min-etal-2023-factscore}
\end{itemize}

\subsubsection{\faEdit $\;$ factuality task}
\label{app:taskname:factuality}

\begin{itemize}
\item \textbf{Task description:} LLM responses that have been manually split into more than 16K individual facts, and each individual fact has been manually labeled as either supported, irrelevant, or not supported.

\item \textbf{Prompts taken from}:~\citep{min-etal-2023-factscore}

\item \textbf{Ground truth annotators:} Crowdworkers (fact-checking experts through Upwork)

\end{itemize}

\subsection{\faDatabase $\;$ hatespeech dataset}
\label{app:datasetname:hatespeech}

\begin{itemize}
\item \textbf{Data Type:} Tweets
\item \textbf{Dataset description:} Filtered set of tweets, retweets, and replies from most prominent U.S hate groups between January 1, 2015 and December 31, 2017.
\item \textbf{Dataset Links:} \url{https://www.dropbox.com/s/24meryhqi1oo0xk/implicit-hate-corpus.zip?dl=0}
\item \textbf{Dataset citation:} \citet{elsherief-etal-2021-latent}
\end{itemize}

\subsubsection{\faEdit $\;$ hatespeech\_explicit task}
\label{app:taskname:hatespeechexplicit}

\begin{itemize}
\item \textbf{Task description:} Classify tweets as explicit hate speech, implicit hate speech, or not hate speech based on whether they attack people based on protected characteristics.

\item \textbf{Prompts taken from}:~\citep{elsherief-etal-2021-latent}

\item \textbf{Ground truth annotators:} Amazon Mechanical Turk annotators

\end{itemize}

\subsubsection{\faEdit $\;$ hatespeech\_implicit task}
\label{app:taskname:hatespeechimplicit}

\begin{itemize}
\item \textbf{Task description:} Categorize implicit hate tweets into six specific types: white grievance, incitement, inferiority, irony, stereotypical, or threatening.
This task is only done on a subset of the dataset, considers tweets that have been categorized as ``implicit hate speech'' in a prior annotation stage (hatespeech\_explicit task).

\item \textbf{Prompts taken from}:~\citep{elsherief-etal-2021-latent, ziems-etal-2024-large}

\item \textbf{Ground truth annotators:} Expert annotators (trained research assistants)

\item \textbf{Notes:} We only consider the six implicit hate speech categories and omit tweets annotated with the 'other' class (less than 2\% of the dataset).
\end{itemize}

\subsubsection{\faEdit $\;$ hatespeech\_target task}
\label{app:taskname:hatespeechtarget}

\begin{itemize}
\item \textbf{Task description:} Identify the target group of implicit hate tweets from 26 possible demographic and political groups.
This task is only done on a subset of the dataset, considers tweets that have been categorized as ``implicit hate speech'' in a prior annotation stage (hatespeech\_explicit task).

\item \textbf{Prompts taken from}:~\citep{elsherief-etal-2021-latent}

\item \textbf{Ground truth annotators:} Expert annotators (trained research assistants)

\item \textbf{Notes:} We consider only the most frequent 26 targets an annotation classes, which reduces the dataset size from 6358 to 5475.
\end{itemize}

\subsection{\faDatabase $\;$ emotion dataset}
\label{app:datasetname:emotion}

\begin{itemize}
\item \textbf{Data Type:} Reddit comments
\item \textbf{Dataset description:} Reddit comments from 2005-2019 selected from subreddits with 10k+ comments, filtered to remove deleted, non-English, and toxic content.
\item \textbf{Dataset Links:} \url{https://github.com/google-research/google-research/tree/master/goemotions}
\item \textbf{Dataset citation:} \citet{demszky-etal-2020-goemotions}
\end{itemize}

\subsubsection{\faEdit $\;$ emotion task}
\label{app:taskname:emotion}

\begin{itemize}
\item \textbf{Task description:} Classify the emotion expressed in Reddit comments using 25 emotion categories plus neutral.

\item \textbf{Prompts taken from}:~\citep{demszky-etal-2020-goemotions}

\item \textbf{Ground truth annotators:} Native English speakers from India

\item \textbf{Notes:} We only include reddit comments with at least three annotators and where all annotators agree on a single label.
\end{itemize}

\subsection{\faDatabase $\;$ essay dataset}
\label{app:datasetname:essay}

\begin{itemize}
\item \textbf{Data Type:} Essays
\item \textbf{Dataset description:} Essays written by eleven year old cohort members from the National Child Development Study (NCDS) in 1969 with the following instructions: "Imagine you are now 25 years old. Write about the life you are leading, your interests, your home life and your work at the age of 25. (You have 30 minutes to do this)." Additionally, they were given a short questionnaire to complete at school about their interests outside school, the school subjects they enjoyed most, and what they thought they were most likely to do when they left secondary school.
\item \textbf{Dataset Links:} \url{https://beta.ukdataservice.ac.uk/datacatalogue/studies/study?id=5790}
\item \textbf{Dataset citation:} \citet{UniversityofLondonNational_Child_Development_Study2024}
\item \textbf{Dataset notes:} For all tasks in this dataset, we formulated prompts based on the original human annotation guidelines.
\end{itemize}

\subsubsection{\faEdit $\;$ essay\_living task}
\label{app:taskname:essayliving}

\begin{itemize}
\item \textbf{Task description:} Description of the anticipated living situation at age 25.

\item \textbf{Ground truth annotators:} Human experts

\end{itemize}

\subsubsection{\faEdit $\;$ essay\_housewife task}
\label{app:taskname:essayhousewife}

\begin{itemize}
\item \textbf{Task description:} Whether being a housewife is mentioned as an occupation or role.

\item \textbf{Ground truth annotators:} Human experts

\end{itemize}

\subsubsection{\faEdit $\;$ essay\_worksocial task}
\label{app:taskname:essayworksocial}

\begin{itemize}
\item \textbf{Task description:} Whether social aspects of work are mentioned.

\item \textbf{Ground truth annotators:} Human experts

\end{itemize}

\subsubsection{\faEdit $\;$ essay\_location task}
\label{app:taskname:essaylocation}

\begin{itemize}
\item \textbf{Task description:} Geographic location mentioned for living at age 25.

\item \textbf{Ground truth annotators:} Human experts

\end{itemize}

\subsubsection{\faEdit $\;$ essay\_domestic task}
\label{app:taskname:essaydomestic}

\begin{itemize}
\item \textbf{Task description:} Type of domestic labor discussed.

\item \textbf{Ground truth annotators:} Human experts

\end{itemize}

\subsubsection{\faEdit $\;$ essay\_narrative task}
\label{app:taskname:essaynarrative}

\begin{itemize}
\item \textbf{Task description:} Level of narrative structure in the essay.

\item \textbf{Ground truth annotators:} Human experts

\end{itemize}

\subsection{\faDatabase $\;$ issue\_survey dataset}
\label{app:datasetname:issuesurvey}

\begin{itemize}
\item \textbf{Data Type:} Open ended survey responses
\item \textbf{Dataset description:} The British Election Study (BES) dataset contains responses from 118,597 respondents to survey questions from 29 Internet Panel waves, conducted between February 2014 and July 2024.
\item \textbf{Dataset Links:} \url{https://www.britishelectionstudy.com/data-object/british-election-study-combined-wave-1-26-internet-panel-open-ended-response-data/}, \url{https://www.britishelectionstudy.com/data-object/british-election-study-combined-wave-1-29-internet-panel/}
\item \textbf{Dataset citation:} \citet{fieldhouse2024british}
\end{itemize}

\subsubsection{\faEdit $\;$ issue\_survey task}
\label{app:taskname:issuesurvey}

\begin{itemize}
\item \textbf{Task description:} In each wave of the British Election Study Internet Panel, respondents are asked the following question, with the option to provide an open text response: "As far as you're concerned, what is the SINGLE MOST important issue facing the country at the present time?".
The task is to classify the response to this most important issue (MII) question into one of 49 categories.

\item \textbf{Prompts taken from}:~\citep{Mellon2024Issue}

\item \textbf{Ground truth annotators:} Expert annotators (British election study research associates)

\item \textbf{Notes:} The open text responses to the MII question from wave 26 onwards of the BES were coded using GPT-4o and not by BES research associates~\citep{fieldhouse2024british}. Therefore, following the study of~\citet{Mellon2024Issue}, we only consider waves 1-25, which were coded by human experts, in our experiments. We also exclude trivial samples where the open text response is equivalent to the coded category. We also exlude MII ground truth code 46, which is used for responses that cannot be coded as they are unintelligible, are excessively vague as to what aspect of politics they are getting at (e.g., "So many issues, not sure where to start", "attitudes", "our future"), do not relate to politics or society in any meaningful way or when it is impossible to choose between multiple possible codes.
\end{itemize}

\subsection{\faDatabase $\;$ humor dataset}
\label{app:datasetname:humor}

\begin{itemize}
\item \textbf{Data Type:} Reddit posts
\item \textbf{Dataset description:} 15910 posts from reddit/r/jokes.
\item \textbf{Dataset Links:} \url{https://github.com/orionw/RedditHumorDetection}
\item \textbf{Dataset citation:} \citet{weller-seppi-2019-humor}
\end{itemize}

\subsubsection{\faEdit $\;$ humor task}
\label{app:taskname:humor}

\begin{itemize}
\item \textbf{Task description:} The task is to predict whether the text is funny to a lot of people. The ground truth is operationalized as a binary variable indicating if the post received more than 200 upvotes~\citep{weller-seppi-2019-humor}.

\item \textbf{Prompts taken from}:~\citep{ziems-etal-2024-large}

\item \textbf{Ground truth annotators:} Reddit users

\item \textbf{Notes:} On average, each tweet received 2918.39 votes.
\end{itemize}

\subsection{\faDatabase $\;$ topic dataset}
\label{app:datasetname:topic}

\begin{itemize}
\item \textbf{Data Type:} Congressional bills
\item \textbf{Dataset description:} A balanced random sample of congressional bills. They are relating to economy (5k), or three other topics (5k).
\item \textbf{Dataset Links:} \url{https://osf.io/gjt87/files/osfstorage}
\item \textbf{Dataset citation:} \citet{egami2023using}
\end{itemize}

\subsubsection{\faEdit $\;$ topic task}
\label{app:taskname:topic}

\begin{itemize}
\item \textbf{Task description:} Binary topic classification, more precisely, identify whether a congressional bill is relating to the economy or not.

\item \textbf{Ground truth annotators:} Trained human annotators

\item \textbf{Notes:} We only use the random sample of 10K provided by the authors, consisting of one balanced random sample from the full data that they use in the paper (5k texts about economy, 5k about 3 other topics). The full dataset contains 400k bills as is available at \url{https://comparativeagendas.s3.amazonaws.com/datasetfiles/US-Legislative-congressional\_bills\_19.3\_3\_2.csv}.
\end{itemize}

\subsection{\faDatabase $\;$ ideology\_tweets dataset}
\label{app:datasetname:ideologytweets}

\begin{itemize}
\item \textbf{Data Type:} Tweets
\item \textbf{Dataset description:} Tweets made by politicians before elections.
\item \textbf{Dataset Links:} \url{https://github.com/cssmodels/llm/tree/main/Data/countries}
\item \textbf{Dataset citation:} \citet{tornberg2024large}
\end{itemize}

\subsubsection{\faEdit $\;$ ideology\_tweets task}
\label{app:taskname:ideologytweets}

\begin{itemize}
\item \textbf{Task description:} Classify tweet author's party affiliation from tweet content.

\item \textbf{Prompts taken from}:~\citep{tornberg2024large}

\item \textbf{Task-specific data sampling:} We used all data.

\item \textbf{Ground truth annotators:} Ground truth values correspond to the actual political party affiliation of the tweet's author.

\item \textbf{Notes:} We use the genderizeR package to predict gender from first names of tweet authors~\citep{genderizeR}.
\end{itemize}

\subsection{\faDatabase $\;$ misinfo dataset}
\label{app:datasetname:misinfo}

\begin{itemize}
\item \textbf{Data Type:} News headlines
\item \textbf{Dataset description:} The Misinfo Reaction Frames corpus is an dataset of 25k news headlines to articles that have been factchecked. The articles can be about Covid-19, Cancer or Climate Change.
\item \textbf{Dataset Links:} \url{https://github.com/skgabriel/mrf-modeling}
\item \textbf{Dataset citation:} \citet{gabriel-etal-2022-misinfo}
\end{itemize}

\subsubsection{\faEdit $\;$ misinfo task}
\label{app:taskname:misinfo}

\begin{itemize}
\item \textbf{Task description:} The annotation task is to predict whether the news article is likely to contain misinformation.

\item \textbf{Prompts taken from}:~\citep{ziems-etal-2024-large}

\item \textbf{Ground truth annotators:} Amazon Mechanical Turk workers (US-based)

\end{itemize}

\subsection{\faDatabase $\;$ tone dataset}
\label{app:datasetname:tone}

\begin{itemize}
\item \textbf{Data Type:} Political ads
\item \textbf{Dataset description:} The data contains 935 political ads from US Senate Candidates. The data was constructed by~\citet{carlson2017pairwise} (accessed via~\citet{OrnsteinBlasingameTruscott2025}).
\item \textbf{Dataset Links:} \url{https://dataverse.harvard.edu/dataset.xhtml?persistentId=doi:10.7910/DVN/DZZ0OM}
\item \textbf{Dataset citation:} \citet{carlson2017pairwise}
\end{itemize}

\subsubsection{\faEdit $\;$ tone task}
\label{app:taskname:tone}

\begin{itemize}
\item \textbf{Task description:} Classifying the tone of a political ad as positive, neutral, or negative.

\item \textbf{Prompts taken from}:~\citep{OrnsteinBlasingameTruscott2025}

\item \textbf{Ground truth annotators:} Expert coders

\item \textbf{Notes:} Ground truth labels were aggregated from a 5-point expert coding scale to a 3-class system. Expert coders originally classified ads on a scale from 1 (positive/promoting) to 5 (attack), considering whether ads promoted a single candidate, contrasted candidates, or attacked candidates. For contrasting ads, coders further distinguished whether they were more promotional, more attacking, or balanced. The 5-point scale was collapsed by mapping scores 1-2 to "positive", score 3 to "neutral", and scores 4-5 to "negative" to create the final tone classification.
\end{itemize}

\subsection{\faDatabase $\;$ manifestos\_uk dataset}
\label{app:datasetname:manifestosuk}

\begin{itemize}
\item \textbf{Data Type:} Party Manifestos from the UK
\item \textbf{Dataset description:} Sentences from manifestos from UK parties, 1987 to 2010.
\item \textbf{Dataset Links:} \url{https://github.com/kbenoit/CSTA-APSR/}, \url{https://dataverse.harvard.edu/file.xhtml?fileId=10595729\&version=1.0}
\item \textbf{Dataset citation:} \citet{benoit2016crowd}
\end{itemize}

\subsubsection{\faEdit $\;$ manifestos\_issue task}
\label{app:taskname:manifestosissue}

\begin{itemize}
\item \textbf{Task description:} Identify two widely used policy dimensions: "economic" policy and "social" policy issue dimension.

\item \textbf{Prompts taken from}:~\citep{barrie2024replication, OrnsteinBlasingameTruscott2025}

\item \textbf{Ground truth annotators:} Crowd workers from 26 CrowdFlower channels

\item \textbf{Notes:} We only include data where more than 80\% of crowdworkers agree on the label. Then, we use the crowd-worker majority-vote as the ground truth. We load the data from~\citep{benoit2016crowd} and~\citep{OrnsteinBlasingameTruscott2025}.
\end{itemize}

\subsubsection{\faEdit $\;$ manifestos\_econ\_ideology task}
\label{app:taskname:manifestoseconideology}

\begin{itemize}
\item \textbf{Task description:} Classify sentences as left or right (for sentences on economic policy).

\item \textbf{Prompts taken from}:~\citep{barrie2024replication, OrnsteinBlasingameTruscott2025}

\item \textbf{Ground truth annotators:} Crowd workers from 26 CrowdFlower channels

\item \textbf{Notes:} We only include data where more than 80\% of crowdworkers agree on the sentence being about economic policy. And we also only include data where more than 80\% of crowdworkers agree on the binarized left ('very left' or 'somewhat left') vs. right ('very right' or 'somewhat right') label. Then, we use the crowd-worker majority-vote as the ground truth. We load the data from~\citep{benoit2016crowd} and~\citep{OrnsteinBlasingameTruscott2025}.
\end{itemize}

\subsubsection{\faEdit $\;$ manifestos\_social\_ideology task}
\label{app:taskname:manifestossocialideology}

\begin{itemize}
\item \textbf{Task description:} Classify sentences as liberal or conservative (for sentences on social policy).

\item \textbf{Prompts taken from}:~\citep{barrie2024replication, OrnsteinBlasingameTruscott2025}

\item \textbf{Ground truth annotators:} Crowd workers from 26 CrowdFlower channels

\item \textbf{Notes:} We only include data where more than 80\% of crowdworkers agree on the sentence being about social policy. And we also only include data where more than 80\% of crowdworkers agree on the liberal vs. conservative label. Then, we use the crowd-worker majority-vote as the ground truth. We load the data from~\citep{benoit2016crowd} and~\citep{OrnsteinBlasingameTruscott2025}.
\end{itemize}

\subsection{\faDatabase $\;$ manifestos dataset}
\label{app:datasetname:manifestos}

\begin{itemize}
\item \textbf{Data Type:} Political Party Manifestos
\item \textbf{Dataset description:} Political manifesto sentences from the manifestoproject corpus, 2000 to 2022.
\item \textbf{Dataset Links:} \url{https://manifesto-project.wzb.eu/}, \url{https://manifesto-project.wzb.eu/down/data/2024a/datasets/MPDataset_MPDS2024a.csv}
\item \textbf{Dataset citation:} \citet{merz2016manifesto, Lehmann2024}
\end{itemize}

\subsubsection{\faEdit $\;$ manifestos\_issues\_detailed task}
\label{app:taskname:manifestosissuesdetailed}

\begin{itemize}
\item \textbf{Task description:} The task is to classify quasi-sentences in party manifestos according to the manifesto project coding scheme. There are 46 categories in total (56 in the original task, where categories that differed only in stance have been aggregated into their main issues).

\item \textbf{Ground truth annotators:} Trained country experts (mostly political scientist or political science students and native speakers)

\item \textbf{Notes:} We used all data available for parties since 2001, with English translation and sentence level annotations.
\end{itemize}

\subsection{\faDatabase $\;$ tweets dataset}
\label{app:datasetname:tweets}

\begin{itemize}
\item \textbf{Data Type:} Tweets
\item \textbf{Dataset description:} A random sample of 2,382 tweets drawn from a dataset of 2.6 million tweets on content moderation posted from January 2020 to April 2021.
\item \textbf{Dataset Links:} \url{https://dataverse.harvard.edu/dataset.xhtml?persistentId=doi:10.7910/DVN/PQYF6M}
\item \textbf{Dataset citation:} \citet{gilardi2023pnas}
\end{itemize}

\subsubsection{\faEdit $\;$ relevance\_tweets task}
\label{app:taskname:relevancetweets}

\begin{itemize}
\item \textbf{Task description:} Binary annotation of whether a tweet is about content moderation (relevant/irrelevant).

\item \textbf{Prompts taken from}:~\citep{gilardi2023pnas, alizadeh2025open}

\item \textbf{Ground truth annotators:} Trained research assistants (political science students)

\item \textbf{Notes:} \citet{gilardi2023pnas} refer to this as the 'Relevance' task for the 'Tweets (2020--2021)' dataset. We employ identical data preprocessing as \citet{gilardi2023pnas}, which means that we consider only tweets for which all annotators agree on the ground truth label. Additionally, we filter out duplicates.
\end{itemize}

\subsubsection{\faEdit $\;$ framesI\_tweets task}
\label{app:taskname:framesItweets}

\begin{itemize}
\item \textbf{Task description:} General frame detection: whether a tweet contains a set of two opposing frames ("problem" and "solution"). The solution frame describes tweets framing content moderation as a solution to other issues (e.g., hate speech). The problem frame describes tweets framing content moderation as a problem on its own as well as to other issues (e.g., free speech).

\item \textbf{Prompts taken from}:~\citep{gilardi2023pnas, alizadeh2025open}

\item \textbf{Ground truth annotators:} Trained research assistants (political science students)

\item \textbf{Notes:} \citet{gilardi2023pnas} refer to this as the 'Frames I' task for the 'Tweets (2020--2021)' dataset. We employ identical data preprocessing as \citet{gilardi2023pnas}, which means that we consider only tweets for which all annotators agree on the ground truth label and that are relevant according to the 'relevance\_tweets' task. Additionally, we filter out duplicates.
\end{itemize}

\subsubsection{\faEdit $\;$ framesII\_tweets task}
\label{app:taskname:framesIItweets}

\begin{itemize}
\item \textbf{Task description:} Policy frame detection: whether a tweet contains a set of fourteen policy frames or other (i.e., 'Economy', 'Capacity and resources', 'Morality', 'Fairness and Equality', 'Constitutionality and Jurisprudence', 'Policy Prescription and Evaluation', 'Law and Order, Crime and Justice', 'Security and Defense', 'Health and Safety', 'Quality of Life', 'Cultural Identity', 'Public Opinion', 'Political', 'External Regulation and Reputation', or 'Other').

\item \textbf{Prompts taken from}:~\citep{gilardi2023pnas, alizadeh2025open}

\item \textbf{Ground truth annotators:} Trained research assistants (political science students)

\item \textbf{Notes:} \citet{gilardi2023pnas} refer to this as the 'Frames II' task for the 'Tweets (2020--2021)' dataset. We employ identical data preprocessing as \citet{gilardi2023pnas}, which means that we consider only tweets for which all annotators agree on the ground truth label and that are relevant according to the 'relevance\_tweets' task. Additionally, we filter out duplicates.
\end{itemize}

\subsubsection{\faEdit $\;$ stance\_tweets task}
\label{app:taskname:stancetweets}

\begin{itemize}
\item \textbf{Task description:} Stance detection: whether a tweet is in favor of, against, or neutral about repealing Section 230 (a piece of US legislation central to content moderation).

\item \textbf{Prompts taken from}:~\citep{gilardi2023pnas, alizadeh2025open}

\item \textbf{Ground truth annotators:} Trained research assistants (political science students)

\item \textbf{Notes:} \citet{gilardi2023pnas} refer to this as the 'Stance' task for the 'Tweets (2020--2021)' dataset. We employ identical data preprocessing as \citet{gilardi2023pnas}, which means that we consider only tweets for which all annotators agree on the ground truth label and that are relevant according to the 'relevance\_tweets' task. Additionally, we filter out duplicates.
\end{itemize}

\subsubsection{\faEdit $\;$ topic\_tweets task}
\label{app:taskname:topictweets}

\begin{itemize}
\item \textbf{Task description:} Topic detection: whether a tweet is about a set of six predefined topics (i.e., Section 230, Trump Ban, Complaint, Platform Policies, Twitter Support, and others).

\item \textbf{Prompts taken from}:~\citep{gilardi2023pnas, alizadeh2025open}

\item \textbf{Ground truth annotators:} Trained research assistants (political science students)

\item \textbf{Notes:} \citet{gilardi2023pnas} refer to this as the 'Topics' task for the 'Tweets (2020--2021)' dataset. We employ identical data preprocessing as \citet{gilardi2023pnas}, which means that we consider only tweets for which all annotators agree on the ground truth label and that are relevant according to the 'relevance\_tweets' task. Additionally, we filter out duplicates.
\end{itemize}

\subsection{\faDatabase $\;$ tweets23 dataset}
\label{app:datasetname:tweets23}

\begin{itemize}
\item \textbf{Data Type:} Tweets
\item \textbf{Dataset description:} This dataset is similar to the tweets dataset, but for January 2023. It includes a random sample of 500 tweets (of which 339 were in English) drawn from a dataset of 1.3 million tweets.
\item \textbf{Dataset Links:} \url{https://dataverse.harvard.edu/dataset.xhtml?persistentId=doi:10.7910/DVN/PQYF6M}
\item \textbf{Dataset citation:} \citet{gilardi2023pnas}
\end{itemize}

\subsubsection{\faEdit $\;$ relevance\_tweets23 task}
\label{app:taskname:relevancetweets23}

\begin{itemize}
\item \textbf{Task description:} Binary annotation of whether a tweet is about content moderation (relevant/irrelevant).

\item \textbf{Prompts taken from}:~\citep{gilardi2023pnas, alizadeh2025open}

\item \textbf{Ground truth annotators:} Trained research assistants (political science students)

\item \textbf{Notes:} \citet{gilardi2023pnas} refer to this as the 'Relevance' task for the 'Tweets (2023)' dataset. We employ identical data preprocessing as \citet{gilardi2023pnas}, which means that we consider only tweets for which all annotators agree on the ground truth label. Additionally, we filter out duplicates.
\end{itemize}

\subsubsection{\faEdit $\;$ framesI\_tweets23 task}
\label{app:taskname:framesItweets23}

\begin{itemize}
\item \textbf{Task description:} General frame detection: whether a tweet contains a set of two opposing frames ("problem" and "solution"). The solution frame describes tweets framing content moderation as a solution to other issues (e.g., hate speech). The problem frame describes tweets framing content moderation as a problem on its own as well as to other issues (e.g., free speech).

\item \textbf{Prompts taken from}:~\citep{gilardi2023pnas, alizadeh2025open}

\item \textbf{Ground truth annotators:} Trained research assistants (political science students)

\item \textbf{Notes:} \citet{gilardi2023pnas} refer to this as the 'Frames I' task for the 'Tweets (2023)' dataset. We employ identical data preprocessing as \citet{gilardi2023pnas}, which means that we consider only tweets for which all annotators agree on the ground truth label and that are relevant according to the 'relevance\_tweets23' task. Additionally, we filter out duplicates.
\end{itemize}

\subsection{\faDatabase $\;$ tweets17 dataset}
\label{app:datasetname:tweets17}

\begin{itemize}
\item \textbf{Data Type:} Tweets
\item \textbf{Dataset description:} A random sample of 1,856 tweets posted by members of the US Congress from 2017 to 2022, drawn from a dataset of 20 million tweets.
\item \textbf{Dataset Links:} \url{https://dataverse.harvard.edu/dataset.xhtml?persistentId=doi:10.7910/DVN/PQYF6M}
\item \textbf{Dataset citation:} \citet{gilardi2023pnas}
\end{itemize}

\subsubsection{\faEdit $\;$ relevance\_tweets17 task}
\label{app:taskname:relevancetweets17}

\begin{itemize}
\item \textbf{Task description:} Binary annotation of whether a tweet is relevant to political content (relevant/irrelevant).

\item \textbf{Prompts taken from}:~\citep{gilardi2023pnas, alizadeh2025open}

\item \textbf{Ground truth annotators:} Trained research assistants (political science students)

\item \textbf{Notes:} \citet{gilardi2023pnas} refer to this as the 'Relevance' task for the 'Tweets (2017--2023)' dataset. We employ identical data preprocessing as \citet{gilardi2023pnas}, which means that we consider only tweets for which all annotators agree on the ground truth label. Additionally, we filter out duplicates.
\end{itemize}

\subsubsection{\faEdit $\;$ framesII\_tweets17 task}
\label{app:taskname:framesIItweets17}

\begin{itemize}
\item \textbf{Task description:} Policy frame detection: whether a tweet contains a set of fourteen policy frames or other (i.e., 'Economy', 'Capacity and resources', 'Morality', 'Fairness and Equality', 'Constitutionality and Jurisprudence', 'Policy Prescription and Evaluation', 'Law and Order, Crime and Justice', 'Security and Defense', 'Health and Safety', 'Quality of Life', 'Cultural Identity', 'Public Opinion', 'Political', 'External Regulation and Reputation', or 'Other').

\item \textbf{Prompts taken from}:~\citep{gilardi2023pnas, alizadeh2025open}

\item \textbf{Ground truth annotators:} Trained research assistants (political science students)

\item \textbf{Notes:} \citet{gilardi2023pnas} refer to this as the 'Frames II' task for the 'Tweets (2017--2023)' dataset. We employ identical data preprocessing as \citet{gilardi2023pnas}, which means that we consider only tweets for which all annotators agree on the ground truth label and that are relevant according to the 'relevance\_tweets17' task. Additionally, we filter out duplicates.
\end{itemize}

\subsection{\faDatabase $\;$ news dataset}
\label{app:datasetname:news}

\begin{itemize}
\item \textbf{Data Type:} News articles
\item \textbf{Dataset description:} A random sample of 1,606 articles newspaper articles on content moderation published from January 2020 to April 2021, drawn from a dataset of 980k articles collected via LexisNexis.
\item \textbf{Dataset Links:} \url{https://dataverse.harvard.edu/dataset.xhtml?persistentId=doi:10.7910/DVN/PQYF6M}
\item \textbf{Dataset citation:} \citet{gilardi2023pnas}
\end{itemize}

\subsubsection{\faEdit $\;$ relevance\_news task}
\label{app:taskname:relevancenews}

\begin{itemize}
\item \textbf{Task description:} Binary annotation of whether a news article is about content moderation (relevant/irrelevant).

\item \textbf{Prompts taken from}:~\citep{gilardi2023pnas, alizadeh2025open}

\item \textbf{Ground truth annotators:} Trained research assistants (political science students)

\item \textbf{Notes:} \citet{gilardi2023pnas} refer to this as the 'Relevance' task for the 'News Articles (2020--2021)' dataset. We employ identical data preprocessing as \citet{gilardi2023pnas}, which means that we consider only tweets for which all annotators agree on the ground truth label. Additionally, we filter out duplicates.
\end{itemize}

\subsubsection{\faEdit $\;$ framesI\_news task}
\label{app:taskname:framesInews}

\begin{itemize}
\item \textbf{Task description:} General frame detection: whether a news article contains a set of two opposing frames ("problem" and "solution"). The solution frame describes news articles framing content moderation as a solution to other issues (e.g., hate speech). The problem frame describes news articles framing content moderation as a problem on its own as well as to other issues (e.g., free speech).

\item \textbf{Prompts taken from}:~\citep{gilardi2023pnas, alizadeh2025open}

\item \textbf{Ground truth annotators:} Trained research assistants (political science students)

\item \textbf{Notes:} \citet{gilardi2023pnas} refer to this as the 'Frames I' task for the 'News Articles (2020--2021)' dataset. We employ identical data preprocessing as \citet{gilardi2023pnas}, which means that we consider only news articles for which all annotators agree on the ground truth label and that are relevant according to the 'relevance\_news' task. Additionally, we filter out duplicates.
\end{itemize}

%% file: main.bbl
\begin{thebibliography}{182}
\providecommand{\natexlab}[1]{#1}
\providecommand{\url}[1]{\texttt{#1}}
\expandafter\ifx\csname urlstyle\endcsname\relax
  \providecommand{\doi}[1]{doi: #1}\else
  \providecommand{\doi}{doi: \begingroup \urlstyle{rm}\Url}\fi

\bibitem[Al~Kuwatly et~al.(2020)Al~Kuwatly, Wich, and Groh]{al-kuwatly-etal-2020-identifying}
Hala Al~Kuwatly, Maximilian Wich, and Georg Groh.
\newblock Identifying and measuring annotator bias based on annotators' demographic characteristics.
\newblock In Seyi Akiwowo, Bertie Vidgen, Vinodkumar Prabhakaran, and Zeerak Waseem, editors, \emph{Proceedings of the Fourth Workshop on Online Abuse and Harms}, pages 184--190, Online, November 2020. Association for Computational Linguistics.
\newblock \doi{10.18653/v1/2020.alw-1.21}.
\newblock URL \url{https://aclanthology.org/2020.alw-1.21/}.

\bibitem[Alizadeh et~al.(2025)Alizadeh, Kubli, Samei, Dehghani, Zahedivafa, Bermeo, Korobeynikova, and Gilardi]{alizadeh2025open}
Meysam Alizadeh, Maël Kubli, Zeynab Samei, Shirin Dehghani, Mohammadmasiha Zahedivafa, Juan~D Bermeo, Maria Korobeynikova, and Fabrizio Gilardi.
\newblock Open-source llms for text annotation: a practical guide for model setting and fine-tuning.
\newblock \emph{Journal of Computational Social Science}, 8\penalty0 (1), 2025.
\newblock \doi{10.1007/s42001-024-00345-9}.

\bibitem[Alonso~del Barrio and Gatica-Perez(2023)]{Barrio2023Framing}
David Alonso~del Barrio and Daniel Gatica-Perez.
\newblock Framing the news: From human perception to large language model inferences.
\newblock In \emph{Proceedings of the 2023 ACM International Conference on Multimedia Retrieval}, ICMR '23, page 627–635, New York, NY, USA, 2023. Association for Computing Machinery.
\newblock ISBN 9798400701788.
\newblock \doi{10.1145/3591106.3592278}.
\newblock URL \url{https://doi.org/10.1145/3591106.3592278}.

\bibitem[Ashwin et~al.(2025)Ashwin, Chhabra, and Rao]{Ashwin2025}
Julian Ashwin, Aditya Chhabra, and Vijayendra Rao.
\newblock Using large language models for qualitative analysis can introduce serious bias.
\newblock \emph{Sociological Methods \& Research}, 2025.
\newblock \doi{10.1177/00491241251338246}.
\newblock URL \url{https://doi.org/10.1177/00491241251338246}.

\bibitem[Atreja et~al.(2025)Atreja, Ashkinaze, Li, Mendelsohn, and Hemphill]{Atreja_Ashkinaze_Li_Mendelsohn_Hemphill_2025}
Shubham Atreja, Joshua Ashkinaze, Lingyao Li, Julia Mendelsohn, and Libby Hemphill.
\newblock What’s in a prompt?: A large-scale experiment to assess the impact of prompt design on the compliance and accuracy of llm-generated text annotations.
\newblock \emph{Proceedings of the International AAAI Conference on Web and Social Media}, 19\penalty0 (1):\penalty0 122--145, 2025.
\newblock \doi{10.1609/icwsm.v19i1.35807}.
\newblock URL \url{https://ojs.aaai.org/index.php/ICWSM/article/view/35807}.

\bibitem[Baly et~al.(2020)Baly, Da~San~Martino, Glass, and Nakov]{baly-etal-2020-detect}
Ramy Baly, Giovanni Da~San~Martino, James Glass, and Preslav Nakov.
\newblock We can detect your bias: Predicting the political ideology of news articles.
\newblock In Bonnie Webber, Trevor Cohn, Yulan He, and Yang Liu, editors, \emph{Proceedings of the 2020 Conference on Empirical Methods in Natural Language Processing (EMNLP)}, pages 4982--4991, Online, November 2020. Association for Computational Linguistics.
\newblock \doi{10.18653/v1/2020.emnlp-main.404}.
\newblock URL \url{https://aclanthology.org/2020.emnlp-main.404/}.

\bibitem[Barrie et~al.(2024{\natexlab{a}})Barrie, Palaiologou, and Törnberg]{barrie2024prompt}
Christopher Barrie, Elli Palaiologou, and Petter Törnberg.
\newblock Prompt stability scoring for text annotation with large language models.
\newblock \emph{arXiv preprint arXiv:2407.02039}, 2024{\natexlab{a}}.

\bibitem[Barrie et~al.(2024{\natexlab{b}})Barrie, Palmer, and Spirling]{barrie2024replication}
Christopher Barrie, Alexis Palmer, and Arthur Spirling.
\newblock Replication for language models problems, principles, and best practice for political science.
\newblock 2024{\natexlab{b}}.
\newblock URL \url{https://arthurspirling.org/documents/BarriePalmerSpirling_TrustMeBro.pdf}.

\bibitem[Battaglia et~al.(2024)Battaglia, Christensen, Hansen, and Sacher]{battaglia2024inference}
Laura Battaglia, Timothy Christensen, Stephen Hansen, and Szymon Sacher.
\newblock Inference for regression with variables generated from unstructured data.
\newblock 2024.
\newblock \doi{http://dx.doi.org/10.2139/ssrn.4842086}.

\bibitem[Benjamini and Hochberg(1995)]{Benjamini1995}
Yoav Benjamini and Yosef Hochberg.
\newblock Controlling the false discovery rate: A practical and powerful approach to multiple testing.
\newblock \emph{Journal of the Royal Statistical Society: Series B (Methodological)}, 57\penalty0 (1):\penalty0 289--300, 1995.
\newblock \doi{https://doi.org/10.1111/j.2517-6161.1995.tb02031.x}.
\newblock URL \url{https://rss.onlinelibrary.wiley.com/doi/abs/10.1111/j.2517-6161.1995.tb02031.x}.

\bibitem[Benoit et~al.(2016)Benoit, Conway, Lauderdale, Laver, and Mikhaylov]{benoit2016crowd}
Kenneth Benoit, Drew Conway, Benjamin~E Lauderdale, Michael Laver, and Slava Mikhaylov.
\newblock Crowd-sourced text analysis: Reproducible and agile production of political data.
\newblock \emph{American Political Science Review}, 110\penalty0 (2):\penalty0 278--295, 2016.

\bibitem[Bhattacharyya et~al.(2023)Bhattacharyya, Singla, Krishnamurthy, Shah, and Chen]{bhattacharyya-etal-2023-video}
Aanisha Bhattacharyya, Yaman~K Singla, Balaji Krishnamurthy, Rajiv~Ratn Shah, and Changyou Chen.
\newblock A video is worth 4096 tokens: Verbalize videos to understand them in zero shot.
\newblock In Houda Bouamor, Juan Pino, and Kalika Bali, editors, \emph{Proceedings of the 2023 Conference on Empirical Methods in Natural Language Processing}, pages 9822--9839, Singapore, December 2023. Association for Computational Linguistics.
\newblock \doi{10.18653/v1/2023.emnlp-main.608}.
\newblock URL \url{https://aclanthology.org/2023.emnlp-main.608/}.

\bibitem[Blei et~al.(2003)Blei, Ng, and Jordan]{Blei2003lds}
David~M. Blei, Andrew~Y. Ng, and Michael~I. Jordan.
\newblock Latent dirichlet allocation.
\newblock \emph{J. Mach. Learn. Res.}, 3\penalty0 (null):\penalty0 993–1022, March 2003.
\newblock ISSN 1532-4435.

\bibitem[Bosley et~al.(2025)Bosley, Kuzushima, Enamorado, and Shiraito]{BOSLEY_KUZUSHIMA_ENAMORADO_SHIRAITO_2025}
Mitchell Bosley, Saki Kuzushima, Ted Enamorado, and Yuki Shiraito.
\newblock Improving probabilistic models in text classification via active learning.
\newblock \emph{American Political Science Review}, 119\penalty0 (2):\penalty0 985–1002, 2025.
\newblock \doi{10.1017/S0003055424000716}.

\bibitem[Bozdag et~al.(2024)Bozdag, Sevim, and Ko\c{c}]{Bozdag2024}
Mustafa Bozdag, Nurullah Sevim, and Aykut Ko\c{c}.
\newblock Measuring and mitigating gender bias in legal contextualized language models.
\newblock \emph{ACM Trans. Knowl. Discov. Data}, 18\penalty0 (4), February 2024.
\newblock \doi{10.1145/3628602}.
\newblock URL \url{https://doi.org/10.1145/3628602}.

\bibitem[Breznau et~al.(2022)Breznau, Rinke, Wuttke, Nguyen, Adem, Adriaans, Alvarez-Benjumea, Andersen, Auer, Azevedo, et~al.]{Breznau2022HypothesisUncertainty}
Nate Breznau, Eike~Mark Rinke, Alexander Wuttke, Hung~HV Nguyen, Muna Adem, Jule Adriaans, Amalia Alvarez-Benjumea, Henrik~K Andersen, Daniel Auer, Flavio Azevedo, et~al.
\newblock Observing many researchers using the same data and hypothesis reveals a hidden universe of uncertainty.
\newblock \emph{Proceedings of the National Academy of Sciences}, 119\penalty0 (44):\penalty0 e2203150119, 2022.
\newblock \doi{10.1073/pnas.2203150119}.
\newblock URL \url{https://www.pnas.org/doi/abs/10.1073/pnas.2203150119}.

\bibitem[Briggs et~al.(2025)Briggs, Mellon, Arel-Bundock, and Larson]{Briggs2025}
Ryan Briggs, Jonathan Mellon, Vincent Arel-Bundock, and Tim Larson.
\newblock We used llms to track methodological and substantive publication patterns in political science and they seem to do a pretty good job.
\newblock \emph{OSF preprint https://osf.io/v7fe8}, 2025.

\bibitem[Bucher and Martini(2024)]{bucher2024fine}
Martin Juan~Jos{\'e} Bucher and Marco Martini.
\newblock Fine-tuned'small'llms (still) significantly outperform zero-shot generative ai models in text classification.
\newblock \emph{arXiv preprint arXiv:2406.08660}, 2024.

\bibitem[Calderon et~al.(2025)Calderon, Reichart, and Dror]{calderon-etal-2025-alternative}
Nitay Calderon, Roi Reichart, and Rotem Dror.
\newblock The alternative annotator test for {LLM}-as-a-judge: How to statistically justify replacing human annotators with {LLM}s.
\newblock In \emph{Proceedings of the 63rd Annual Meeting of the Association for Computational Linguistics (Volume 1: Long Papers)}, pages 16051--16081. Association for Computational Linguistics, 2025.
\newblock \doi{10.18653/v1/2025.acl-long.782}.
\newblock URL \url{https://aclanthology.org/2025.acl-long.782/}.

\bibitem[Carlson and Montgomery(2017)]{carlson2017pairwise}
David Carlson and Jacob~M Montgomery.
\newblock A pairwise comparison framework for fast, flexible, and reliable human coding of political texts.
\newblock \emph{American Political Science Review}, 111\penalty0 (4):\penalty0 835--843, 2017.

\bibitem[Carlson and Dell(2025)]{carlson2025unifying}
Jacob Carlson and Melissa Dell.
\newblock A unifying framework for robust and efficient inference with unstructured data.
\newblock \emph{arXiv preprint arXiv:2505.00282}, 2025.

\bibitem[Chae and Davidson(2023)]{chae2023large}
Youngjin Chae and Thomas Davidson.
\newblock Large language models for text classification: From zero-shot learning to fine-tuning.
\newblock \emph{Open Science Foundation}, 10, 2023.

\bibitem[Chan et~al.(2024)Chan, Chen, Su, Yu, Xue, Zhang, Fu, and Liu]{chan2024chateval}
Chi-Min Chan, Weize Chen, Yusheng Su, Jianxuan Yu, Wei Xue, Shanghang Zhang, Jie Fu, and Zhiyuan Liu.
\newblock Chateval: Towards better {LLM}-based evaluators through multi-agent debate.
\newblock In \emph{The Twelfth International Conference on Learning Representations}, 2024.
\newblock URL \url{https://openreview.net/forum?id=FQepisCUWu}.

\bibitem[chen et~al.(2023)chen, Zhao, Zhang, Chern, Gao, Liu, and He]{NEURIPS2023_8b8a7960}
shiqi chen, Yiran Zhao, Jinghan Zhang, I-Chun Chern, Siyang Gao, Pengfei Liu, and Junxian He.
\newblock Felm: Benchmarking factuality evaluation of large language models.
\newblock In A.~Oh, T.~Naumann, A.~Globerson, K.~Saenko, M.~Hardt, and S.~Levine, editors, \emph{Advances in Neural Information Processing Systems}, volume~36, pages 44502--44523, 2023.
\newblock URL \url{https://proceedings.neurips.cc/paper_files/paper/2023/file/8b8a7960d343e023a6a0afe37eee6022-Paper-Datasets_and_Benchmarks.pdf}.

\bibitem[Chew et~al.(2023)Chew, Bollenbacher, Wenger, Speer, and Kim]{chew2023llm}
Robert Chew, John Bollenbacher, Michael Wenger, Jessica Speer, and Annice Kim.
\newblock Llm-assisted content analysis: Using large language models to support deductive coding.
\newblock \emph{arXiv preprint arXiv:2306.14924}, 2023.

\bibitem[Chhun et~al.(2022)Chhun, Colombo, Suchanek, and Clavel]{chhun-etal-2022-human}
Cyril Chhun, Pierre Colombo, Fabian~M. Suchanek, and Chlo{\'e} Clavel.
\newblock Of human criteria and automatic metrics: A benchmark of the evaluation of story generation.
\newblock In Nicoletta Calzolari, Chu-Ren Huang, Hansaem Kim, James Pustejovsky, Leo Wanner, Key-Sun Choi, Pum-Mo Ryu, Hsin-Hsi Chen, Lucia Donatelli, Heng Ji, Sadao Kurohashi, Patrizia Paggio, Nianwen Xue, Seokhwan Kim, Younggyun Hahm, Zhong He, Tony~Kyungil Lee, Enrico Santus, Francis Bond, and Seung-Hoon Na, editors, \emph{Proceedings of the 29th International Conference on Computational Linguistics}, pages 5794--5836, Gyeongju, Republic of Korea, October 2022. International Committee on Computational Linguistics.
\newblock URL \url{https://aclanthology.org/2022.coling-1.509/}.

\bibitem[Chiang and Lee(2023)]{chiang-lee-2023-large}
Cheng-Han Chiang and Hung-yi Lee.
\newblock Can large language models be an alternative to human evaluations?
\newblock In Anna Rogers, Jordan Boyd-Graber, and Naoaki Okazaki, editors, \emph{Proceedings of the 61st Annual Meeting of the Association for Computational Linguistics (Volume 1: Long Papers)}, pages 15607--15631, Toronto, Canada, July 2023. Association for Computational Linguistics.
\newblock \doi{10.18653/v1/2023.acl-long.870}.
\newblock URL \url{https://aclanthology.org/2023.acl-long.870/}.

\bibitem[Choi and Ferrara(2024)]{Choi2024Automated}
Eun~Cheol Choi and Emilio Ferrara.
\newblock Automated claim matching with large language models: Empowering fact-checkers in the fight against misinformation.
\newblock In \emph{Companion Proceedings of the ACM Web Conference 2024}, WWW '24, page 1441–1449, New York, NY, USA, 2024. Association for Computing Machinery.
\newblock ISBN 9798400701726.
\newblock \doi{10.1145/3589335.3651910}.
\newblock URL \url{https://doi.org/10.1145/3589335.3651910}.

\bibitem[Danescu-Niculescu-Mizil et~al.(2013)Danescu-Niculescu-Mizil, Sudhof, Jurafsky, Leskovec, and Potts]{danescu-niculescu-mizil-etal-2013-computational}
Cristian Danescu-Niculescu-Mizil, Moritz Sudhof, Dan Jurafsky, Jure Leskovec, and Christopher Potts.
\newblock A computational approach to politeness with application to social factors.
\newblock In Hinrich Schuetze, Pascale Fung, and Massimo Poesio, editors, \emph{Proceedings of the 51st Annual Meeting of the Association for Computational Linguistics (Volume 1: Long Papers)}, pages 250--259, Sofia, Bulgaria, August 2013. Association for Computational Linguistics.
\newblock URL \url{https://aclanthology.org/P13-1025/}.

\bibitem[Davani et~al.(2024)Davani, D\'{\i}az, Baker, and Prabhakaran]{Davani2024}
Aida Davani, Mark D\'{\i}az, Dylan Baker, and Vinodkumar Prabhakaran.
\newblock Disentangling perceptions of offensiveness: Cultural and moral correlates.
\newblock In \emph{Proceedings of the 2024 ACM Conference on Fairness, Accountability, and Transparency}, FAccT '24, page 2007–2021, New York, NY, USA, 2024. Association for Computing Machinery.
\newblock ISBN 9798400704505.
\newblock \doi{10.1145/3630106.3659021}.
\newblock URL \url{https://doi.org/10.1145/3630106.3659021}.

\bibitem[Davidson(2024)]{davidson2024start}
Thomas Davidson.
\newblock Start generating: Harnessing generative artificial intelligence for sociological research.
\newblock \emph{Socius}, 10, 2024.

\bibitem[Demszky et~al.(2020)Demszky, Movshovitz-Attias, Ko, Cowen, Nemade, and Ravi]{demszky-etal-2020-goemotions}
Dorottya Demszky, Dana Movshovitz-Attias, Jeongwoo Ko, Alan Cowen, Gaurav Nemade, and Sujith Ravi.
\newblock {G}o{E}motions: A dataset of fine-grained emotions.
\newblock In Dan Jurafsky, Joyce Chai, Natalie Schluter, and Joel Tetreault, editors, \emph{Proceedings of the 58th Annual Meeting of the Association for Computational Linguistics}, pages 4040--4054, Online, July 2020. Association for Computational Linguistics.
\newblock \doi{10.18653/v1/2020.acl-main.372}.
\newblock URL \url{https://aclanthology.org/2020.acl-main.372/}.

\bibitem[Demszky et~al.(2023)Demszky, Yang, Yeager, Bryan, Clapper, Chandhok, Eichstaedt, Hecht, Jamieson, Johnson, et~al.]{demszky2023using}
Dorottya Demszky, Diyi Yang, David~S Yeager, Christopher~J Bryan, Margarett Clapper, Susannah Chandhok, Johannes~C Eichstaedt, Cameron Hecht, Jeremy Jamieson, Meghann Johnson, et~al.
\newblock Using large language models in psychology.
\newblock \emph{Nature Reviews Psychology}, 2\penalty0 (11):\penalty0 688--701, 2023.

\bibitem[Deroy and Maity(2023)]{deroy2023questioning}
Aniket Deroy and Subhankar Maity.
\newblock Questioning biases in case judgment summaries: Legal datasets or large language models?
\newblock \emph{arXiv preprint arXiv:2312.00554}, 2023.

\bibitem[dos Santos et~al.(2024)dos Santos, Santos, Lynn, and Benatallah]{dosSantos2024}
Vitor~Gaboardi dos Santos, Guto~Leoni Santos, Theo Lynn, and Boualem Benatallah.
\newblock Identifying citizen-related issues from social media using llm-based data augmentation.
\newblock In Giancarlo Guizzardi, Flavia Santoro, Haralambos Mouratidis, and Pnina Soffer, editors, \emph{Advanced Information Systems Engineering}, pages 531--546, Cham, 2024. Springer Nature Switzerland.
\newblock ISBN 978-3-031-61057-8.

\bibitem[Efron(2004)]{Efron2004}
Bradley Efron.
\newblock Large-scale simultaneous hypothesis testing.
\newblock \emph{Journal of the American Statistical Association}, 99\penalty0 (465):\penalty0 96--104, 2004.
\newblock \doi{10.1198/016214504000000089}.
\newblock URL \url{https://doi.org/10.1198/016214504000000089}.

\bibitem[Efron(2005)]{efron2005local}
Bradley Efron.
\newblock Local false discovery rates, 2005.

\bibitem[Egami et~al.(2023)Egami, Hinck, Stewart, and Wei]{egami2023using}
Naoki Egami, Musashi Hinck, Brandon Stewart, and Hanying Wei.
\newblock Using imperfect surrogates for downstream inference: Design-based supervised learning for social science applications of large language models.
\newblock In A.~Oh, T.~Naumann, A.~Globerson, K.~Saenko, M.~Hardt, and S.~Levine, editors, \emph{Advances in Neural Information Processing Systems}, volume~36, pages 68589--68601. Curran Associates, Inc., 2023.
\newblock URL \url{https://proceedings.neurips.cc/paper_files/paper/2023/file/d862f7f5445255090de13b825b880d59-Paper-Conference.pdf}.

\bibitem[Egami et~al.(2024)Egami, Hinck, Stewart, and Wei]{egami2024using}
Naoki Egami, Musashi Hinck, Brandon~M Stewart, and Hanying Wei.
\newblock Using large language model annotations for the social sciences: A general framework of using predicted variables in downstream analyses.
\newblock 2024.
\newblock URL \url{https://naokiegami.com/paper/dsl_ss.pdf}.

\bibitem[Egami et~al.(2025)Egami, Hinck, {M. Stewart}, and Wei]{dslRpackage}
Naoki Egami, Musashi Hinck, Brandon {M. Stewart}, and Hanying Wei.
\newblock \emph{dsl: Design-based Supervised Learning}, 2025.
\newblock URL \url{http://naokiegami.com/dsl/}.
\newblock R package version 0.1.0.

\bibitem[ElSherief et~al.(2021)ElSherief, Ziems, Muchlinski, Anupindi, Seybolt, De~Choudhury, and Yang]{elsherief-etal-2021-latent}
Mai ElSherief, Caleb Ziems, David Muchlinski, Vaishnavi Anupindi, Jordyn Seybolt, Munmun De~Choudhury, and Diyi Yang.
\newblock Latent hatred: A benchmark for understanding implicit hate speech.
\newblock In Marie-Francine Moens, Xuanjing Huang, Lucia Specia, and Scott Wen-tau Yih, editors, \emph{Proceedings of the 2021 Conference on Empirical Methods in Natural Language Processing}, pages 345--363, Online and Punta Cana, Dominican Republic, November 2021. Association for Computational Linguistics.
\newblock \doi{10.18653/v1/2021.emnlp-main.29}.
\newblock URL \url{https://aclanthology.org/2021.emnlp-main.29/}.

\bibitem[Fatemi et~al.(2023)Fatemi, Rabbi, and Opdahl]{Fatemi2023}
Bahareh Fatemi, Fazle Rabbi, and Andreas~L. Opdahl.
\newblock Evaluating the effectiveness of gpt large language model for news classification in the iptc news ontology.
\newblock \emph{IEEE Access}, 11:\penalty0 145386--145394, 2023.
\newblock \doi{10.1109/ACCESS.2023.3345414}.

\bibitem[Felkner et~al.(2024)Felkner, Thompson, and May]{felkner-etal-2024-gpt}
Virginia Felkner, Jennifer Thompson, and Jonathan May.
\newblock {GPT} is not an annotator: The necessity of human annotation in fairness benchmark construction.
\newblock In Lun-Wei Ku, Andre Martins, and Vivek Srikumar, editors, \emph{Proceedings of the 62nd Annual Meeting of the Association for Computational Linguistics (Volume 1: Long Papers)}, pages 14104--14115, Bangkok, Thailand, August 2024. Association for Computational Linguistics.
\newblock \doi{10.18653/v1/2024.acl-long.760}.
\newblock URL \url{https://aclanthology.org/2024.acl-long.760/}.

\bibitem[Feuerriegel et~al.(2025)Feuerriegel, Maarouf, B{\"a}r, Geissler, Schweisthal, Pr{\"o}llochs, Robertson, Rathje, Hartmann, Mohammad, et~al.]{feuerriegel2025using}
Stefan Feuerriegel, Abdurahman Maarouf, Dominik B{\"a}r, Dominique Geissler, Jonas Schweisthal, Nicolas Pr{\"o}llochs, Claire~E Robertson, Steve Rathje, Jochen Hartmann, Saif~M Mohammad, et~al.
\newblock Using natural language processing to analyse text data in behavioural science.
\newblock \emph{Nature Reviews Psychology}, 4\penalty0 (2):\penalty0 96--111, 2025.

\bibitem[Fieldhouse et~al.(2024)Fieldhouse, Green, Evans, Mellon, Prosser, Bailey, Griffiths, and Perrett]{fieldhouse2024british}
Edward Fieldhouse, Jane Green, Geoffrey Evans, Jonathan Mellon, Christopher Prosser, Jack Bailey, James Griffiths, and Stuart Perrett.
\newblock British election study internet panel waves 1-29.
\newblock \emph{The University of Manchester, Manchester}, 2024.
\newblock \doi{10.5255/UKDA-SN-8202-2}.

\bibitem[Fu et~al.(2024)Fu, Hsu, Chan, Lau, Liu, and Yip]{Fu2024}
Ziru Fu, Yu~Cheng Hsu, Christian~S Chan, Chaak~Ming Lau, Joyce Liu, and Paul Siu~Fai Yip.
\newblock Efficacy of chatgpt in cantonese sentiment analysis: Comparative study.
\newblock \emph{J Med Internet Res}, 26, Jan 2024.
\newblock \doi{10.2196/51069}.
\newblock URL \url{https://www.jmir.org/2024/1/e51069}.

\bibitem[Gabriel et~al.(2022)Gabriel, Hallinan, Sap, Nguyen, Roesner, Choi, and Choi]{gabriel-etal-2022-misinfo}
Saadia Gabriel, Skyler Hallinan, Maarten Sap, Pemi Nguyen, Franziska Roesner, Eunsol Choi, and Yejin Choi.
\newblock Misinfo reaction frames: Reasoning about readers' reactions to news headlines.
\newblock In Smaranda Muresan, Preslav Nakov, and Aline Villavicencio, editors, \emph{Proceedings of the 60th Annual Meeting of the Association for Computational Linguistics (Volume 1: Long Papers)}, pages 3108--3127, Dublin, Ireland, 2022. Association for Computational Linguistics.
\newblock \doi{10.18653/v1/2022.acl-long.222}.
\newblock URL \url{https://aclanthology.org/2022.acl-long.222/}.

\bibitem[Gelman and Carlin(2014)]{Gelman2014Beyond}
Andrew Gelman and John Carlin.
\newblock Beyond power calculations: Assessing type s (sign) and type m (magnitude) errors.
\newblock \emph{Perspectives on Psychological Science}, 9\penalty0 (6):\penalty0 641--651, 2014.
\newblock \doi{10.1177/1745691614551642}.
\newblock URL \url{https://doi.org/10.1177/1745691614551642}.

\bibitem[Gelman and Loken(2013)]{gelman2013garden}
Andrew Gelman and Eric Loken.
\newblock The garden of forking paths: Why multiple comparisons can be a problem, even when there is no “fishing expedition” or “p-hacking” and the research hypothesis was posited ahead of time.
\newblock \emph{Department of Statistics, Columbia University}, 348\penalty0 (1-17):\penalty0 3, 2013.

\bibitem[Gelman and Stern(2006)]{Gelman2006Significant}
Andrew Gelman and Hal Stern.
\newblock The difference between “significant” and “not significant” is not itself statistically significant.
\newblock \emph{The American Statistician}, 60\penalty0 (4):\penalty0 328--331, 2006.
\newblock \doi{10.1198/000313006X152649}.
\newblock URL \url{https://doi.org/10.1198/000313006X152649}.

\bibitem[{Gemma Team, Google DeepMind}(2025)]{team2025gemma}
{Gemma Team, Google DeepMind}.
\newblock Gemma 3 technical report.
\newblock \emph{arXiv preprint arXiv:2503.19786}, 2025.

\bibitem[Geva et~al.(2019)Geva, Goldberg, and Berant]{geva-etal-2019-modeling}
Mor Geva, Yoav Goldberg, and Jonathan Berant.
\newblock Are we modeling the task or the annotator? an investigation of annotator bias in natural language understanding datasets.
\newblock In Kentaro Inui, Jing Jiang, Vincent Ng, and Xiaojun Wan, editors, \emph{Proceedings of the 2019 Conference on Empirical Methods in Natural Language Processing and the 9th International Joint Conference on Natural Language Processing (EMNLP-IJCNLP)}, pages 1161--1166, Hong Kong, China, November 2019. Association for Computational Linguistics.
\newblock \doi{10.18653/v1/D19-1107}.
\newblock URL \url{https://aclanthology.org/D19-1107/}.

\bibitem[Gilardi et~al.(2023)Gilardi, Alizadeh, and Kubli]{gilardi2023pnas}
Fabrizio Gilardi, Meysam Alizadeh, and Maël Kubli.
\newblock Chatgpt outperforms crowd workers for text-annotation tasks.
\newblock \emph{Proceedings of the National Academy of Sciences}, 120\penalty0 (30):\penalty0 e2305016120, 2023.
\newblock \doi{10.1073/pnas.2305016120}.
\newblock URL \url{https://www.pnas.org/doi/abs/10.1073/pnas.2305016120}.

\bibitem[Gligoric et~al.(2025)Gligoric, Zrnic, Lee, Candes, and Jurafsky]{gligoric-etal-2025-unconfident}
Kristina Gligoric, Tijana Zrnic, Cinoo Lee, Emmanuel Candes, and Dan Jurafsky.
\newblock Can unconfident {LLM} annotations be used for confident conclusions?
\newblock In Luis Chiruzzo, Alan Ritter, and Lu~Wang, editors, \emph{Proceedings of the 2025 Conference of the Nations of the Americas Chapter of the Association for Computational Linguistics: Human Language Technologies (Volume 1: Long Papers)}, pages 3514--3533, Albuquerque, New Mexico, April 2025. Association for Computational Linguistics.
\newblock ISBN 979-8-89176-189-6.
\newblock URL \url{https://aclanthology.org/2025.naacl-long.179/}.

\bibitem[Goldfarb and King(2016)]{Goldfarb2016}
Brent Goldfarb and Andrew~A. King.
\newblock Scientific apophenia in strategic management research: Significance tests \& mistaken inference.
\newblock \emph{Strategic Management Journal}, 37\penalty0 (1):\penalty0 167--176, 2016.
\newblock \doi{https://doi.org/10.1002/smj.2459}.
\newblock URL \url{https://sms.onlinelibrary.wiley.com/doi/abs/10.1002/smj.2459}.

\bibitem[Grimmer and Stewart(2013)]{grimmer2013text}
Justin Grimmer and Brandon~M Stewart.
\newblock Text as data: The promise and pitfalls of automatic content analysis methods for political texts.
\newblock \emph{Political analysis}, 21\penalty0 (3):\penalty0 267--297, 2013.

\bibitem[Groemping(2006)]{relaimpo}
Ulrike Groemping.
\newblock Relative importance for linear regression in r: The package relaimpo.
\newblock \emph{Journal of Statistical Software}, 17\penalty0 (1):\penalty0 1--27, 2006.

\bibitem[Gu et~al.(2024)Gu, Zhu, Ye, Zhang, Wang, Zhu, Jiang, Xiong, Li, Wu, He, Xu, Huang, Liu, Wang, Wang, Zheng, Feng, and Xiao]{Gu_Zhu_Ye_2024}
Zhouhong Gu, Xiaoxuan Zhu, Haoning Ye, Lin Zhang, Jianchen Wang, Yixin Zhu, Sihang Jiang, Zhuozhi Xiong, Zihan Li, Weijie Wu, Qianyu He, Rui Xu, Wenhao Huang, Jingping Liu, Zili Wang, Shusen Wang, Weiguo Zheng, Hongwei Feng, and Yanghua Xiao.
\newblock Xiezhi: An ever-updating benchmark for holistic domain knowledge evaluation.
\newblock \emph{Proceedings of the AAAI Conference on Artificial Intelligence}, 38\penalty0 (16):\penalty0 18099--18107, Mar. 2024.
\newblock \doi{10.1609/aaai.v38i16.29767}.
\newblock URL \url{https://ojs.aaai.org/index.php/AAAI/article/view/29767}.

\bibitem[Halterman and Keith(2025)]{Halterman_Keith_2025}
Andrew Halterman and Katherine~A. Keith.
\newblock Codebook llms: Evaluating llms as measurement tools for political science concepts.
\newblock \emph{Political Analysis}, page 1–17, 2025.
\newblock \doi{10.1017/pan.2025.10017}.

\bibitem[He et~al.(2024{\natexlab{a}})He, Lin, Gong, Jin, Zhang, Lin, Jiao, Yiu, Duan, and Chen]{he-etal-2024-annollm}
Xingwei He, Zhenghao Lin, Yeyun Gong, A-Long Jin, Hang Zhang, Chen Lin, Jian Jiao, Siu~Ming Yiu, Nan Duan, and Weizhu Chen.
\newblock {A}nno{LLM}: Making large language models to be better crowdsourced annotators.
\newblock In Yi~Yang, Aida Davani, Avi Sil, and Anoop Kumar, editors, \emph{Proceedings of the 2024 Conference of the North American Chapter of the Association for Computational Linguistics: Human Language Technologies (Volume 6: Industry Track)}, pages 165--190, Mexico City, Mexico, June 2024{\natexlab{a}}. Association for Computational Linguistics.
\newblock \doi{10.18653/v1/2024.naacl-industry.15}.
\newblock URL \url{https://aclanthology.org/2024.naacl-industry.15/}.

\bibitem[He et~al.(2024{\natexlab{b}})He, Huang, Ding, Rohatgi, and Huang]{He2024If}
Zeyu He, Chieh-Yang Huang, Chien-Kuang~Cornelia Ding, Shaurya Rohatgi, and Ting-Hao~Kenneth Huang.
\newblock If in a crowdsourced data annotation pipeline, a gpt-4.
\newblock In \emph{Proceedings of the 2024 CHI Conference on Human Factors in Computing Systems}, CHI '24, New York, NY, USA, 2024{\natexlab{b}}. Association for Computing Machinery.
\newblock ISBN 9798400703300.
\newblock \doi{10.1145/3613904.3642834}.
\newblock URL \url{https://doi.org/10.1145/3613904.3642834}.

\bibitem[Head et~al.(2015{\natexlab{a}})Head, Holman, Lanfear, Kahn, and Jennions]{Head2015phacking}
Megan~L. Head, Luke Holman, Rob Lanfear, Andrew~T. Kahn, and Michael~D. Jennions.
\newblock The extent and consequences of p-hacking in science.
\newblock \emph{PLOS Biology}, 13\penalty0 (3):\penalty0 1--15, 03 2015{\natexlab{a}}.
\newblock \doi{10.1371/journal.pbio.1002106}.
\newblock URL \url{https://doi.org/10.1371/journal.pbio.1002106}.

\bibitem[Head et~al.(2015{\natexlab{b}})Head, Holman, Lanfear, Kahn, and Jennions]{head2015extent}
Megan~L Head, Luke Holman, Rob Lanfear, Andrew~T Kahn, and Michael~D Jennions.
\newblock The extent and consequences of p-hacking in science.
\newblock \emph{PLoS biology}, 13\penalty0 (3):\penalty0 e1002106, 2015{\natexlab{b}}.

\bibitem[Heseltine and von Hohenberg(2024)]{Heseltine2024}
Michael Heseltine and Bernhard~Clemm von Hohenberg.
\newblock Large language models as a substitute for human experts in annotating political text.
\newblock \emph{Research \& Politics}, 11\penalty0 (1):\penalty0 20531680241236239, 2024.
\newblock \doi{10.1177/20531680241236239}.
\newblock URL \url{https://doi.org/10.1177/20531680241236239}.

\bibitem[Hitchcock and Sober(2004)]{Hitchcock2004}
Christopher Hitchcock and Elliott Sober.
\newblock Prediction versus accommodation and the risk of overfitting.
\newblock \emph{The British Journal for the Philosophy of Science}, 55\penalty0 (1):\penalty0 1--34, 2004.
\newblock \doi{10.1093/bjps/55.1.1}.
\newblock URL \url{https://doi.org/10.1093/bjps/55.1.1}.

\bibitem[Hoes et~al.(2023)Hoes, Altay, and Bermeo]{hoes2023leveraging}
Emma Hoes, Sacha Altay, and Juan Bermeo.
\newblock Leveraging chatgpt for efficient fact-checking.
\newblock \emph{PsyArXiv. April}, 3, 2023.

\bibitem[Hussain et~al.(2024)Hussain, Binz, Mata, and Wulff]{hussain2024tutorial}
Zak Hussain, Marcel Binz, Rui Mata, and Dirk~U Wulff.
\newblock A tutorial on open-source large language models for behavioral science.
\newblock \emph{Behavior Research Methods}, 56\penalty0 (8):\penalty0 8214--8237, 2024.

\bibitem[Ioannidis(2008)]{ioannidis2008most}
John~PA Ioannidis.
\newblock Why most discovered true associations are inflated.
\newblock \emph{Epidemiology}, 19\penalty0 (5):\penalty0 640--648, 2008.

\bibitem[Jiang et~al.(2024)Jiang, Tan, Nirmal, and Liu]{Jiang2024Disinformation}
Bohan Jiang, Zhen Tan, Ayushi Nirmal, and Huan Liu.
\newblock \emph{Disinformation Detection: An Evolving Challenge in the Age of LLMs}, pages 427--435.
\newblock 2024.
\newblock \doi{10.1137/1.9781611978032.50}.
\newblock URL \url{https://epubs.siam.org/doi/abs/10.1137/1.9781611978032.50}.

\bibitem[Jin et~al.(2024)Jin, Choi, Verma, Wang, and Kumar]{jin-etal-2024-mm}
Yiqiao Jin, Minje Choi, Gaurav Verma, Jindong Wang, and Srijan Kumar.
\newblock {MM}-{SOC}: Benchmarking multimodal large language models in social media platforms.
\newblock In Lun-Wei Ku, Andre Martins, and Vivek Srikumar, editors, \emph{Findings of the Association for Computational Linguistics: ACL 2024}, pages 6192--6210, Bangkok, Thailand, August 2024. Association for Computational Linguistics.
\newblock \doi{10.18653/v1/2024.findings-acl.370}.
\newblock URL \url{https://aclanthology.org/2024.findings-acl.370/}.

\bibitem[Kadavath et~al.(2022)Kadavath, Conerly, Askell, Henighan, Drain, Perez, Schiefer, Hatfield-Dodds, DasSarma, Tran-Johnson, et~al.]{kadavath2022language}
Saurav Kadavath, Tom Conerly, Amanda Askell, Tom Henighan, Dawn Drain, Ethan Perez, Nicholas Schiefer, Zac Hatfield-Dodds, Nova DasSarma, Eli Tran-Johnson, et~al.
\newblock Language models (mostly) know what they know.
\newblock \emph{arXiv preprint arXiv:2207.05221}, 2022.

\bibitem[Karjus(2025)]{karjus2025machine}
Andres Karjus.
\newblock Machine-assisted quantitizing designs: augmenting humanities and social sciences with artificial intelligence.
\newblock \emph{Humanities and Social Sciences Communications}, 12\penalty0 (1):\penalty0 1--18, 2025.

\bibitem[Karjus and Cuskley(2024)]{karjus2024evolving}
Andres Karjus and Christine Cuskley.
\newblock Evolving linguistic divergence on polarizing social media.
\newblock \emph{Humanities and Social Sciences Communications}, 11\penalty0 (1):\penalty0 1--14, 2024.

\bibitem[Kerr(1998)]{kerr1998harking}
Norbert~L. Kerr.
\newblock Harking: Hypothesizing after the results are known.
\newblock \emph{Personality and Social Psychology Review}, 2\penalty0 (3):\penalty0 196--217, 1998.
\newblock \doi{10.1207/s15327957pspr0203\_4}.
\newblock URL \url{https://doi.org/10.1207/s15327957pspr0203_4}.
\newblock PMID: 15647155.

\bibitem[Kholodna et~al.(2024)Kholodna, Julka, Khodadadi, Gumus, and Granitzer]{Kholodna2024}
Nataliia Kholodna, Sahib Julka, Mohammad Khodadadi, Muhammed~Nurullah Gumus, and Michael Granitzer.
\newblock Llms in the loop: Leveraging large language model annotations for active learning in low-resource languages.
\newblock In \emph{Machine Learning and Knowledge Discovery in Databases. Applied Data Science Track}, pages 397--412, Cham, 2024. Springer Nature Switzerland.
\newblock ISBN 978-3-031-70381-2.

\bibitem[Khondaker et~al.(2023)Khondaker, Waheed, Nagoudi, and Abdul-Mageed]{khondaker-etal-2023-gptaraeval}
Md~Tawkat~Islam Khondaker, Abdul Waheed, El~Moatez~Billah Nagoudi, and Muhammad Abdul-Mageed.
\newblock {GPTA}ra{E}val: A comprehensive evaluation of {C}hat{GPT} on {A}rabic {NLP}.
\newblock In Houda Bouamor, Juan Pino, and Kalika Bali, editors, \emph{Proceedings of the 2023 Conference on Empirical Methods in Natural Language Processing}, pages 220--247, Singapore, December 2023. Association for Computational Linguistics.
\newblock \doi{10.18653/v1/2023.emnlp-main.16}.
\newblock URL \url{https://aclanthology.org/2023.emnlp-main.16/}.

\bibitem[Kim et~al.(2024)Kim, Mitra, Li~Chen, Rahman, and Zhang]{kim-etal-2024-meganno}
Hannah Kim, Kushan Mitra, Rafael Li~Chen, Sajjadur Rahman, and Dan Zhang.
\newblock {MEGA}nno+: A human-{LLM} collaborative annotation system.
\newblock In Nikolaos Aletras and Orphee De~Clercq, editors, \emph{Proceedings of the 18th Conference of the European Chapter of the Association for Computational Linguistics: System Demonstrations}, pages 168--176, St. Julians, Malta, March 2024. Association for Computational Linguistics.
\newblock URL \url{https://aclanthology.org/2024.eacl-demo.18/}.

\bibitem[Krippendorff(2018)]{krippendorff2018content}
Klaus Krippendorff.
\newblock \emph{Content analysis: An introduction to its methodology}.
\newblock Sage publications, 2018.

\bibitem[Krugmann and Hartmann(2024)]{krugmann2024sentiment}
Jan~Ole Krugmann and Jochen Hartmann.
\newblock Sentiment analysis in the age of generative ai.
\newblock \emph{Customer Needs and Solutions}, 11\penalty0 (1):\penalty0 3, 2024.

\bibitem[Kumar et~al.(2024)Kumar, AbuHashem, and Durumeric]{Kumar_AbuHashem_Durumeric_2024}
Deepak Kumar, Yousef~Anees AbuHashem, and Zakir Durumeric.
\newblock Watch your language: Investigating content moderation with large language models.
\newblock \emph{Proceedings of the International AAAI Conference on Web and Social Media}, 18\penalty0 (1):\penalty0 865--878, May 2024.
\newblock \doi{10.1609/icwsm.v18i1.31358}.
\newblock URL \url{https://ojs.aaai.org/index.php/ICWSM/article/view/31358}.

\bibitem[Laskar et~al.(2023{\natexlab{a}})Laskar, Bari, Rahman, Bhuiyan, Joty, and Huang]{laskar-etal-2023-systematic}
Md~Tahmid~Rahman Laskar, M~Saiful Bari, Mizanur Rahman, Md~Amran~Hossen Bhuiyan, Shafiq Joty, and Jimmy Huang.
\newblock A systematic study and comprehensive evaluation of {C}hat{GPT} on benchmark datasets.
\newblock In Anna Rogers, Jordan Boyd-Graber, and Naoaki Okazaki, editors, \emph{Findings of the Association for Computational Linguistics: ACL 2023}, pages 431--469, Toronto, Canada, July 2023{\natexlab{a}}. Association for Computational Linguistics.
\newblock \doi{10.18653/v1/2023.findings-acl.29}.
\newblock URL \url{https://aclanthology.org/2023.findings-acl.29/}.

\bibitem[Laskar et~al.(2023{\natexlab{b}})Laskar, Rahman, Jahan, Hoque, and Huang]{laskar2023cqsumdp}
Md~Tahmid~Rahman Laskar, Mizanur Rahman, Israt Jahan, Enamul Hoque, and Jimmy Huang.
\newblock Cqsumdp: a chatgpt-annotated resource for query-focused abstractive summarization based on debatepedia.
\newblock \emph{arXiv preprint arXiv:2305.06147}, 2023{\natexlab{b}}.

\bibitem[Laver et~al.(2003)Laver, Benoit, and Garry]{laver2003extracting}
Michael Laver, Kenneth Benoit, and John Garry.
\newblock Extracting policy positions from political texts using words as data.
\newblock \emph{American political science review}, 97\penalty0 (2):\penalty0 311--331, 2003.
\newblock \doi{10.1017/S0003055403000698}.

\bibitem[Lehmann et~al.(2024)Lehmann, Franzmann, Al-Gaddooa, Burst, Ivanusch, Regel, Riethmüller, Volkens, Weßels, and Zehnter]{Lehmann2024}
Pola Lehmann, Simon Franzmann, Denise Al-Gaddooa, Tobias Burst, Christoph Ivanusch, Sven Regel, Felicia Riethmüller, Andrea Volkens, Bernhard Weßels, and Lisa Zehnter.
\newblock The manifesto data collection. manifesto project (mrg/cmp/marpor). version 2024a, 2024.
\newblock URL \url{https://doi.org/10.25522/manifesto.mpds.2024a}.

\bibitem[Li(2024)]{li-2024-human}
Jiyi Li.
\newblock Human-{LLM} hybrid text answer aggregation for crowd annotations.
\newblock In Yaser Al-Onaizan, Mohit Bansal, and Yun-Nung Chen, editors, \emph{Proceedings of the 2024 Conference on Empirical Methods in Natural Language Processing}, pages 15609--15622, Miami, Florida, USA, November 2024. Association for Computational Linguistics.
\newblock \doi{10.18653/v1/2024.emnlp-main.874}.
\newblock URL \url{https://aclanthology.org/2024.emnlp-main.874/}.

\bibitem[Li et~al.(2024{\natexlab{a}})Li, Li, Chen, Gui, Yang, Yu, Wang, Cai, Zhou, Shen, et~al.]{li2024political}
Lincan Li, Jiaqi Li, Catherine Chen, Fred Gui, Hongjia Yang, Chenxiao Yu, Zhengguang Wang, Jianing Cai, Junlong~Aaron Zhou, Bolin Shen, et~al.
\newblock Political-llm: Large language models in political science.
\newblock \emph{arXiv preprint arXiv:2412.06864}, 2024{\natexlab{a}}.

\bibitem[Li et~al.(2024{\natexlab{b}})Li, Fan, Atreja, and Hemphill]{Li2024HOT}
Lingyao Li, Lizhou Fan, Shubham Atreja, and Libby Hemphill.
\newblock “hot” chatgpt: The promise of chatgpt in detecting and discriminating hateful, offensive, and toxic comments on social media.
\newblock \emph{ACM Trans. Web}, 18\penalty0 (2), March 2024{\natexlab{b}}.
\newblock ISSN 1559-1131.
\newblock \doi{10.1145/3643829}.
\newblock URL \url{https://doi.org/10.1145/3643829}.

\bibitem[Li et~al.(2023)Li, Shi, Ziems, Kan, Chen, Liu, and Yang]{li-etal-2023-coannotating}
Minzhi Li, Taiwei Shi, Caleb Ziems, Min-Yen Kan, Nancy Chen, Zhengyuan Liu, and Diyi Yang.
\newblock {C}o{A}nnotating: Uncertainty-guided work allocation between human and large language models for data annotation.
\newblock In Houda Bouamor, Juan Pino, and Kalika Bali, editors, \emph{Proceedings of the 2023 Conference on Empirical Methods in Natural Language Processing}, pages 1487--1505, Singapore, December 2023. Association for Computational Linguistics.
\newblock \doi{10.18653/v1/2023.emnlp-main.92}.
\newblock URL \url{https://aclanthology.org/2023.emnlp-main.92/}.

\bibitem[Liang et~al.(2023)Liang, Bommasani, Lee, Tsipras, Soylu, Yasunaga, Zhang, Narayanan, Wu, Kumar, Newman, Yuan, Yan, Zhang, Cosgrove, Manning, Re, Acosta-Navas, Hudson, Zelikman, Durmus, Ladhak, Rong, Ren, Yao, WANG, Santhanam, Orr, Zheng, Yuksekgonul, Suzgun, Kim, Guha, Chatterji, Khattab, Henderson, Huang, Chi, Xie, Santurkar, Ganguli, Hashimoto, Icard, Zhang, Chaudhary, Wang, Li, Mai, Zhang, and Koreeda]{liang2023holistic}
Percy Liang, Rishi Bommasani, Tony Lee, Dimitris Tsipras, Dilara Soylu, Michihiro Yasunaga, Yian Zhang, Deepak Narayanan, Yuhuai Wu, Ananya Kumar, Benjamin Newman, Binhang Yuan, Bobby Yan, Ce~Zhang, Christian Cosgrove, Christopher~D Manning, Christopher Re, Diana Acosta-Navas, Drew~A. Hudson, Eric Zelikman, Esin Durmus, Faisal Ladhak, Frieda Rong, Hongyu Ren, Huaxiu Yao, Jue WANG, Keshav Santhanam, Laurel Orr, Lucia Zheng, Mert Yuksekgonul, Mirac Suzgun, Nathan Kim, Neel Guha, Niladri~S. Chatterji, Omar Khattab, Peter Henderson, Qian Huang, Ryan~Andrew Chi, Sang~Michael Xie, Shibani Santurkar, Surya Ganguli, Tatsunori Hashimoto, Thomas Icard, Tianyi Zhang, Vishrav Chaudhary, William Wang, Xuechen Li, Yifan Mai, Yuhui Zhang, and Yuta Koreeda.
\newblock Holistic evaluation of language models.
\newblock \emph{Transactions on Machine Learning Research}, 2023.
\newblock ISSN 2835-8856.
\newblock URL \url{https://openreview.net/forum?id=iO4LZibEqW}.
\newblock Featured Certification, Expert Certification, Outstanding Certification.

\bibitem[Liao et~al.(2025)Liao, Antoniak, Cheong, Cheng, Lee, Lo, Chang, and Zhang]{liao2025llms}
Zhehui Liao, Maria Antoniak, Inyoung Cheong, Evie Yu-Yen Cheng, Ai-Heng Lee, Kyle Lo, Joseph~Chee Chang, and Amy~X Zhang.
\newblock {LLM}s as research tools: A large scale survey of researchers{\textquoteright} usage and perceptions.
\newblock In \emph{Second Conference on Language Modeling}, 2025.
\newblock URL \url{https://openreview.net/forum?id=p0BwJk3R1p}.

\bibitem[Lin et~al.(2022)Lin, Hilton, and Evans]{lin2022teaching}
Stephanie Lin, Jacob Hilton, and Owain Evans.
\newblock Teaching models to express their uncertainty in words.
\newblock \emph{Transactions on Machine Learning Research}, 2022.
\newblock ISSN 2835-8856.
\newblock URL \url{https://openreview.net/forum?id=8s8K2UZGTZ}.

\bibitem[Lindeman et~al.(1980)Lindeman, Merenda, Gold, et~al.]{lindeman1980introduction}
Richard~Harold Lindeman, Peter~Francis Merenda, Ruth~Z Gold, et~al.
\newblock \emph{Introduction to bivariate and multivariate analysis}, volume~4.
\newblock Scott, Foresman Glenview, IL, 1980.

\bibitem[Linegar et~al.(2023)Linegar, Kocielnik, and Alvarez]{Linegar2023}
Mitchell Linegar, Rafal Kocielnik, and R.~Michael Alvarez.
\newblock Large language models and political science.
\newblock \emph{Frontiers in Political Science}, Volume 5 - 2023, 2023.
\newblock ISSN 2673-3145.
\newblock \doi{10.3389/fpos.2023.1257092}.
\newblock URL \url{https://www.frontiersin.org/journals/political-science/articles/10.3389/fpos.2023.1257092}.

\bibitem[Liu et~al.(2023)Liu, Yao, Ton, Zhang, Guo, Cheng, Klochkov, Taufiq, and Li]{liu2023trustworthy}
Yang Liu, Yuanshun Yao, Jean-Francois Ton, Xiaoying Zhang, Ruocheng Guo, Hao Cheng, Yegor Klochkov, Muhammad~Faaiz Taufiq, and Hang Li.
\newblock Trustworthy llms: a survey and guideline for evaluating large language models' alignment.
\newblock \emph{arXiv preprint arXiv:2308.05374}, 2023.

\bibitem[Liyanage et~al.(2024)Liyanage, Gokani, and Mago]{Liyanage2024}
Chandreen~R. Liyanage, Ravi Gokani, and Vijay Mago.
\newblock Gpt-4 as an x data annotator: Unraveling its performance on a stance classification task.
\newblock \emph{PLOS ONE}, 19\penalty0 (8):\penalty0 1--21, 08 2024.
\newblock \doi{10.1371/journal.pone.0307741}.
\newblock URL \url{https://doi.org/10.1371/journal.pone.0307741}.

\bibitem[{Llama Team, AI @ Meta}(2024)]{grattafiori2024llama}
{Llama Team, AI @ Meta}.
\newblock The llama 3 herd of models.
\newblock \emph{arXiv preprint arXiv:2407.21783}, 2024.

\bibitem[Lombard et~al.(2002)Lombard, Snyder-Duch, and Bracken]{Lombard2002}
Matthew Lombard, Jennifer Snyder-Duch, and Cheryl~Campanella Bracken.
\newblock Content analysis in mass communication: Assessment and reporting of intercoder reliability.
\newblock \emph{Human Communication Research}, 28\penalty0 (4):\penalty0 587--604, 2002.
\newblock \doi{https://doi.org/10.1111/j.1468-2958.2002.tb00826.x}.
\newblock URL \url{https://onlinelibrary.wiley.com/doi/abs/10.1111/j.1468-2958.2002.tb00826.x}.

\bibitem[Lu et~al.(2024)Lu, Hu, and Chen]{Lu_2024}
Tianxuan Lu, Jin Hu, and Pingping Chen.
\newblock Benchmarking llama 3 for chinese news summation: Accuracy, cultural nuance, and societal value alignment.
\newblock June 2024.
\newblock \doi{10.36227/techrxiv.171742386.68305769/v1}.
\newblock URL \url{http://dx.doi.org/10.36227/techrxiv.171742386.68305769/v1}.

\bibitem[Ludwig et~al.(2025)Ludwig, Mullainathan, and Rambachan]{Ludwig2025EconometricFramework}
Jens Ludwig, Sendhil Mullainathan, and Ashesh Rambachan.
\newblock Large language models: An applied econometric framework.
\newblock Working Paper 33344, National Bureau of Economic Research, January 2025.
\newblock URL \url{http://www.nber.org/papers/w33344}.

\bibitem[Luo et~al.(2020)Luo, Card, and Jurafsky]{luo-etal-2020-detecting}
Yiwei Luo, Dallas Card, and Dan Jurafsky.
\newblock Detecting stance in media on global warming.
\newblock In Trevor Cohn, Yulan He, and Yang Liu, editors, \emph{Findings of the Association for Computational Linguistics: EMNLP 2020}, pages 3296--3315, Online, November 2020. Association for Computational Linguistics.
\newblock \doi{10.18653/v1/2020.findings-emnlp.296}.
\newblock URL \url{https://aclanthology.org/2020.findings-emnlp.296/}.

\bibitem[Lupo et~al.(2023)Lupo, Magnusson, Hovy, Naurin, and W{\"a}ngnerud]{lupo2023towards}
Lorenzo Lupo, Oscar Magnusson, Dirk Hovy, Elin Naurin, and Lena W{\"a}ngnerud.
\newblock Towards human-level text coding with llms: The case of fatherhood roles in public policy documents.
\newblock \emph{arXiv preprint arXiv:2311.11844}, 2023.

\bibitem[M{\ae}hlum et~al.(2024)M{\ae}hlum, Samuel, Norman, Jelin, Bjertn{\ae}s, {\O}vrelid, and Velldal]{maehlum-etal-2024-difficult}
Petter M{\ae}hlum, David Samuel, Rebecka~Maria Norman, Elma Jelin, {\O}yvind~Andresen Bjertn{\ae}s, Lilja {\O}vrelid, and Erik Velldal.
\newblock It{'}s difficult to be neutral {--} human and {LLM}-based sentiment annotation of patient comments.
\newblock In Dina Demner-Fushman, Sophia Ananiadou, Paul Thompson, and Brian Ondov, editors, \emph{Proceedings of the First Workshop on Patient-Oriented Language Processing (CL4Health) @ LREC-COLING 2024}, pages 8--19, Torino, Italia, May 2024. ELRA and ICCL.
\newblock URL \url{https://aclanthology.org/2024.cl4health-1.2/}.

\bibitem[Margatina et~al.(2021)Margatina, Vernikos, Barrault, and Aletras]{margatina-etal-2021-active}
Katerina Margatina, Giorgos Vernikos, Lo{\"i}c Barrault, and Nikolaos Aletras.
\newblock Active learning by acquiring contrastive examples.
\newblock In Marie-Francine Moens, Xuanjing Huang, Lucia Specia, and Scott Wen-tau Yih, editors, \emph{Proceedings of the 2021 Conference on Empirical Methods in Natural Language Processing}, pages 650--663, Online and Punta Cana, Dominican Republic, November 2021. Association for Computational Linguistics.
\newblock \doi{10.18653/v1/2021.emnlp-main.51}.
\newblock URL \url{https://aclanthology.org/2021.emnlp-main.51/}.

\bibitem[Meehl(1967)]{Meehl_1967}
Paul~E. Meehl.
\newblock Theory-testing in psychology and physics: A methodological paradox.
\newblock \emph{Philosophy of Science}, 34\penalty0 (2):\penalty0 103–115, 1967.
\newblock \doi{10.1086/288135}.

\bibitem[Mellon et~al.(2024)Mellon, Bailey, Scott, Breckwoldt, Miori, and Schmedeman]{Mellon2024Issue}
Jonathan Mellon, Jack Bailey, Ralph Scott, James Breckwoldt, Marta Miori, and Phillip Schmedeman.
\newblock Do ais know what the most important issue is? using language models to code open-text social survey responses at scale.
\newblock \emph{Research \& Politics}, 11\penalty0 (1):\penalty0 20531680241231468, 2024.
\newblock \doi{10.1177/20531680241231468}.
\newblock URL \url{https://doi.org/10.1177/20531680241231468}.

\bibitem[Meng(2018)]{Meng2018paradises}
Xiao-Li Meng.
\newblock {Statistical paradises and paradoxes in big data (I): Law of large populations, big data paradox, and the 2016 US presidential election}.
\newblock \emph{The Annals of Applied Statistics}, 12\penalty0 (2):\penalty0 685 -- 726, 2018.
\newblock \doi{10.1214/18-AOAS1161SF}.
\newblock URL \url{https://doi.org/10.1214/18-AOAS1161SF}.

\bibitem[Mens et~al.(2023)Mens, Kovács, Hannan, and Pros]{LeMenspnas2309350120}
Gaël~Le Mens, Balázs Kovács, Michael~T. Hannan, and Guillem Pros.
\newblock Uncovering the semantics of concepts using gpt-4.
\newblock \emph{Proceedings of the National Academy of Sciences}, 120\penalty0 (49):\penalty0 e2309350120, 2023.
\newblock \doi{10.1073/pnas.2309350120}.
\newblock URL \url{https://www.pnas.org/doi/abs/10.1073/pnas.2309350120}.

\bibitem[Merz et~al.(2016)Merz, Regel, and Lewandowski]{merz2016manifesto}
Nicolas Merz, Sven Regel, and Jirka Lewandowski.
\newblock The manifesto corpus: A new resource for research on political parties and quantitative text analysis.
\newblock \emph{Research \& Politics}, 3\penalty0 (2):\penalty0 2053168016643346, 2016.

\bibitem[Mets et~al.(2024)Mets, Karjus, Ibrus, and Schich]{Mets2024}
Mark Mets, Andres Karjus, Indrek Ibrus, and Maximilian Schich.
\newblock Automated stance detection in complex topics and small languages: The challenging case of immigration in polarizing news media.
\newblock \emph{PLOS ONE}, 19\penalty0 (4):\penalty0 1--16, 04 2024.
\newblock \doi{10.1371/journal.pone.0302380}.
\newblock URL \url{https://doi.org/10.1371/journal.pone.0302380}.

\bibitem[Min et~al.(2023)Min, Krishna, Lyu, Lewis, Yih, Koh, Iyyer, Zettlemoyer, and Hajishirzi]{min-etal-2023-factscore}
Sewon Min, Kalpesh Krishna, Xinxi Lyu, Mike Lewis, Wen-tau Yih, Pang Koh, Mohit Iyyer, Luke Zettlemoyer, and Hannaneh Hajishirzi.
\newblock {FA}ct{S}core: Fine-grained atomic evaluation of factual precision in long form text generation.
\newblock In Houda Bouamor, Juan Pino, and Kalika Bali, editors, \emph{Proceedings of the 2023 Conference on Empirical Methods in Natural Language Processing}, pages 12076--12100, Singapore, December 2023. Association for Computational Linguistics.
\newblock \doi{10.18653/v1/2023.emnlp-main.741}.
\newblock URL \url{https://aclanthology.org/2023.emnlp-main.741/}.

\bibitem[Mishra and Chatterjee(2024)]{Mishra2024}
Shyamal Mishra and Preetha Chatterjee.
\newblock Exploring chatgpt for toxicity detection in github.
\newblock In \emph{Proceedings of the 2024 ACM/IEEE 44th International Conference on Software Engineering: New Ideas and Emerging Results}, ICSE-NIER'24, page 6–10, New York, NY, USA, 2024. Association for Computing Machinery.
\newblock ISBN 9798400705007.
\newblock \doi{10.1145/3639476.3639777}.
\newblock URL \url{https://doi.org/10.1145/3639476.3639777}.

\bibitem[Misiejuk et~al.(2024)Misiejuk, Kaliisa, and Scianna]{MISIEJUK2024100216}
Kamila Misiejuk, Rogers Kaliisa, and Jennifer Scianna.
\newblock Augmenting assessment with ai coding of online student discourse: A question of reliability.
\newblock \emph{Computers and Education: Artificial Intelligence}, 6:\penalty0 100216, 2024.
\newblock ISSN 2666-920X.
\newblock \doi{https://doi.org/10.1016/j.caeai.2024.100216}.
\newblock URL \url{https://www.sciencedirect.com/science/article/pii/S2666920X24000171}.

\bibitem[Mizumoto and Eguchi(2023)]{Mizumoto2023}
Atsushi Mizumoto and Masaki Eguchi.
\newblock Exploring the potential of using an ai language model for automated essay scoring.
\newblock \emph{Research Methods in Applied Linguistics}, 2\penalty0 (2):\penalty0 100050, 2023.
\newblock ISSN 2772-7661.
\newblock \doi{https://doi.org/10.1016/j.rmal.2023.100050}.
\newblock URL \url{https://www.sciencedirect.com/science/article/pii/S2772766123000101}.

\bibitem[Modarressi et~al.(2025)Modarressi, Spiess, and Venugopal]{modarressi2025causal}
Iman Modarressi, Jann Spiess, and Amar Venugopal.
\newblock Causal inference on outcomes learned from text.
\newblock \emph{arXiv preprint arXiv:2503.00725}, 2025.

\bibitem[Moghimifar et~al.(2024)Moghimifar, Li, Thomson, and Haffari]{moghimifar2024modelling}
Farhad Moghimifar, Yuan-Fang Li, Robert Thomson, and Gholamreza Haffari.
\newblock Modelling political coalition negotiations using llm-based agents.
\newblock \emph{arXiv preprint arXiv:2402.11712}, 2024.

\bibitem[M{\o}ller et~al.(2024)M{\o}ller, Pera, Dalsgaard, and Aiello]{moller-etal-2024-parrot}
Anders~Giovanni M{\o}ller, Arianna Pera, Jacob Dalsgaard, and Luca Aiello.
\newblock The parrot dilemma: Human-labeled vs. {LLM}-augmented data in classification tasks.
\newblock In Yvette Graham and Matthew Purver, editors, \emph{Proceedings of the 18th Conference of the European Chapter of the Association for Computational Linguistics (Volume 2: Short Papers)}, pages 179--192, St. Julian{'}s, Malta, March 2024. Association for Computational Linguistics.
\newblock \doi{10.18653/v1/2024.eacl-short.17}.
\newblock URL \url{https://aclanthology.org/2024.eacl-short.17/}.

\bibitem[Moretti(2013)]{moretti2013distant}
Franco Moretti.
\newblock \emph{Distant reading}.
\newblock Verso Books, 2013.

\bibitem[Movva et~al.(2024)Movva, Koh, and Pierson]{movva-etal-2024-annotation}
Rajiv Movva, Pang~Wei Koh, and Emma Pierson.
\newblock Annotation alignment: Comparing {LLM} and human annotations of conversational safety.
\newblock In Yaser Al-Onaizan, Mohit Bansal, and Yun-Nung Chen, editors, \emph{Proceedings of the 2024 Conference on Empirical Methods in Natural Language Processing}, pages 9048--9062, Miami, Florida, USA, November 2024. Association for Computational Linguistics.
\newblock \doi{10.18653/v1/2024.emnlp-main.511}.
\newblock URL \url{https://aclanthology.org/2024.emnlp-main.511/}.

\bibitem[Mu et~al.(2024)Mu, Wu, Thorne, Robinson, Aletras, Scarton, Bontcheva, and Song]{mu-etal-2024-navigating}
Yida Mu, Ben~P. Wu, William Thorne, Ambrose Robinson, Nikolaos Aletras, Carolina Scarton, Kalina Bontcheva, and Xingyi Song.
\newblock Navigating prompt complexity for zero-shot classification: A study of large language models in computational social science.
\newblock In Nicoletta Calzolari, Min-Yen Kan, Veronique Hoste, Alessandro Lenci, Sakriani Sakti, and Nianwen Xue, editors, \emph{Proceedings of the 2024 Joint International Conference on Computational Linguistics, Language Resources and Evaluation (LREC-COLING 2024)}, pages 12074--12086, Torino, Italia, May 2024. ELRA and ICCL.
\newblock URL \url{https://aclanthology.org/2024.lrec-main.1055/}.

\bibitem[Ni et~al.(2024)Ni, Shi, Stammbach, Sachan, Ash, and Leippold]{ni-etal-2024-afacta}
Jingwei Ni, Minjing Shi, Dominik Stammbach, Mrinmaya Sachan, Elliott Ash, and Markus Leippold.
\newblock {AF}a{CTA}: Assisting the annotation of factual claim detection with reliable {LLM} annotators.
\newblock In Lun-Wei Ku, Andre Martins, and Vivek Srikumar, editors, \emph{Proceedings of the 62nd Annual Meeting of the Association for Computational Linguistics (Volume 1: Long Papers)}, pages 1890--1912, Bangkok, Thailand, August 2024. Association for Computational Linguistics.
\newblock \doi{10.18653/v1/2024.acl-long.104}.
\newblock URL \url{https://aclanthology.org/2024.acl-long.104/}.

\bibitem[Ollion et~al.(2024)Ollion, Shen, Macanovic, and Chatelain]{ollion2024dangers}
{\'E}tienne Ollion, Rubing Shen, Ana Macanovic, and Arnault Chatelain.
\newblock The dangers of using proprietary llms for research.
\newblock \emph{Nature Machine Intelligence}, 6\penalty0 (1):\penalty0 4--5, 2024.

\bibitem[OpenAI(2024)]{openai2024gpt4ocard}
OpenAI.
\newblock Gpt-4o system card.
\newblock \emph{arXiv preprint arXiv:2410.21276}, 2024.

\bibitem[Ornstein et~al.(2025)Ornstein, Blasingame, and Truscott]{OrnsteinBlasingameTruscott2025}
Joseph~T. Ornstein, Elise~N. Blasingame, and Jake~S. Truscott.
\newblock How to train your stochastic parrot: large language models for political texts.
\newblock \emph{Political Science Research and Methods}, 13\penalty0 (2):\penalty0 264–281, 2025.
\newblock \doi{10.1017/psrm.2024.64}.

\bibitem[Ouyang et~al.(2022{\natexlab{a}})Ouyang, Wu, Jiang, Almeida, Wainwright, Mishkin, Zhang, Agarwal, Slama, Ray, Schulman, Hilton, Kelton, Miller, Simens, Askell, Welinder, Christiano, Leike, and Lowe]{NEURIPS2022_b1efde53}
Long Ouyang, Jeffrey Wu, Xu~Jiang, Diogo Almeida, Carroll Wainwright, Pamela Mishkin, Chong Zhang, Sandhini Agarwal, Katarina Slama, Alex Ray, John Schulman, Jacob Hilton, Fraser Kelton, Luke Miller, Maddie Simens, Amanda Askell, Peter Welinder, Paul~F Christiano, Jan Leike, and Ryan Lowe.
\newblock Training language models to follow instructions with human feedback.
\newblock In \emph{Advances in Neural Information Processing Systems}, volume~35, pages 27730--27744, 2022{\natexlab{a}}.
\newblock URL \url{https://proceedings.neurips.cc/paper_files/paper/2022/file/b1efde53be364a73914f58805a001731-Paper-Conference.pdf}.

\bibitem[Ouyang et~al.(2022{\natexlab{b}})Ouyang, Wu, Jiang, Almeida, Wainwright, Mishkin, Zhang, Agarwal, Slama, Ray, Schulman, Hilton, Kelton, Miller, Simens, Askell, Welinder, Christiano, Leike, and Lowe]{NEURIPS2022b1efde53}
Long Ouyang, Jeffrey Wu, Xu~Jiang, Diogo Almeida, Carroll Wainwright, Pamela Mishkin, Chong Zhang, Sandhini Agarwal, Katarina Slama, Alex Ray, John Schulman, Jacob Hilton, Fraser Kelton, Luke Miller, Maddie Simens, Amanda Askell, Peter Welinder, Paul~F Christiano, Jan Leike, and Ryan Lowe.
\newblock Training language models to follow instructions with human feedback.
\newblock In S.~Koyejo, S.~Mohamed, A.~Agarwal, D.~Belgrave, K.~Cho, and A.~Oh, editors, \emph{Advances in Neural Information Processing Systems}, volume~35, pages 27730--27744. Curran Associates, Inc., 2022{\natexlab{b}}.
\newblock URL \url{https://proceedings.neurips.cc/paper_files/paper/2022/file/b1efde53be364a73914f58805a001731-Paper-Conference.pdf}.

\bibitem[Pangakis et~al.(2023)Pangakis, Wolken, and Fasching]{pangakis2023automated}
Nicholas Pangakis, Samuel Wolken, and Neil Fasching.
\newblock Automated annotation with generative ai requires validation.
\newblock \emph{arXiv preprint arXiv:2306.00176}, 2023.

\bibitem[Pangakis and Wolken(2025)]{pangakis2025keeping}
Nick Pangakis and Sam Wolken.
\newblock Keeping humans in the loop: Human-centered automated annotation with generative ai.
\newblock In \emph{International AAAI Conference on Web and Social Media}, June 2025.
\newblock URL \url{https://www.microsoft.com/en-us/research/publication/keeping-humans-in-the-loop-human-centered-automated-annotation-with-generative-ai/}.

\bibitem[Pavlovic and Poesio(2024)]{pavlovic-poesio-2024-effectiveness}
Maja Pavlovic and Massimo Poesio.
\newblock The effectiveness of {LLM}s as annotators: A comparative overview and empirical analysis of direct representation.
\newblock In Gavin Abercrombie, Valerio Basile, Davide Bernadi, Shiran Dudy, Simona Frenda, Lucy Havens, and Sara Tonelli, editors, \emph{Proceedings of the 3rd Workshop on Perspectivist Approaches to NLP (NLPerspectives) @ LREC-COLING 2024}, pages 100--110, Torino, Italia, May 2024. ELRA and ICCL.
\newblock URL \url{https://aclanthology.org/2024.nlperspectives-1.11/}.

\bibitem[Peskine et~al.(2023)Peskine, Koren{\v{c}}i{\'c}, Grubisic, Papotti, Troncy, and Rosso]{peskine-etal-2023-definitions}
Youri Peskine, Damir Koren{\v{c}}i{\'c}, Ivan Grubisic, Paolo Papotti, Raphael Troncy, and Paolo Rosso.
\newblock Definitions matter: Guiding {GPT} for multi-label classification.
\newblock In Houda Bouamor, Juan Pino, and Kalika Bali, editors, \emph{Findings of the Association for Computational Linguistics: EMNLP 2023}, pages 4054--4063, Singapore, December 2023. Association for Computational Linguistics.
\newblock \doi{10.18653/v1/2023.findings-emnlp.267}.
\newblock URL \url{https://aclanthology.org/2023.findings-emnlp.267/}.

\bibitem[Qiu and Lan(2024)]{qiu2024interactive}
Huachuan Qiu and Zhenzhong Lan.
\newblock Interactive agents: Simulating counselor-client psychological counseling via role-playing llm-to-llm interactions.
\newblock \emph{arXiv preprint arXiv:2408.15787}, 2024.

\bibitem[{Qwen Team}(2025{\natexlab{a}})]{qwen2025qwen25technicalreport}
{Qwen Team}.
\newblock Qwen2.5 technical report.
\newblock \emph{arXiv preprint arXiv:2412.15115}, 2025{\natexlab{a}}.

\bibitem[{Qwen Team}(2025{\natexlab{b}})]{qwen2025qwen3technicalreport}
{Qwen Team}.
\newblock Qwen3 technical report.
\newblock \emph{arXiv preprint arXiv:2505.09388}, 2025{\natexlab{b}}.

\bibitem[Rathje et~al.(2024)Rathje, Mirea, Sucholutsky, Marjieh, Robertson, and Bavel]{Rathje2024pnas}
Steve Rathje, Dan-Mircea Mirea, Ilia Sucholutsky, Raja Marjieh, Claire~E. Robertson, and Jay J.~Van Bavel.
\newblock Gpt is an effective tool for multilingual psychological text analysis.
\newblock \emph{Proceedings of the National Academy of Sciences}, 121\penalty0 (34):\penalty0 e2308950121, 2024.
\newblock \doi{10.1073/pnas.2308950121}.
\newblock URL \url{https://www.pnas.org/doi/abs/10.1073/pnas.2308950121}.

\bibitem[Rogers and Zhang(2024)]{Rogers2024}
Richard Rogers and Xiaoke Zhang.
\newblock The russia–ukraine war in chinese social media: Llm analysis yields a bias toward neutrality.
\newblock \emph{Social Media + Society}, 10\penalty0 (2):\penalty0 20563051241254379, 2024.
\newblock \doi{10.1177/20563051241254379}.
\newblock URL \url{https://doi.org/10.1177/20563051241254379}.

\bibitem[Ro{\ss} et~al.(2016)Ro{\ss}, Rist, Carbonell, Cabrera, Kurowsky, Wojatzki, and {NLP4CMC III: 3rdWorkshop on Natural Language Processing for Computer-Mediated Communication 22 September 2016}]{duepublico_mods_00042132}
Bj{\"o}rn Ro{\ss}, Michael Rist, Guillermo Carbonell, Benjamin Cabrera, Nils Kurowsky, Michael Wojatzki, and {NLP4CMC III: 3rdWorkshop on Natural Language Processing for Computer-Mediated Communication 22 September 2016}.
\newblock Measuring the reliability of hate speech annotations: the case of the european refugee crisis.
\newblock Sep 2016.
\newblock \doi{10.17185/duepublico/42132}.
\newblock URL \url{https://doi.org/10.17185/duepublico/42132}.

\bibitem[Rothstein et~al.(2005)Rothstein, Sutton, and Borenstein]{rothstein2005publication}
Hannah~R Rothstein, Alexander~J Sutton, and Michael Borenstein.
\newblock Publication bias in meta-analysis.
\newblock \emph{Publication bias in meta-analysis: Prevention, assessment and adjustments}, pages 1--7, 2005.

\bibitem[Rouzegar and Makrehchi(2024)]{rouzegar-makrehchi-2024-enhancing}
Hamidreza Rouzegar and Masoud Makrehchi.
\newblock Enhancing text classification through {LLM}-driven active learning and human annotation.
\newblock In Sophie Henning and Manfred Stede, editors, \emph{Proceedings of the 18th Linguistic Annotation Workshop (LAW-XVIII)}, pages 98--111, St. Julians, Malta, March 2024. Association for Computational Linguistics.
\newblock URL \url{https://aclanthology.org/2024.law-1.10/}.

\bibitem[Roy et~al.(2023)Roy, Harshvardhan, Mukherjee, and Saha]{roy-etal-2023-probing}
Sarthak Roy, Ashish Harshvardhan, Animesh Mukherjee, and Punyajoy Saha.
\newblock Probing {LLM}s for hate speech detection: strengths and vulnerabilities.
\newblock In Houda Bouamor, Juan Pino, and Kalika Bali, editors, \emph{Findings of the Association for Computational Linguistics: EMNLP 2023}, pages 6116--6128, Singapore, December 2023. Association for Computational Linguistics.
\newblock \doi{10.18653/v1/2023.findings-emnlp.407}.
\newblock URL \url{https://aclanthology.org/2023.findings-emnlp.407/}.

\bibitem[Roy et~al.(2022)Roy, Nakshatri, and Goldwasser]{roy-etal-2022-towards}
Shamik Roy, Nishanth~Sridhar Nakshatri, and Dan Goldwasser.
\newblock Towards few-shot identification of morality frames using in-context learning.
\newblock In David Bamman, Dirk Hovy, David Jurgens, Katherine Keith, Brendan O'Connor, and Svitlana Volkova, editors, \emph{Proceedings of the Fifth Workshop on Natural Language Processing and Computational Social Science (NLP+CSS)}, pages 183--196, Abu Dhabi, UAE, November 2022. Association for Computational Linguistics.
\newblock \doi{10.18653/v1/2022.nlpcss-1.20}.
\newblock URL \url{https://aclanthology.org/2022.nlpcss-1.20/}.

\bibitem[Salinas and Morstatter(2024)]{salinas-morstatter-2024-butterfly}
Abel Salinas and Fred Morstatter.
\newblock The butterfly effect of altering prompts: How small changes and jailbreaks affect large language model performance.
\newblock In Lun-Wei Ku, Andre Martins, and Vivek Srikumar, editors, \emph{Findings of the Association for Computational Linguistics: ACL 2024}, pages 4629--4651, Bangkok, Thailand, August 2024. Association for Computational Linguistics.
\newblock \doi{10.18653/v1/2024.findings-acl.275}.
\newblock URL \url{https://aclanthology.org/2024.findings-acl.275/}.

\bibitem[Sclar et~al.(2024)Sclar, Choi, Tsvetkov, and Suhr]{sclar2024quantifying}
Melanie Sclar, Yejin Choi, Yulia Tsvetkov, and Alane Suhr.
\newblock Quantifying language models' sensitivity to spurious features in prompt design or: How i learned to start worrying about prompt formatting.
\newblock In \emph{The Twelfth International Conference on Learning Representations}, 2024.
\newblock URL \url{https://openreview.net/forum?id=RIu5lyNXjT}.

\bibitem[Shu et~al.(2020)Shu, Mahudeswaran, Wang, Lee, and Liu]{shu2020fakenewsnet}
Kai Shu, Deepak Mahudeswaran, Suhang Wang, Dongwon Lee, and Huan Liu.
\newblock Fakenewsnet: A data repository with news content, social context, and spatiotemporal information for studying fake news on social media.
\newblock \emph{Big Data}, 8\penalty0 (3):\penalty0 171--188, 2020.
\newblock \doi{10.1089/big.2020.0062}.
\newblock URL \url{https://doi.org/10.1089/big.2020.0062}.
\newblock PMID: 32491943.

\bibitem[Simmons et~al.(2011)Simmons, Nelson, and Simonsohn]{Simmons2011}
Joseph~P. Simmons, Leif~D. Nelson, and Uri Simonsohn.
\newblock False-positive psychology: Undisclosed flexibility in data collection and analysis allows presenting anything as significant.
\newblock \emph{Psychological Science}, 22\penalty0 (11):\penalty0 1359--1366, 2011.
\newblock \doi{10.1177/0956797611417632}.
\newblock URL \url{https://doi.org/10.1177/0956797611417632}.
\newblock PMID: 22006061.

\bibitem[Simonsohn et~al.(2014)Simonsohn, Nelson, and Simmons]{simonsohn2014p}
Uri Simonsohn, Leif~D Nelson, and Joseph~P Simmons.
\newblock P-curve: a key to the file-drawer.
\newblock \emph{Journal of experimental psychology: General}, 143\penalty0 (2):\penalty0 534, 2014.
\newblock \doi{10.1037/a0033242}.

\bibitem[Steegen et~al.(2016)Steegen, Tuerlinckx, Gelman, and Vanpaemel]{Steegen2016Multiverse}
Sara Steegen, Francis Tuerlinckx, Andrew Gelman, and Wolf Vanpaemel.
\newblock Increasing transparency through a multiverse analysis.
\newblock \emph{Perspectives on Psychological Science}, 11\penalty0 (5):\penalty0 702--712, 2016.
\newblock \doi{10.1177/1745691616658637}.
\newblock URL \url{https://doi.org/10.1177/1745691616658637}.
\newblock PMID: 27694465.

\bibitem[Stefan and Schönbrodt(2023)]{Stefan2023}
Angelika~M. Stefan and Felix~D. Schönbrodt.
\newblock Big little lies: a compendium and simulation of <i>p</i>-hacking strategies.
\newblock \emph{Royal Society Open Science}, 10\penalty0 (2):\penalty0 220346, 2023.
\newblock \doi{10.1098/rsos.220346}.
\newblock URL \url{https://royalsocietypublishing.org/doi/abs/10.1098/rsos.220346}.

\bibitem[Sterling(1959)]{sterling1959publication}
Theodore~D Sterling.
\newblock Publication decisions and their possible effects on inferences drawn from tests of significance—or vice versa.
\newblock \emph{Journal of the American statistical association}, 54\penalty0 (285):\penalty0 30--34, 1959.

\bibitem[Stuhler et~al.(2025)Stuhler, Ton, and Ollion]{Stuhler2025}
Oscar Stuhler, Cat~Dang Ton, and Etienne Ollion.
\newblock From codebooks to promptbooks: Extracting information from text with generative large language models.
\newblock \emph{Sociological Methods \& Research}, 54\penalty0 (3):\penalty0 794--848, 2025.
\newblock \doi{10.1177/00491241251336794}.
\newblock URL \url{https://doi.org/10.1177/00491241251336794}.

\bibitem[Thirunavukarasu et~al.(2023)Thirunavukarasu, Ting, Elangovan, Gutierrez, Tan, and Ting]{thirunavukarasu2023large}
Arun~James Thirunavukarasu, Darren Shu~Jeng Ting, Kabilan Elangovan, Laura Gutierrez, Ting~Fang Tan, and Daniel Shu~Wei Ting.
\newblock Large language models in medicine.
\newblock \emph{Nature medicine}, 29\penalty0 (8):\penalty0 1930--1940, 2023.

\bibitem[T{\"o}rnberg(2023)]{tornberg2023use}
Petter T{\"o}rnberg.
\newblock How to use llms for text analysis.
\newblock \emph{arXiv preprint arXiv:2307.13106}, 2023.

\bibitem[T{\"o}rnberg(2024{\natexlab{a}})]{tornberg2024best}
Petter T{\"o}rnberg.
\newblock Best practices for text annotation with large language models.
\newblock \emph{arXiv preprint arXiv:2402.05129}, 2024{\natexlab{a}}.

\bibitem[T{\"o}rnberg(2024{\natexlab{b}})]{tornberg2024large}
Petter T{\"o}rnberg.
\newblock Large language models outperform expert coders and supervised classifiers at annotating political social media messages.
\newblock \emph{Social Science Computer Review}, page 08944393241286471, 2024{\natexlab{b}}.

\bibitem[Tseng et~al.(2025)Tseng, Chen, Chen, and Chen]{tseng2025evaluating}
Yu-Min Tseng, Wei-Lin Chen, Chung-Chi Chen, and Hsin-Hsi Chen.
\newblock Evaluating large language models as expert annotators.
\newblock In \emph{Second Conference on Language Modeling}, 2025.
\newblock URL \url{https://openreview.net/forum?id=DktAODDdbt}.

\bibitem[Törnberg(2024)]{Törnberg2024Large}
Petter Törnberg.
\newblock Large language models outperform expert coders and supervised classifiers at annotating political social media messages.
\newblock \emph{Social Science Computer Review}, 0\penalty0 (0):\penalty0 08944393241286471, 2024.
\newblock \doi{10.1177/08944393241286471}.
\newblock URL \url{https://doi.org/10.1177/08944393241286471}.

\bibitem[University~of London(2024)]{UniversityofLondonNational_Child_Development_Study2024}
Centre for Longitudinal~Studies University~of London, Institute of~Education.
\newblock National child development study: Age 11, sweep 2, sample of essays, 1969.
\newblock SN 5790, 2024.
\newblock \doi{10.5255/UKDA-SN-5790-2}.
\newblock URL \url{http://doi.org/10.5255/UKDA-SN-5790-2}.

\bibitem[Vaccaro et~al.(2024)Vaccaro, Almaatouq, and Malone]{vaccaro2024combinations}
Michelle Vaccaro, Abdullah Almaatouq, and Thomas Malone.
\newblock When combinations of humans and ai are useful: A systematic review and meta-analysis.
\newblock \emph{Nature Human Behaviour}, 8\penalty0 (12):\penalty0 2293--2303, 2024.

\bibitem[van~der Meer et~al.(2024)van~der Meer, Falk, Murukannaiah, and Liscio]{van-der-meer-etal-2024-annotator}
Michiel van~der Meer, Neele Falk, Pradeep~K. Murukannaiah, and Enrico Liscio.
\newblock Annotator-centric active learning for subjective {NLP} tasks.
\newblock In Yaser Al-Onaizan, Mohit Bansal, and Yun-Nung Chen, editors, \emph{Proceedings of the 2024 Conference on Empirical Methods in Natural Language Processing}, pages 18537--18555, Miami, Florida, USA, November 2024. Association for Computational Linguistics.
\newblock \doi{10.18653/v1/2024.emnlp-main.1031}.
\newblock URL \url{https://aclanthology.org/2024.emnlp-main.1031/}.

\bibitem[Wais(2006)]{genderizeR}
Kamil Wais.
\newblock {Gender Prediction Methods Based on First Names with genderizeR}.
\newblock \emph{{The R Journal}}, 8\penalty0 (1):\penalty0 17--37, 2006.
\newblock \doi{10.32614/RJ-2016-002}.
\newblock URL \url{https://doi.org/10.32614/RJ-2016-002}.

\bibitem[Wallach et~al.(2025)Wallach, Desai, Cooper, Wang, Atalla, Barocas, Blodgett, Chouldechova, Corvi, Dow, Garcia-Gathright, Olteanu, Pangakis, Reed, Sheng, Vann, Vaughan, Vogel, Washington, and Jacobs]{wallach2025position}
Hanna Wallach, Meera Desai, A.~Feder Cooper, Angelina Wang, Chad Atalla, Solon Barocas, Su~Lin Blodgett, Alexandra Chouldechova, Emily Corvi, P.~Alex Dow, Jean Garcia-Gathright, Alexandra Olteanu, Nicholas~J Pangakis, Stefanie Reed, Emily Sheng, Dan Vann, Jennifer~Wortman Vaughan, Matthew Vogel, Hannah Washington, and Abigail~Z. Jacobs.
\newblock Position: Evaluating generative {AI} systems is a social science measurement challenge.
\newblock In \emph{Forty-second International Conference on Machine Learning Position Paper Track}, 2025.
\newblock URL \url{https://openreview.net/forum?id=1ZC4RNjqzU}.

\bibitem[Wan et~al.(2024)Wan, Safavi, Jauhar, Kim, Counts, Neville, Suri, Shah, White, Yang, Andersen, Buscher, Joshi, and Rangan]{Wan2024TnT}
Mengting Wan, Tara Safavi, Sujay~Kumar Jauhar, Yujin Kim, Scott Counts, Jennifer Neville, Siddharth Suri, Chirag Shah, Ryen~W. White, Longqi Yang, Reid Andersen, Georg Buscher, Dhruv Joshi, and Nagu Rangan.
\newblock Tnt-llm: Text mining at scale with large language models.
\newblock In \emph{Proceedings of the 30th ACM SIGKDD Conference on Knowledge Discovery and Data Mining}, KDD '24, page 5836–5847, New York, NY, USA, 2024. Association for Computing Machinery.
\newblock ISBN 9798400704901.
\newblock \doi{10.1145/3637528.3671647}.
\newblock URL \url{https://doi.org/10.1145/3637528.3671647}.

\bibitem[Wang et~al.(2021)Wang, Liu, Xu, Zhu, and Zeng]{wang-etal-2021-want-reduce}
Shuohang Wang, Yang Liu, Yichong Xu, Chenguang Zhu, and Michael Zeng.
\newblock Want to reduce labeling cost? {GPT}-3 can help.
\newblock In Marie-Francine Moens, Xuanjing Huang, Lucia Specia, and Scott Wen-tau Yih, editors, \emph{Findings of the Association for Computational Linguistics: EMNLP 2021}, pages 4195--4205, Punta Cana, Dominican Republic, November 2021. Association for Computational Linguistics.
\newblock \doi{10.18653/v1/2021.findings-emnlp.354}.
\newblock URL \url{https://aclanthology.org/2021.findings-emnlp.354/}.

\bibitem[Wang et~al.(2024{\natexlab{a}})Wang, Kim, Rahman, Mitra, and Miao]{Wang2024Human}
Xinru Wang, Hannah Kim, Sajjadur Rahman, Kushan Mitra, and Zhengjie Miao.
\newblock Human-llm collaborative annotation through effective verification of llm labels.
\newblock In \emph{Proceedings of the 2024 CHI Conference on Human Factors in Computing Systems}, CHI '24, New York, NY, USA, 2024{\natexlab{a}}. Association for Computing Machinery.
\newblock ISBN 9798400703300.
\newblock \doi{10.1145/3613904.3641960}.
\newblock URL \url{https://doi.org/10.1145/3613904.3641960}.

\bibitem[Wang et~al.(2024{\natexlab{b}})Wang, Ma, Zhang, Ni, Chandra, Guo, Ren, Arulraj, He, Jiang, Li, Ku, Wang, Zhuang, Fan, Yue, and Chen]{MMLU_PRO_NEURIPS2024}
Yubo Wang, Xueguang Ma, Ge~Zhang, Yuansheng Ni, Abhranil Chandra, Shiguang Guo, Weiming Ren, Aaran Arulraj, Xuan He, Ziyan Jiang, Tianle Li, Max Ku, Kai Wang, Alex Zhuang, Rongqi Fan, Xiang Yue, and Wenhu Chen.
\newblock Mmlu-pro: A more robust and challenging multi-task language understanding benchmark.
\newblock In \emph{Advances in Neural Information Processing Systems}, volume~37, pages 95266--95290, 2024{\natexlab{b}}.
\newblock URL \url{https://proceedings.neurips.cc/paper_files/paper/2024/file/ad236edc564f3e3156e1b2feafb99a24-Paper-Datasets_and_Benchmarks_Track.pdf}.

\bibitem[Weber and Reichardt(2023)]{weber2023evaluation}
Maximilian Weber and Merle Reichardt.
\newblock Evaluation is all you need. prompting generative large language models for annotation tasks in the social sciences. a primer using open models.
\newblock \emph{arXiv preprint arXiv:2401.00284}, 2023.

\bibitem[Weber(1990)]{weber1990basic}
Robert~Philip Weber.
\newblock \emph{Basic content analysis}, volume~49.
\newblock Sage, 1990.

\bibitem[Weerasooriya et~al.(2023)Weerasooriya, Dutta, Ranasinghe, Zampieri, Homan, and KhudaBukhsh]{weerasooriya-etal-2023-vicarious}
Tharindu Weerasooriya, Sujan Dutta, Tharindu Ranasinghe, Marcos Zampieri, Christopher Homan, and Ashiqur KhudaBukhsh.
\newblock Vicarious offense and noise audit of offensive speech classifiers: Unifying human and machine disagreement on what is offensive.
\newblock In Houda Bouamor, Juan Pino, and Kalika Bali, editors, \emph{Proceedings of the 2023 Conference on Empirical Methods in Natural Language Processing}, pages 11648--11668, Singapore, December 2023. Association for Computational Linguistics.
\newblock \doi{10.18653/v1/2023.emnlp-main.713}.
\newblock URL \url{https://aclanthology.org/2023.emnlp-main.713/}.

\bibitem[Wei et~al.(2022)Wei, Bosma, Zhao, Guu, Yu, Lester, Du, Dai, and Le]{wei2022finetuned}
Jason Wei, Maarten Bosma, Vincent Zhao, Kelvin Guu, Adams~Wei Yu, Brian Lester, Nan Du, Andrew~M. Dai, and Quoc~V Le.
\newblock Finetuned language models are zero-shot learners.
\newblock In \emph{International Conference on Learning Representations}, 2022.
\newblock URL \url{https://openreview.net/forum?id=gEZrGCozdqR}.

\bibitem[Wei et~al.(2024)Wei, Yang, Song, Lu, Hu, Huang, Tran, Peng, Liu, Huang, Du, and Le]{NEURIPS2024_937ae0e8}
Jerry Wei, Chengrun Yang, Xinying Song, Yifeng Lu, Nathan Hu, Jie Huang, Dustin Tran, Daiyi Peng, Ruibo Liu, Da~Huang, Cosmo Du, and Quoc~V. Le.
\newblock Long-form factuality in large language models.
\newblock In A.~Globerson, L.~Mackey, D.~Belgrave, A.~Fan, U.~Paquet, J.~Tomczak, and C.~Zhang, editors, \emph{Advances in Neural Information Processing Systems}, volume~37, pages 80756--80827. Curran Associates, Inc., 2024.
\newblock URL \url{https://proceedings.neurips.cc/paper_files/paper/2024/file/937ae0e83eb08d2cb8627fe1def8c751-Paper-Conference.pdf}.

\bibitem[Weller and Seppi(2019)]{weller-seppi-2019-humor}
Orion Weller and Kevin Seppi.
\newblock Humor detection: A transformer gets the last laugh.
\newblock In Kentaro Inui, Jing Jiang, Vincent Ng, and Xiaojun Wan, editors, \emph{Proceedings of the 2019 Conference on Empirical Methods in Natural Language Processing and the 9th International Joint Conference on Natural Language Processing (EMNLP-IJCNLP)}, pages 3621--3625, Hong Kong, China, November 2019. Association for Computational Linguistics.
\newblock \doi{10.18653/v1/D19-1372}.
\newblock URL \url{https://aclanthology.org/D19-1372/}.

\bibitem[White(2000)]{White2000snooping}
Halbert White.
\newblock A reality check for data snooping.
\newblock \emph{Econometrica}, 68\penalty0 (5):\penalty0 1097--1126, 2000.
\newblock \doi{https://doi.org/10.1111/1468-0262.00152}.
\newblock URL \url{https://onlinelibrary.wiley.com/doi/abs/10.1111/1468-0262.00152}.

\bibitem[Wolf et~al.(2020)Wolf, Debut, Sanh, Chaumond, Delangue, Moi, Cistac, Rault, Louf, Funtowicz, Davison, Shleifer, von Platen, Ma, Jernite, Plu, Xu, Scao, Gugger, Drame, Lhoest, and Rush]{wolf-etal-2020-transformers}
Thomas Wolf, Lysandre Debut, Victor Sanh, Julien Chaumond, Clement Delangue, Anthony Moi, Pierric Cistac, Tim Rault, Rémi Louf, Morgan Funtowicz, Joe Davison, Sam Shleifer, Patrick von Platen, Clara Ma, Yacine Jernite, Julien Plu, Canwen Xu, Teven~Le Scao, Sylvain Gugger, Mariama Drame, Quentin Lhoest, and Alexander~M. Rush.
\newblock Transformers: State-of-the-art natural language processing.
\newblock In \emph{Proceedings of the 2020 Conference on Empirical Methods in Natural Language Processing: System Demonstrations}, pages 38--45, Online, October 2020. Association for Computational Linguistics.
\newblock URL \url{https://www.aclweb.org/anthology/2020.emnlp-demos.6}.

\bibitem[Wu et~al.(2024{\natexlab{a}})Wu, Guo, and Hooi]{Wu2024Fake}
Jiaying Wu, Jiafeng Guo, and Bryan Hooi.
\newblock Fake news in sheep's clothing: Robust fake news detection against llm-empowered style attacks.
\newblock In \emph{Proceedings of the 30th ACM SIGKDD Conference on Knowledge Discovery and Data Mining}, KDD '24, page 3367–3378, New York, NY, USA, 2024{\natexlab{a}}. Association for Computing Machinery.
\newblock ISBN 9798400704901.
\newblock \doi{10.1145/3637528.3671977}.
\newblock URL \url{https://doi.org/10.1145/3637528.3671977}.

\bibitem[Wu et~al.(2024{\natexlab{b}})Wu, Guo, and Hooi]{Wu2024SheepDog}
Jiaying Wu, Jiafeng Guo, and Bryan Hooi.
\newblock Fake news in sheep's clothing: Robust fake news detection against llm-empowered style attacks.
\newblock In \emph{Proceedings of the 30th ACM SIGKDD Conference on Knowledge Discovery and Data Mining}, KDD '24, page 3367–3378, New York, NY, USA, 2024{\natexlab{b}}. Association for Computing Machinery.
\newblock ISBN 9798400704901.
\newblock \doi{10.1145/3637528.3671977}.
\newblock URL \url{https://doi.org/10.1145/3637528.3671977}.

\bibitem[Wu et~al.(2023)Wu, Nagler, Tucker, and Messing]{wu2023large}
Patrick~Y Wu, Jonathan Nagler, Joshua~A Tucker, and Solomon Messing.
\newblock Large language models can be used to estimate the latent positions of politicians.
\newblock \emph{arXiv preprint arXiv:2303.12057}, 2023.

\bibitem[Xu et~al.(2024)Xu, Sun, Ren, Guo, Pan, Lin, Sun, and Han]{XU2024103665}
Ruoxi Xu, Yingfei Sun, Mengjie Ren, Shiguang Guo, Ruotong Pan, Hongyu Lin, Le~Sun, and Xianpei Han.
\newblock Ai for social science and social science of ai: A survey.
\newblock \emph{Information Processing \& Management}, 61\penalty0 (3):\penalty0 103665, 2024.
\newblock ISSN 0306-4573.
\newblock \doi{https://doi.org/10.1016/j.ipm.2024.103665}.
\newblock URL \url{https://www.sciencedirect.com/science/article/pii/S0306457324000256}.

\bibitem[Yao et~al.(2023)Yao, Jindal, Popa, Katsis, Ghosh, He, Lu, Srivastava, Li, Hendler, and Wang]{yao-etal-2023-beyond}
Bingsheng Yao, Ishan Jindal, Lucian Popa, Yannis Katsis, Sayan Ghosh, Lihong He, Yuxuan Lu, Shashank Srivastava, Yunyao Li, James Hendler, and Dakuo Wang.
\newblock Beyond labels: Empowering human annotators with natural language explanations through a novel active-learning architecture.
\newblock In Houda Bouamor, Juan Pino, and Kalika Bali, editors, \emph{Findings of the Association for Computational Linguistics: EMNLP 2023}, pages 11629--11643, Singapore, December 2023. Association for Computational Linguistics.
\newblock \doi{10.18653/v1/2023.findings-emnlp.778}.
\newblock URL \url{https://aclanthology.org/2023.findings-emnlp.778/}.

\bibitem[Yu et~al.(2023)Yu, Yang, Pelrine, Godbout, and Rabbany]{yu2023open}
Hao Yu, Zachary Yang, Kellin Pelrine, Jean~Francois Godbout, and Reihaneh Rabbany.
\newblock Open, closed, or small language models for text classification?
\newblock \emph{arXiv preprint arXiv:2308.10092}, 2023.

\bibitem[Zhang et~al.(2023)Zhang, Chen, and Yang]{zhang-etal-2023-mitigating}
Zhehao Zhang, Jiaao Chen, and Diyi Yang.
\newblock Mitigating biases in hate speech detection from a causal perspective.
\newblock In Houda Bouamor, Juan Pino, and Kalika Bali, editors, \emph{Findings of the Association for Computational Linguistics: EMNLP 2023}, pages 6610--6625, Singapore, December 2023. Association for Computational Linguistics.
\newblock \doi{10.18653/v1/2023.findings-emnlp.440}.
\newblock URL \url{https://aclanthology.org/2023.findings-emnlp.440/}.

\bibitem[Zhang et~al.(2022)Zhang, Strubell, and Hovy]{zhang-etal-2022-survey}
Zhisong Zhang, Emma Strubell, and Eduard Hovy.
\newblock A survey of active learning for natural language processing.
\newblock In Yoav Goldberg, Zornitsa Kozareva, and Yue Zhang, editors, \emph{Proceedings of the 2022 Conference on Empirical Methods in Natural Language Processing}, pages 6166--6190, Abu Dhabi, United Arab Emirates, December 2022. Association for Computational Linguistics.
\newblock \doi{10.18653/v1/2022.emnlp-main.414}.
\newblock URL \url{https://aclanthology.org/2022.emnlp-main.414/}.

\bibitem[Zhong et~al.(2022)Zhong, Snell, Klein, and Steinhardt]{pmlr-v162-zhong22a}
Ruiqi Zhong, Charlie Snell, Dan Klein, and Jacob Steinhardt.
\newblock Describing differences between text distributions with natural language.
\newblock In Kamalika Chaudhuri, Stefanie Jegelka, Le~Song, Csaba Szepesvari, Gang Niu, and Sivan Sabato, editors, \emph{Proceedings of the 39th International Conference on Machine Learning}, volume 162 of \emph{Proceedings of Machine Learning Research}, pages 27099--27116. PMLR, 17--23 Jul 2022.
\newblock URL \url{https://proceedings.mlr.press/v162/zhong22a.html}.

\bibitem[Zhou et~al.(2025)Zhou, Xu, Wang, Lu, Gao, and Ai]{Zhou_Xu_Wang_Lu_Gao_Ai_2025}
Yuhang Zhou, Paiheng Xu, Xiyao Wang, Xuan Lu, Ge~Gao, and Wei Ai.
\newblock Emojis decoded: Leveraging chatgpt for enhanced understanding in social media communications.
\newblock \emph{Proceedings of the International AAAI Conference on Web and Social Media}, 19\penalty0 (1):\penalty0 2302--2316, Jun. 2025.
\newblock \doi{10.1609/icwsm.v19i1.35935}.
\newblock URL \url{https://ojs.aaai.org/index.php/ICWSM/article/view/35935}.

\bibitem[Ziems et~al.(2024)Ziems, Held, Shaikh, Chen, Zhang, and Yang]{ziems-etal-2024-large}
Caleb Ziems, William Held, Omar Shaikh, Jiaao Chen, Zhehao Zhang, and Diyi Yang.
\newblock Can large language models transform computational social science?
\newblock \emph{Computational Linguistics}, 50\penalty0 (1):\penalty0 237--291, March 2024.
\newblock \doi{10.1162/coli_a_00502}.
\newblock URL \url{https://aclanthology.org/2024.cl-1.8/}.

\end{thebibliography}
